\documentclass[journal,twoside]{IEEEtran}
\IEEEoverridecommandlockouts
\usepackage{cite}
\ifCLASSINFOpdf
  \usepackage[pdftex]{graphicx}
  \graphicspath{{imgs/}}
  \DeclareGraphicsExtensions{.pdf,.jpeg,.png}
\else
  \usepackage[dvips]{graphicx}
  \graphicspath{{imgs/}}
  \DeclareGraphicsExtensions{.eps}
\fi
\usepackage[cmex10]{amsmath}
\usepackage{amssymb}
\ifCLASSOPTIONcompsoc
  \usepackage[caption=false,font=normalsize,labelfont=sf,textfont=sf]{subfig}
\else
  \usepackage[caption=false,font=footnotesize]{subfig}
\fi
\usepackage{abbrevs}
\usepackage{tikz}
\usepackage{url}

\newcommand{\tblcite}[1]{\begin{minipage}{0.55\linewidth} \cite{{#1}} \end{minipage}}
\newcommand{\rgb}{\textsc{rgb}}

\newcommand{\smallt}{``small object''\@\xspace} %State x
\newcommand{\xShape}[1]{\ensuremath{\mathbf{p}_{#1}}} %State x
\newcommand{\sxa}{\ensuremath{\mathbf{{x}}} } %State x
\newcommand{\sxk}{\ensuremath{\mathbf{{c}}} } %State x
\newcommand{\sxm}{\ensuremath{\mathbf{{d}}} } %State x

\newcommand*\tvect[1]{\big[#1\big]^T}

\newcommand{\rv}[1]{\ensuremath{{{#1}}}}

\newcommand{\mat}[1]{{\ensuremath{{\mathbf{#1}}}}}

\begin{document}
\tikzstyle{state}=[shape=circle,draw=blue!90,fill=blue!10,line width=1pt]
\title{Extended Object Tracking:\\ Introduction, Overview and Applications}
\author{Karl~Granstr\"om, Marcus~Baum, and Stephan~Reuter% <-this % stops a space
%\thanks{M. Shell is with the Department of Electrical and Computer Engineering, Georgia Institute of Technology, Atlanta, GA, 30332 USA e-mail: (see http://www.michaelshell.org/contact.html).}% <-this % stops a space
\thanks{Karl~Granstr\"om is with the Department of Signals and Systems, Chalmers University of Technology, Gothenburg, Sweden. E-mail: \texttt{karl.granstrom@chalmers.se}

Marcus~Baum is with the Institute of Computer Science, University of Goettingen, Goettingen, Germany. E-mail: \texttt{marcus.baum@cs.uni-goettingen.de}

Stephan~Reuter is with the Institute of Measurement, Control, and Microtechnology, Ulm University, Ulm, Germany. E-mail: \texttt{stephan.reuter@uni-ulm.de}}% 
}

%Telephone: (800) 555--1212, Fax: (888) 555--1212}


\maketitle

\begin{abstract}
This article provides an elaborate overview of  current research in extended object tracking.
We provide a clear definition of the extended object tracking problem and discuss its delimitation to other types of object tracking.
Next,  different aspects of extended object modelling are extensively discussed.  Subsequently, we give a  tutorial introduction to two basic and well used extended object tracking approaches -- the random matrix approach and the Kalman filter-based approach for star-convex  shapes. The next part treats the tracking of  multiple extended objects and elaborates how the large number of feasible association hypotheses can be tackled using both Random Finite Set (\rfs) and Non-\rfs multi-object trackers.
The article concludes with a summary of current applications, where four example applications involving camera, X-band radar, light detection and ranging (\lidar), red-green-blue-depth (\rgbd) sensors are highlighted. 
\end{abstract}


% \fi
\IEEEpeerreviewmaketitle

%\tableofcontents

%%%%%%%%%%%%%%%%%%%%%%%%%%%%%%%%%%%%%%%%%%%%%%%%%%%%%%%%%%%%%%%%%%%%%%%%%%%%%%%%%%%%%%%%%%
%%%%%%%%%%%%%%%%%%%%%%%%%%%%%%%%%%%%%%%%%%%%%%%%%%%%%%%%%%%%%%%%%%%%%%%%%%%%%%%%%%%%%%%%%%
%%%%%%%%%%%%%%%%%%%%%%%%%%%%%%%%%%%%%%%%%%%%%%%%%%%%%%%%%%%%%%%%%%%%%%%%%%%%%%%%%%%%%%%%%%

\section{Introduction}
Multiple Target Tracking (\mtt) denotes the process of successively determining the number and states of multiple dynamic objects based on noisy sensor measurements. Tracking systems are a key technology for  many technical applications in areas such as robotics, surveillance, autonomous driving, automation, medicine, and sensor networks.

Traditionally, \mtt algorithms have been tailored for scenarios with  multiple remote objects that are far away from the sensor, e.g., as in radar-based air surveillance. In such scenarios, an object is not always detected by the sensor, and if it is detected, at most one sensor resolution cell is occupied by the object. 
From traditional scenarios, specific assumptions on the mathematical model of  \mtt problems have evolved including the so-called \smallt assumptions: 
\begin{itemize}
 \item The objects evolve independently,
\item each object can be modelled as a point without any spatial extent, and
\item each object gives rise to at most a single measurement per time frame/scan. 
\end{itemize}

\mtt based on the \smallt assumptions is a highly complex problem due to sensor noise, missed detections, clutter detections, measurement origin uncertainty, and an unknown and time-varying number of targets. The most common approaches to \mtt are:
\begin{itemize}
	\item Multiple Hypothesis Tracking (\mht) \cite{Reid:1979,Kurien:1990,BlackmanP:1999},
	\item Joint Probabilistic Data Association (\jpda) \cite{BarShalom:1974,FortmannBSS:1983,BarShalomDH:2009},
	\item Probabilistic Multiple Hypothesis Tracking (\pmht) \cite{Streit1995,Willett2002}, and
	\item Random Finite Sets (\rfs) approaches \cite{mahler_book_2007,Mahler:2014}.
\end{itemize}
In the hypothesis-oriented \mht \cite{Reid:1979} and track-oriented \mht \cite{Kurien:1990}, the probability and log-likelihood ratio of a track, respectively, are calculated recursively. The \jpda type approaches blend data association probabilities on a scan-by-scan basis. The \pmht approach allows multiple measurement assignments to the same object\footnote{Note that allowing multiple assignments to the same object is in violation of the ``small object'' assumption, which assumes at most a single measurement per time frame/scan.}, which results in an efficient method using the Expectation-Maximization (EM) framework, see, e.g., \cite[Ch. 9]{Bishop:2006}.  The \rfs type approaches rely on modelling the objects and the measurements as random sets. A recent overview article about \mtt, with a main focus on small, so-called point objects, is given in \cite{VoMBSCOMV:2015}.

Today, there is still a huge variety of applications for which the \smallt assumptions are reasonable. 
However, due to  rapid advances in sensor technology in the recent years, it is becoming increasingly common that objects occupy several sensor resolution cells. Furthermore, novel applications with  objects in the near-field of sensors, e.g., in mobile robotics and autonomous driving, often render the  \smallt assumptions invalid.

The tracking of an object that might occupy more than one sensor cell leads to the so-called \emph{extended object tracking} or \emph{extended target tracking} problem.
In extended object tracking the objects give rise to a varying number of potentially noisy measurements from different spatially distributed measurement
sources, also referred to as reflection points. 
The shape of the object is usually unknown and can even vary over time, and the objective is to recursively determine the
shape of the object plus its kinematic parameters. Due to the nonlinearity of the resulting estimation problem, already tracking a single extended object is in general a highly complex problem for which elaborate nonlinear estimation techniques are required.

\begin{figure}
\center
\subfloat[Point object tracking example: Frame 1 (left) \& Frame 2 (right)]{
\begin{minipage}{8cm}\center
 \fbox{\includegraphics[page=1,width=0.44\columnwidth] {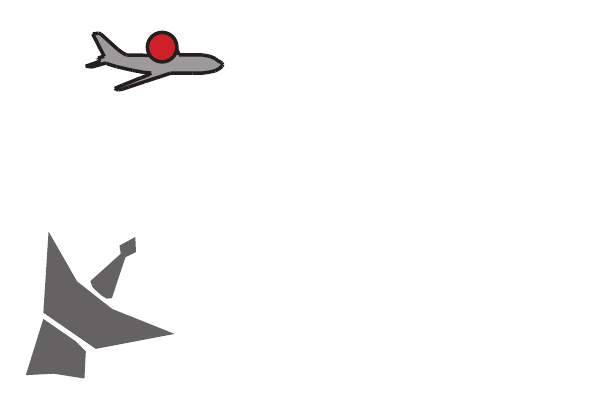}}
 \hspace{0.5cm}
 \fbox{\includegraphics[page=2,width=0.44\columnwidth] {PointVsExt2015_v2.pdf}}
 \label{fig:pointtarget}
 \end{minipage}
} 

\subfloat[Extended object tracking example: Frame 1 (left) \& Frame 2 (right)]{
\begin{minipage}{8cm}\center
 \fbox{\includegraphics[page=1,width=0.44\columnwidth] {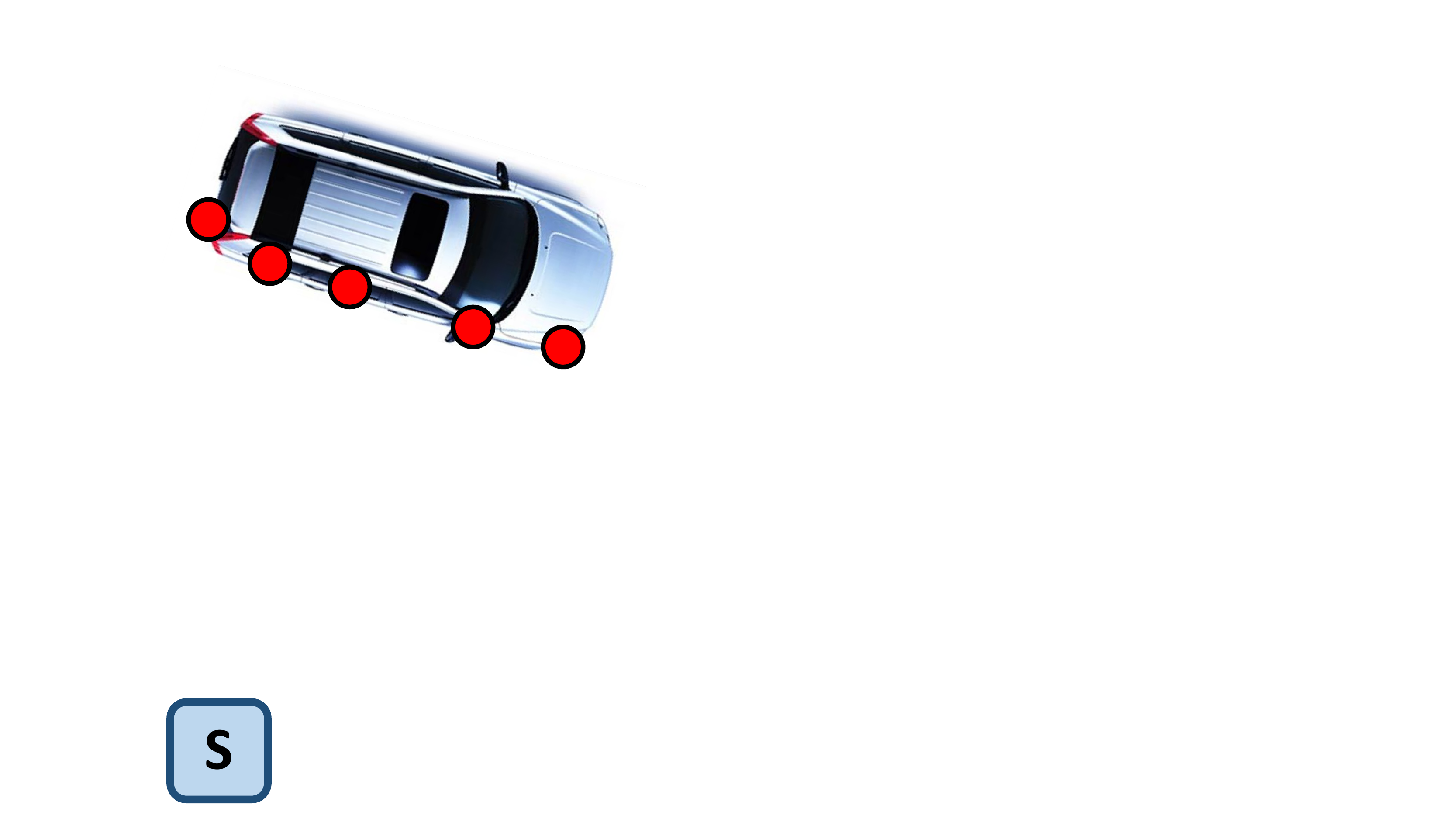}}
 \hspace{0.5cm}
 \fbox{\includegraphics[page=2,width=0.44\columnwidth] {ETTexampleLidarCar.pdf}}
 \label{fig:extendedtarget}
 \end{minipage}
} 

\subfloat[Contour Tracking example: Frame 1 (left) \& Frame 2 (right)]{
\begin{minipage}{8cm}\center
 \fbox{\includegraphics[page=5,width=0.44\columnwidth] {PointVsExt2015_v2.pdf}}
 \hspace{0.5cm}
 \fbox{\includegraphics[page=6,width=0.44\columnwidth] {PointVsExt2015_v2.pdf}}
 \label{fig:contourtracking}
 \end{minipage}
} 
\caption{Illustration of different types of tracking problems: a) In point object tracking, at most one measurement (red markers) per frame is received. b)  In extended object tracking, multiple measurements (red markers) from a varying number of measurement sources/reflection centers are obtained per frame. c) In contour tracking, a single contour (red) is extracted from each single image frame. Hence, one can say that in contour tracking, the measurements are contours, while in extended object tracking the measurements are (Cartesian) points. However, in both extended object tracking and contour tracking one aims at estimating  the shape, i.e., a contour, based on the received measurements.}
\label{fig:targetExtentModelTypes}
\end{figure}

Although often misunderstood -- extended object tracking, as defined above, is fundamentally different from  typical contour tracking problems in computer vision \cite{Yilmaz2006}.
In vision-based contour tracking \cite{Yilmaz2006}, a complete red-green-blue (\rgb) image is available at each time frame and one extracts a contour from each image that is tracked over time.
In extended object tracking, one works with  a few  (usually two or three-dimensional)  measurements per time step, i.e.,  a sparse point cloud. It is nearly always impossible to extract a shape only based on the measurement from one time instant.
The object shape can only be determined if measurements over several time steps are systematically accumulated and fused under incorporation of the (unknown) object motion and sensor noise. An illustration of the difference between point object tracking, extended object tracking, and contour tracking is given in Figure~\ref{fig:targetExtentModelTypes}.

In many practical applications it is necessary to track multiple extended objects, where no measurement-to-object associations are available. Unfortunately, data association becomes even more challenging in multiple \emph{extended} object tracking as a huge number of association events are possible: all possible partitions of the set of measurements have to be enumerated, followed by all possible ways to assign partition cells to object estimates. The first computationally feasible multi-extended object tracking algorithms have recently been developed, and rely on approximations of the partitioning problem in the context of \rfss.

The objective of this article is to
\begin{itemize}
\item[(i)] provide an elaborate and up-to-date introduction to the extended object tracking problem,
\item[(ii)] introduce basic concepts, models, and methods for  shape estimation of a single extended object,
\item[(iii)] introduce the basic concepts, models, and methods for tracking multiple extended objects,
\item[(iv)] point out recent applications and future trends.
\end{itemize}

Historically, the first works on extended object tracking can be traced back to \cite{Drummond1988,Drummond1990}.
Already in 2004, \cite{Waxman2004} gave a short literature overview of cluster (group) tracking and extended object tracking problems. However, since then, huge progress has been made in both shape estimation of a single object and multi-(extended)-object tracking.
An overview of Sequential Monte Carlo (SMC) methods for group and extended object tracking can be found in  \cite{MihaylovaCSGPG:2014}.
The focus of \cite{MihaylovaCSGPG:2014} lies on group object tracking and SMC methods. Hence, the content of  \cite{MihaylovaCSGPG:2014} is orthogonal to this article, and the two articles complement each other. A comparison of early versions of the random matrix and random hypersurface approach was performed in \cite{SDF10_Baum}. Since the publication of \cite{SDF10_Baum}, both methods have been significantly further developed.

The rest of the article is organised as follows. In the next section some definitions are introduced, and modelling of object shape, number of measurements, and object dynamics is overviewed. Section~\ref{sec:ShapeEstimation} overviews two popular approaches to extent modelling and estimation: the random matrix model, Section~\ref{sec:RandomMatrixModel}, and the random hypersurface model, Section~\ref{sec:RandomHypersurfaceModel}. Multiple extended object tracking is overviewed in Section~\ref{sec:MultipleExtendedTargetTracking}, and in Section~\ref{sec:ETTapplications} three applications are presented: tracking cars using a \lidar, marine vessel tracking using X-band radar,  tracking groups of pedestrians using a camera, and tracking complex shapes using a \rgbd sensor. The paper is concluded in Section~\ref{sec:Conclusions}.

\setlength{\fboxsep}{0pt}%
\setlength{\fboxrule}{1pt}%

%%%%%%%%%%%%%%%%%%%%%%%%%%%%%%%%%%%%%%%%%%%%%%%%%%%%%%%%%%%%%%%%%%%%%%%%%%%%%%%%%%%%%%%%%%
%%%%%%%%%%%%%%%%%%%%%%%%%%%%%%%%%%%%%%%%%%%%%%%%%%%%%%%%%%%%%%%%%%%%%%%%%%%%%%%%%%%%%%%%%%
%%%%%%%%%%%%%%%%%%%%%%%%%%%%%%%%%%%%%%%%%%%%%%%%%%%%%%%%%%%%%%%%%%%%%%%%%%%%%%%%%%%%%%%%%%

\section{Definitions and extended object modelling}
In this section we will first give a definition of the extended object tracking problem and some related types of object tracking. We will then overview extended object state modelling, measurement modelling, shape modelling, and dynamics modelling.

\subsection{Definitions}
% 
% In some literature, an extended object is understood to be an object that has a spatial extent, in comparison to a point object that does not have a spatial extent. The spatial extent then causes the possibility of multiple measurements, instead of just the single one. These two definitions are somewhat unfortunate though, because they are a poor representation of the underlying reality: 

In tracking problems the physical, real-world, objects-of-interest always have spatial extents. This is true for relatively large objects-of-interest, like ships, boat, cars, bicyclists, humans and animals, and it is true for relatively small objects-of-interest, like cells. The differences between extended object tracking and point object tracking is due to sensor properties, especially the sensor resolution, than object properties such as spatial extent. If the resolution, relative to the size of the objects, is high enough, then an object may occupy several resolution cells. Thus, each object may generate multiple detections per time step in this case. In other words, depending on the sensor properties, specifically the sensor resolution, different types of object tracking will arise, and it is therefore instructive to distinguish between different types of object tracking problems. The following are definitions of types of tracking problems that are relevant to this article.

\begin{itemize}
\item \emph{Point object tracking:}\\ Each object generates at most a single measurement per time step, i.e., a single resolution cell is occupied by an object.
\item \emph{Extended object tracking:}\\ Each object generates multiple measurements per time step and the measurements are spatially structured around the objects, i.e.,  multiple resolution cells are occupied by an object.  
\item \emph{Group object tracking:}\\ Each object generates multiple measurements per time step, and the measurements are spatially structured around the object. A group object consists of two or more \emph{subobjects} that share some common motion. Further, the objects are not tracked individually but are instead treated as a single entity. Thus, the group object occupies several resolution cells; each subobject may occupy either one or several resolution cells.
\item \emph{Tracking with multi-path propagation:}\\ Each object generates multiple measurements per time step that are due to multi-path propagation. Thus, the measurements are not spatially structured around the object.
\end{itemize}
All of the tracking approaches, except for point object tracking, assume the possibility of multiple measurements per target. Due to the required differences in motion and measurement modelling, we differentiate between the three tracking approaches rather than defining a single type called \emph{multi-detection tracking}. Most literature considers one type of tracking problem, however, for the same sensor it can be the case that when an object is far away from the sensor it occupies at most one resolution cell, but when it is closer to the sensor it occupies several resolution cells. 

The focus of the article lies on extended object tracking. However, we note that it is possible -- and quite common -- to employ extended object tracking methods to track the shape of a group object, see, e.g., \cite{MihaylovaCSGPG:2014} and the example in Section~\ref{sec:PedGroupTracking}. It is easy to see that extended object tracking and group object tracking are two very similar problems. However some distinctions can be made that warrant two definitions instead of just one.

In extended object tracking, each object is a single entity, e.g., a car, an airplane, a human, or an animal. Often the shape can be assumed to be a rigid body,\footnote{With the exception of the orientation of the extent, the size and shape of the object does not change over time} however extended objects with deformable extents is also possible. In group object tracking, each object is a collection of (smaller) objects that share some common dynamics, while still allowing for individual dynamics within the group. For example, in a group of pedestrians, there is an overall group motion, but the individual pedestrians may also shift their positions within the group.

The measurements from an extended object are caused by \textit{measurement sources}, which has different meaning depending on the sensor that is used and the types of objects that are tracked. In some cases, e.g., see \cite{HammarstrandLS:2012,Bordonaro2015,BordonaroWBSLB:2016_JAIF}, one can model a finite number of measurement sources, while in other cases it is better to model an infinite number of sources.  For example, in \cite{HammarstrandLS:2012} automotive radar is used to track cars, and the measurements are located around the wheelhouses of the tracked cars, i.e. there are four measurement sources. In \cite{ScheelGMRD:2014,ScheelRD:2016} {\lidar}s are used to track cars, and the measurements are then located on the chassi of the car. This can be interpreted as an infinite number of points that may act as measurement sources.

Note that certain sensors measure the object's cross-range and down-range extents (or similar object features), allowing for the extent (size and shape) of the object to be estimated, see e.g., \cite{SalmondP:2003,zhong2008_extTarg_IMM_RBUKF,AngelovaM:2008,sun2014modeling,sun2014modeling2,sun2014joint}.  However, by the definitions used here, this is not extended object tracking unless there are multiple such measurements.

Lastly, multi-path phenomenon occur, e.g., when data from over-the-horizon-radar (\textsc{othr}) is used, see, e.g.,  \cite{SathyanCAS:2013,HabtemariamTTMK:2013,TangCMMTK:2015}. An important difference between extended object tracking and tracking with multi-path phenomenon lies in the distribution of the measurements: for the plain multi-path problem a spatial distribution is not assumed.

\begin{figure}
	\centering
	
	\includegraphics[width=0.90\columnwidth]{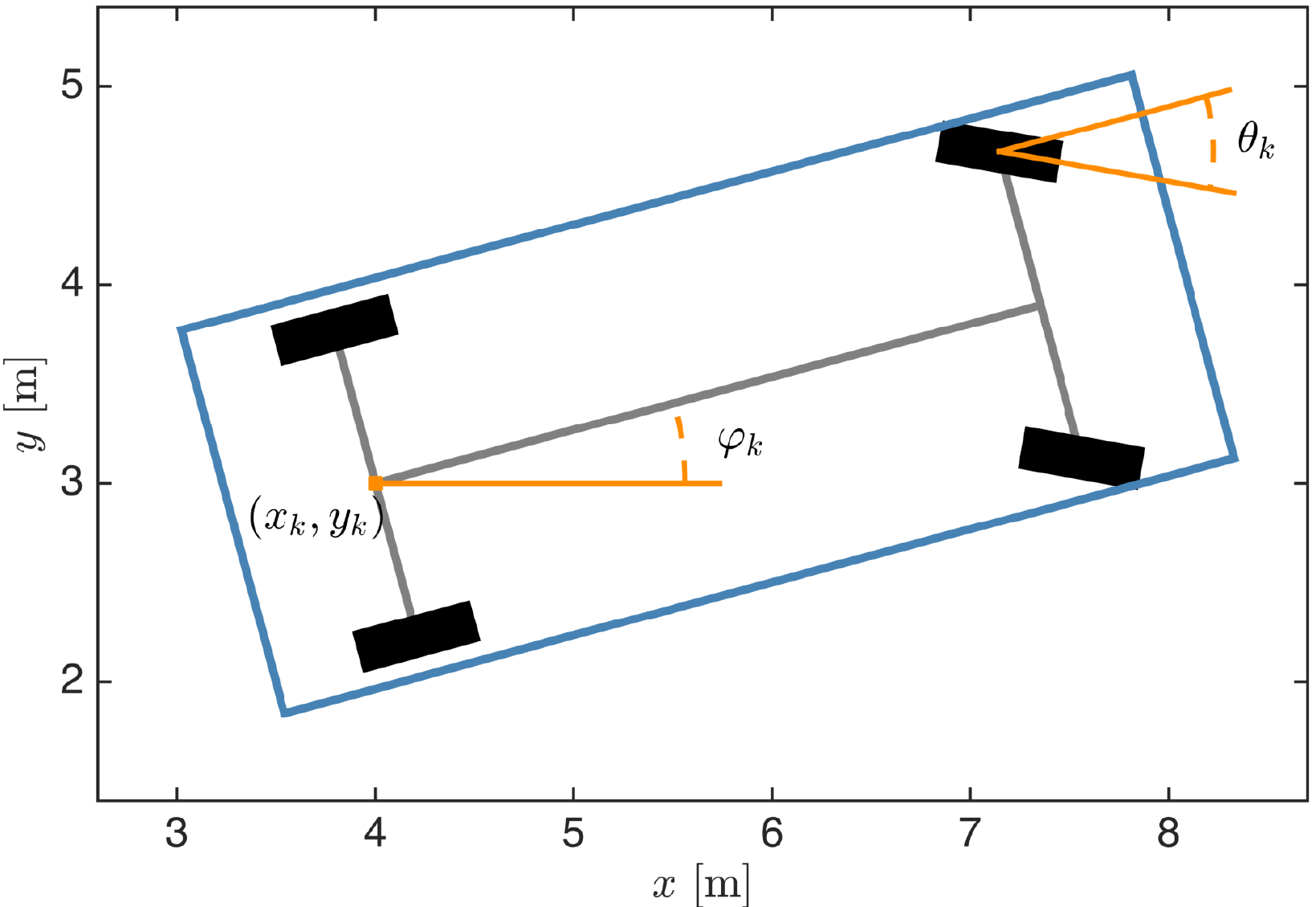}

	\caption{Example illustration of car state. The state vector $\sx$ models position $\cx,cy$, velocity $\cv$, heading $\varphi$, turning-angle $\theta$, length $\ell$ and width $w$. Note that velocity, length and width are not marked in the illustration.}
	\label{fig:illustrationOfCarState}
\end{figure}

\subsection{Object state}
\label{subsec:StateModels}

The extended object state models where the object is located, where it is going, and what its spatial extent (shape and size) is. The state typically includes the following:
\begin{itemize}
	\item \emph{Position:} Either $(x,y)$-position in  2D or $(x,y,z)$-position in 3D. 
	\item \emph{Kinematic state:} The motion parameters of the object, such as velocity, acceleration, heading and turn-rate. 
	\item \emph{Extent state:} Parameters that determine the shape and the size of the object, as well as the orientation of the shape. 
\end{itemize}
An example object state, appropriate for a car that is tracked using a horizontally mounted 2D \lidar sensor \cite{GranstromRMS:2014}, is illustrated in Figure~\ref{fig:illustrationOfCarState}. In this example the state vector at time step $k$, denoted $\sx_k$, is
\begin{align}
	\sx_{k} = \begin{bmatrix} \cx_{k} & \cy_{k} & \cv_{k} & \varphi_{k} & \theta_{k} & \ell_{k} & w_{k} \end{bmatrix}^{\tp} \label{eq:RectangularCarStateVector}
\end{align}
where $\cx_{k},\cy_{k}$ is 2D position, the kinematic state is comprised by velocity $\cv_{k}$, heading $\varphi_{k}$ and turning angle $\theta_{k}$, and the extent state is comprised by length $\ell_{k}$ and width $ w_{k}$. Note that the car's shape is assumed to be a rectangle, and the orientation this rectangular shape is assumed to be aligned with the car's heading. This state model is used in the car tracking example that is presented in Section~\ref{sec:LidarCarTracking}.

In general, exactly what parameters the object state includes---e.g., 2D or 3D position? Which kinematics? Any assumed shape?---depends very much on the type of object that is tracked, the type of sensor data that is used, and the type(s) of object motion that one wishes to describe.

For example, for tracking cars it is often sufficient to only model the 2D position on the road, while airborne objects typically require 3D position. The position state may coincide with the objects centre-of-mass, however this is not always the appropriate choice. When cars are tracked it is suitable to take the position as the mid-point on the rear-axle, because this facilitates the use of single-track-bicycle models in the motion modelling. Motion modelling, or dynamic modelling, for extended objects is address further in Section~\ref{sec:DynamicsModelling}.

If 2D position is modelled, the heading/orientation of the object can be described by a single angle, while 3D position may require more angles to accurately describe the heading/orientation, e.g., roll, pitch and yaw angles. Often the orientation of the extent is aligned with the heading, however this is not always the choice. For example, some motion models for cars include a so called slip angle that describes the angular difference between the car's heading and the orientation of the car's shape, see, e.g., \cite{SchrammHB:2014} for an introduction to vehicle dynamics modelling.

The extent state is determined by the type of shape that one wishes to describe; it could be a simple geometric shape like the rectangle used in Figure~\ref{fig:illustrationOfCarState}, or it could be a more general shape. There are many different alternatives for this, and an overview is given in Section~\ref{subsec:shapemodels}.

\subsection{Measurement modelling}
\label{subsec:MeasModels}

\begin{figure}
	\centering
	{
	\subfloat[]{\includegraphics[width=0.5\textwidth]{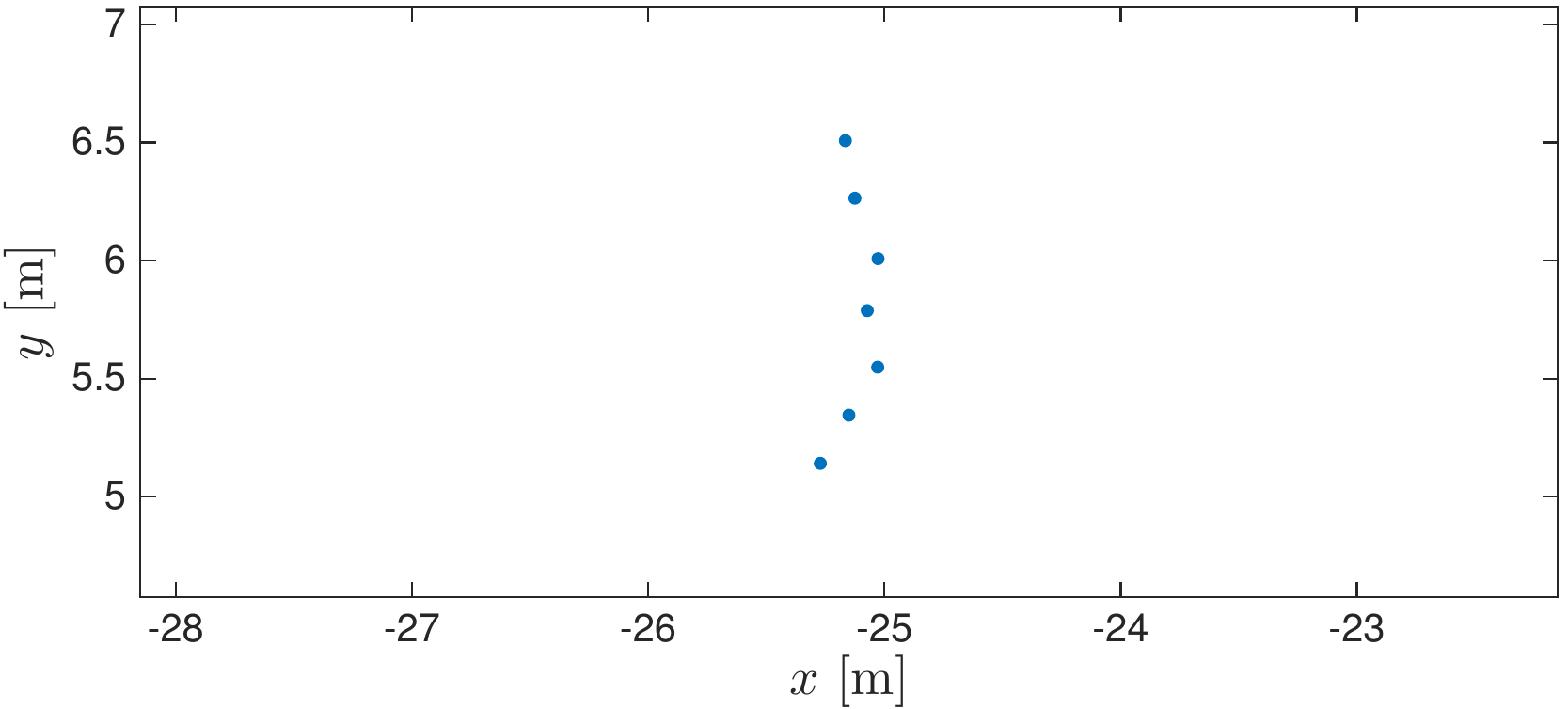}}
	\hfil
	\subfloat[]{\includegraphics[width=0.50\textwidth]{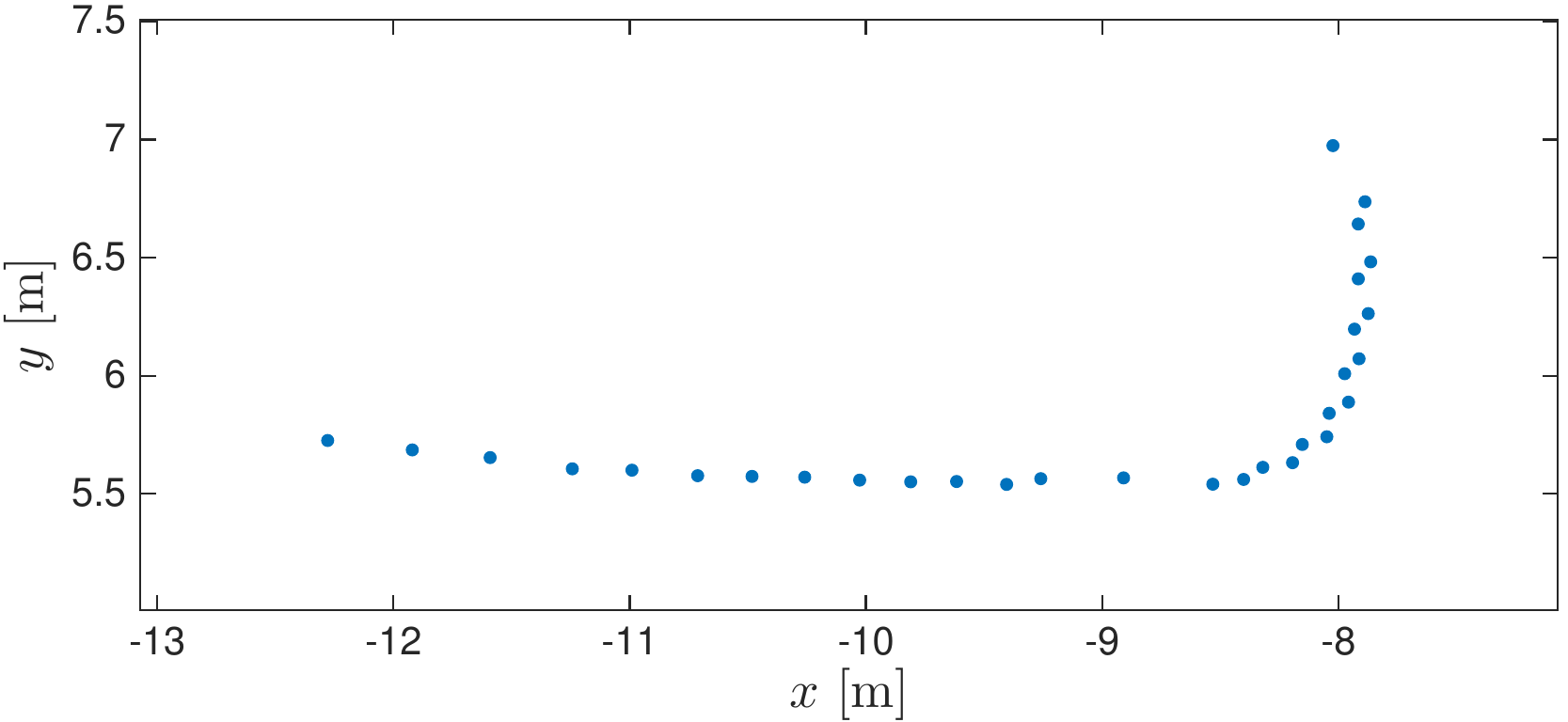}}
	\hfil
	\subfloat[]{\includegraphics[width=0.50\textwidth]{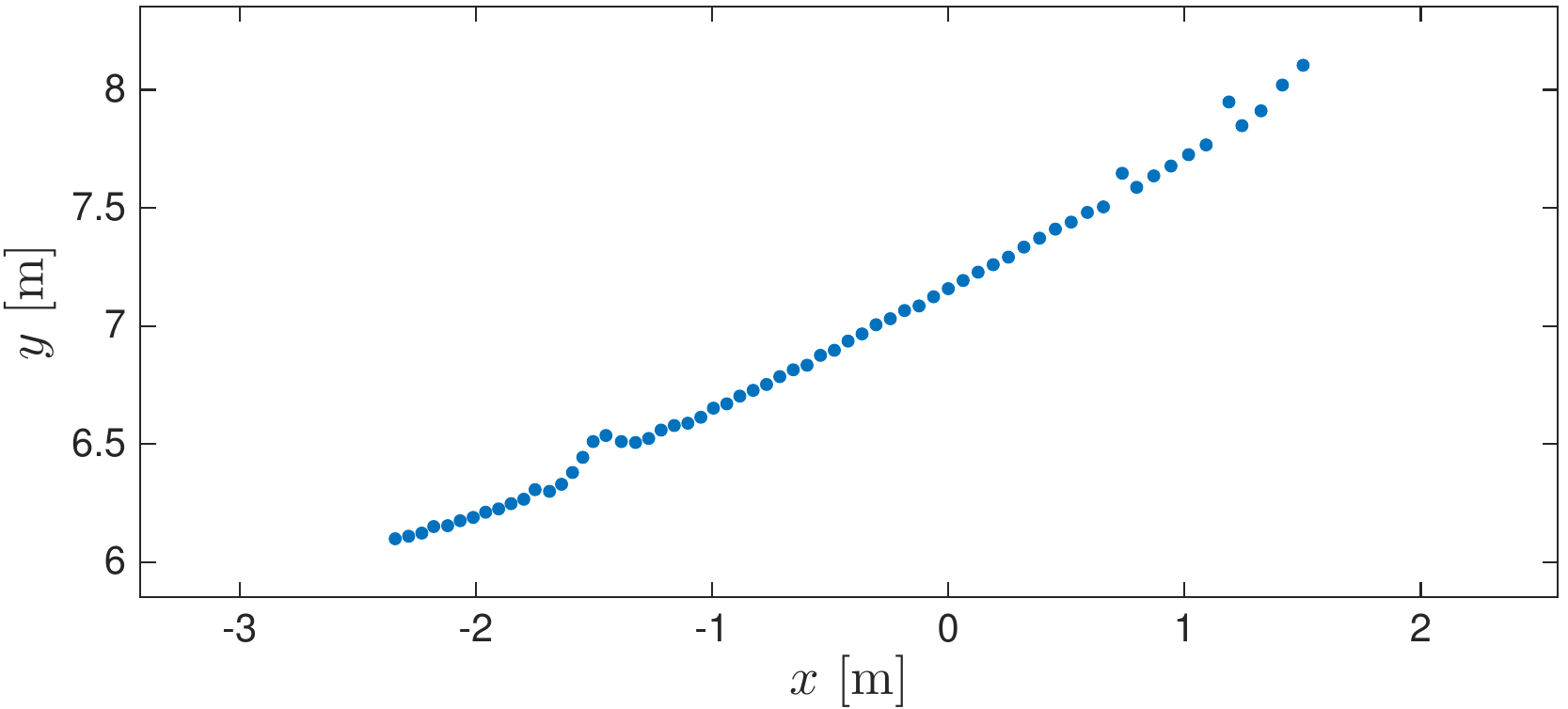}}
	}
	\caption{Example of real-world \lidar detections. The sensor is located in the origin, the measured object is a car. a)-c) shows detections from the same car from three different time steps. When the sensor-to-target geometry changes, the set of detections changes. In a) only the front side of the car is visible to the sensor, and the detections form a line. In b) the front and right sides are visible, and the detections (approximately) form an $L$-shape. In c) only the right side is visible to the sensor. Note that the car is farthest from the sensor in a), and closest in c).}
	\label{fig:CarLidar}
\end{figure}

Depending on what type of sensor is used, where the measured object is located w.r.t. the sensor, and how the object is oriented, the sensor will produce a different number of detections, originating from different points on the object. In addition to this, sensor noise will affect the detections, and all these properties have to be taken into account in the measurement modelling. 

An example with real-world \lidar data is given in Figure~\ref{fig:CarLidar}. Here the 2D-\lidar was used to track a car; in the Figure \lidar detections from three different time steps are shown. We can see that the number of detections, as well as their locations relative to the target, changes with the sensor-to-target geometry.

Due to sensor noise and model uncertainties, the measurement modelling is typically handled using probabilistic tools. Let the extended object state be denoted $\sx$, and let
\begin{align}
	\setZ=\left\{\sz^{(j)}\right\}_{j=1}^{n}
\end{align}
be a set of measurements that were caused by the object. Modelling the extended object measurements means to model the conditional distribution
\begin{align}
	p(\setZ | \sx), \label{eq:ExtObjMeasLik}
\end{align}
often referred to as the extended object measurement likelihood. The likelihood \eqref{eq:ExtObjMeasLik} needs to capture the number of detections, and how the detections are spatially distributed around the target state $\sx$. This modelling can be approached in several different ways; we overview the most common ways in the following sub-sections.

\subsubsection{Set of points on a rigid body}
\label{sec:SetOfRigidPoints}
One way is to model that the extended object has some number $L$ of reflection points\footnote{For some sensors, e.g., high resolution radar, the term \emph{scattering} point may be a more accurate description of the underlying sensor properties. Further, reflection \emph{source} may be a more accurate terminology in some cases, because the reflector may not be a discrete point but a larger structure, e.g., in automotive radar where the entire side of the car can be a reflector \cite{BuhrenY:2006}. However, reflection point appears to be the more common expression in extended object tracking literature, so in the remainder of the paper we adhere to this terminology.} located on a rigid body shape, as described in, e.g., \cite[Sec. 12.7.1]{mahler_book_2007}. We denote this as a Set of Points on a Rigid Body (\textsc{sprb}) model. 

\begin{figure}
	\centering
	
	\includegraphics[width=0.75\columnwidth]{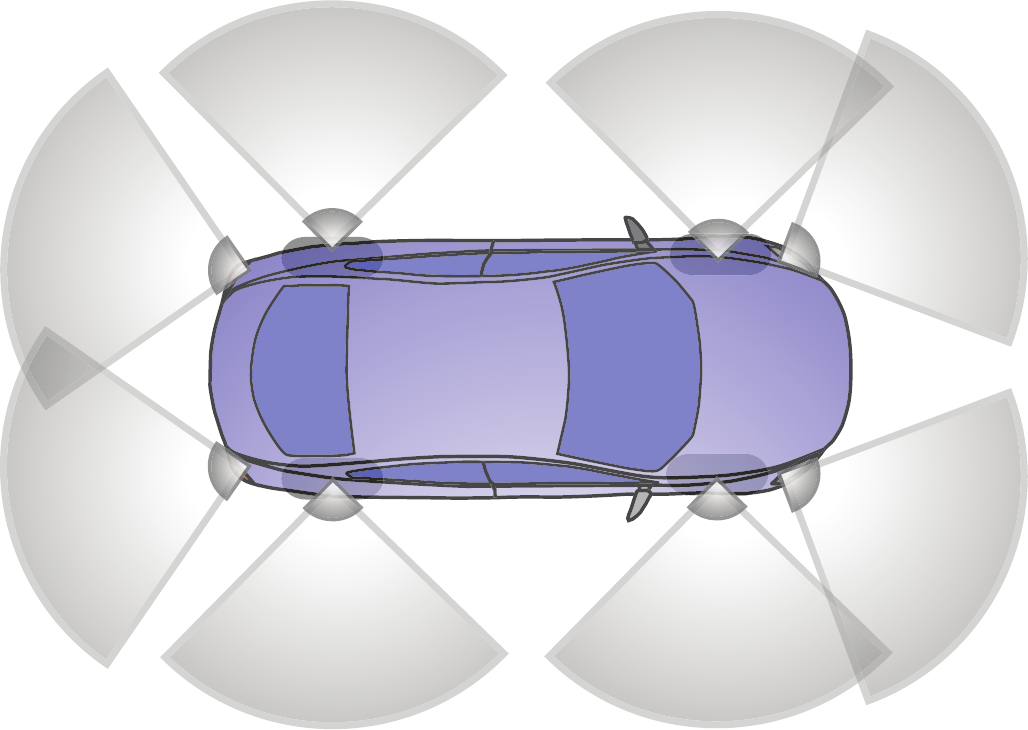}

\caption{Car with eight modelled radar reflection points: four points on the corners of the car, and four points on the wheel-houses. Also illustrated are the visibility regions. Image courtesy of Hammarstrand \etal \cite{HammarstrandSSS:2012}.}

	\label{fig:CarWithReflectionPoints}
\end{figure}

In \textsc{sprb} models the reflection points are detected independently of each other, and the $\ell$th reflection point has a detection probability $p_{D}^{\ell}$ that is a function of the object state. The measurement likelihood is \cite[Eq. 12.208]{mahler_book_2007}
\begin{align}
	p(\setZ | \sx) = \sum_{\theta} \prod_{\theta_{\ell}=0} \left(1-p_{D}^{\ell} \right) \prod_{\theta_{\ell}>0} p_{D}^{\ell} p^{\ell}\left(\sz^{(\theta_{\ell})} | \sx\right) \label{eq:ExtObjMeasLik_Lpoints}
\end{align}
if $|\setZ|\leq L$ and $p(\setZ | \sx) = 0$ otherwise. Here $|\setZ|$ is the cardinality of the measurement set, and $\theta$ is an assignment variable\footnote{$\theta_{\ell}=0$ means that the $\ell$th point is not associated to any measurement, and $\theta_{\ell}=j$ means that the $\ell$th point is associated to the $j$th measurement. Each measurement in $\setZ$ is associated to one of the reflection points, however no reflection point is associated to more than one measurement.}. In mathematical terms, the measurement process for each reflection point can be described as a Bernoulli \rfs \cite{mahler_book_2007,Mahler:2014}, and the measurement process for the extended object is a multi-Bernoulli \rfs \cite{mahler_book_2007,Mahler:2014}. 

\textsc{sprb} models were used in some early work on extended object tracking, see e.g., \cite{BroidaCC:1990,BroidaC:1991,HuangN:1994,Dezert:1998}, and were applied to data from vision sensors \cite{BroidaCC:1990,BroidaC:1991}. \textsc{sprb} modelling has also been applied to automotive radar, e.g., to model the reflection points on cars \cite{BuhrenY:2006,GunnarssonSDB:2007,HammarstrandSSS:2012}. An illustration of the $L=8$ automotive radar reflection points modelled in \cite{GunnarssonSDB:2007,HammarstrandSSS:2012} is shown in Figure~\ref{fig:CarWithReflectionPoints}.

A challenge with the \textsc{sprb} approach is that in a Bayesian estimation setting it requires data association between the $L$ points on the extended object and the target detections, see the summation over the assignments $\theta$ in \eqref{eq:ExtObjMeasLik_Lpoints}. This association problem can be quite challenging in settings where the number of points, and their respective locations on the object, are (highly) uncertain. There are some standard methods for handling association problems, such as finding the best assignment using the auction algorithm \cite{Bertsekas:1988}, finding the $M$ best assignments using Murty's algorithm \cite{Murty:1968}, or computing marginal association probabilities using, e.g, Probabilistic Data Associastion (\textsc{pda}) \cite{BarShalom:1974} or fast-\textsc{pda} \cite{Fitzgerald:1990}. A framework for handling the association uncertainty when automotive radar is used to track a single extended object is presented in \cite{HammarstrandLS:2012}.
In \cite{Bordonaro2015,BordonaroWBSLB:2016_JAIF}, the association problem for the \textsc{sprb} approach is by-passed by allowing more than one measurement from a point on the extended object and 
using the expectation maximization (EM) algorithm.

\subsubsection{Spatial model}
\label{sec:SpatialMeasurementModelling}
It was proposed by Gilholm \etal \cite{GilholmS:2005,GilholmGMS:2005} to model the target detections by an inhomogeneous Poisson Point Process (\ppp). This models the number of detections as Poisson distributed with a rate $\gamma(\sx)$ that is a function of the object's state, and the detections are spatially distributed around the target. By this means, the data association problem  is entirely avoided. The name \emph{spatial model} derives from the assumption that the detections are spatially distributed. In this model the measurement likelihood is \cite[Eq. 12.216]{mahler_book_2007}
\begin{align}
	p(\setZ | \sx) = e^{-\gamma(\sx)}  \gamma(\sx)^{|\setZ|} \prod_{\sz\in\setZ} p(\sz| \sx). \label{eq:PPPmeasLik}
\end{align}
Using a \ppp model is motivated in part by mathematical convenience -- it is simple to use in both single object and multiple object scenarios, and avoiding an explicit summation over associations between measurements and points on the object is very attractive \cite{GilholmS:2005,GilholmGMS:2005}.

The single measurement likelihood $p(\sz | \sx)$ in \eqref{eq:PPPmeasLik} is called spatial distribution, and it captures the structure of the measurements by using a model of the object extent and a model of the sensor noise. One alternative is to model $p(\sz | \sx)$ directly, e.g., using physics based modelling of the sensor. Another alternative is to model each detection $\sz$ as a noisy measurement of a source $\sy$ located somewhere on the object. The distribution $p(\sz|\sy)$ models the sensor noise, the distribution $p(\sy | \sx)$ models the extent and the spatial distribution $p(\sz | \sx)$ is given by the convolution 
\begin{align}
	p(\sz | \sx) = \int p(\sz | \sy) p(\sy | \sx) \diff \sy . \label{eq:ReflPointNoiseConvolution}
\end{align}
In other words, the measurement likelihood \eqref{eq:ReflPointNoiseConvolution} is the marginalization of the reflection point $\sy$ out of the estimation problem. For the noise model $p(\sz | \sy)$ the Gaussian distribution is a common choice, however other noise models are possible. An appropriate choice for the measurement source distribution $p(\sy | \sx)$ depends heavily on the type of sensor that is used and the representation of the object's shape.

In \cite[Sec. 2.3]{mahler_FUSION_2009_extTarg} the \ppp model \eqref{eq:PPPmeasLik} is interpreted to imply that the extended object is far enough away from the sensor for the measurements to resemble a cluster of points, rather than a structured ensemble. However, the \ppp model has been used successfully in multiple object scenarios where the object measurements show a high degree of structure, see, e.g., \cite{GranstromO:2012a,GranstromL:2013,GranstromRMS:2014}. 

Multiple extended target tracking using the \ppp model \eqref{eq:PPPmeasLik} has shown that the tracking results are sensitive to the state dependent Poisson rate $\gamma(\sx)$, see \cite{GranstromLO:2012}. The Bayesian conjugate prior for an unknown Poisson rate is the gamma distribution, see, e.g., \cite{GelmanCSR:2004}. By augmenting the state distribution with a gamma distribution for the Poisson rate, an individual Poisson rate can be estimated for each extended object \cite{GranstromO:2012c}.

\begin{figure}
	\includegraphics[width=0.5\columnwidth]{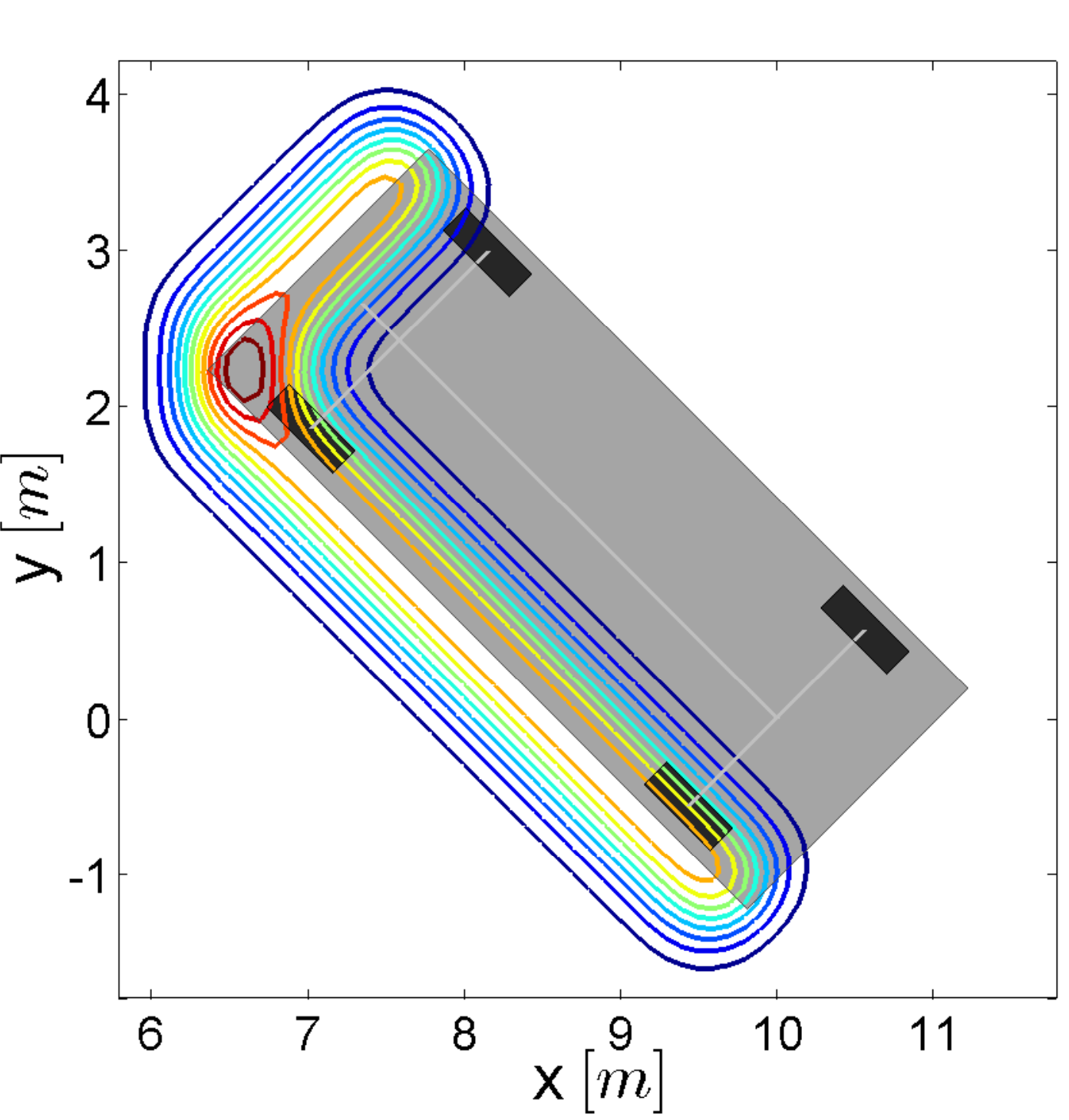}\includegraphics[width=0.5\columnwidth]{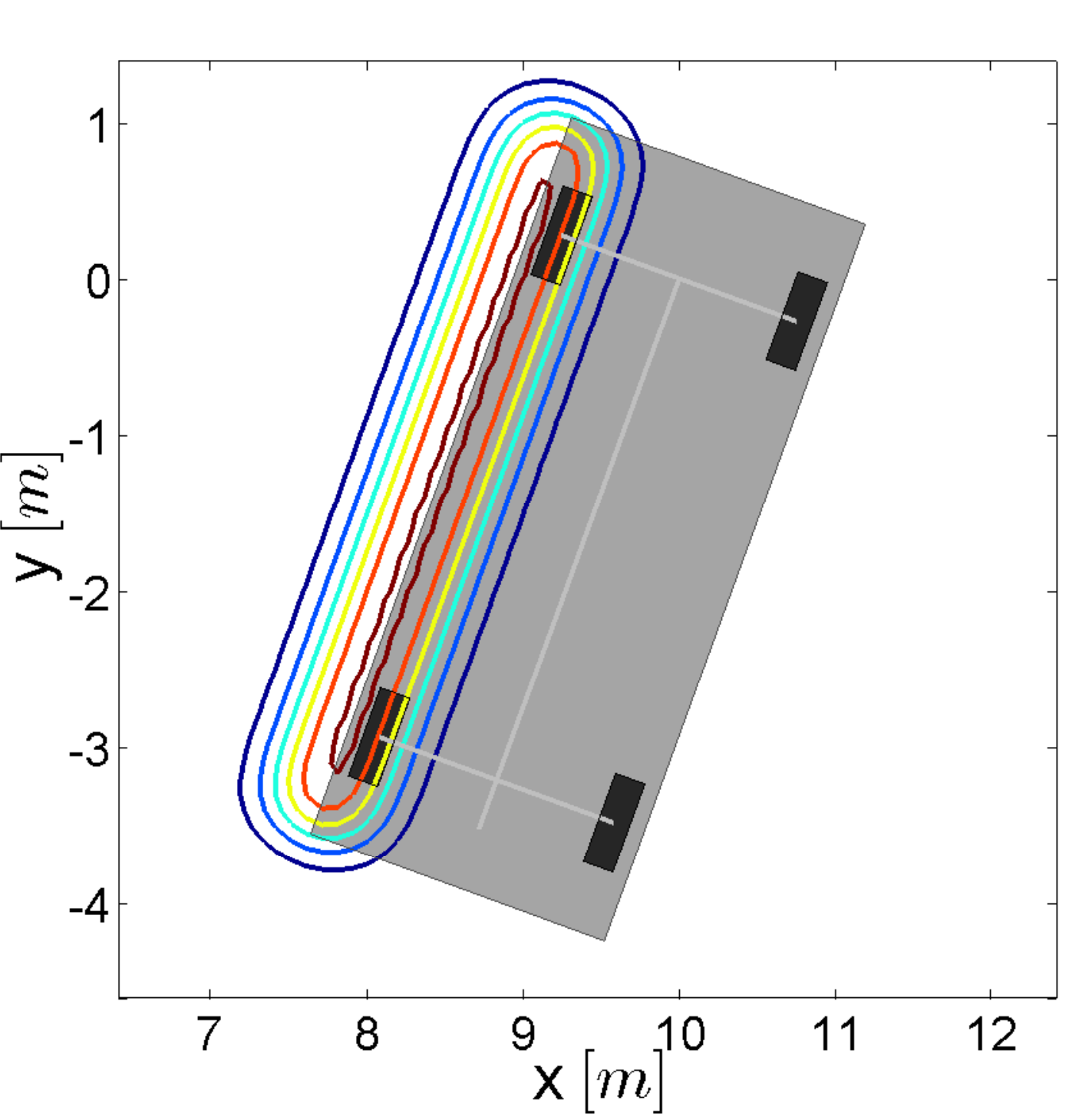}
	\caption{Example of the spatial measurement model. The sensor is a 2D \lidar located in the origin, and the tracked object is a car. The sensor can either recieve measurements from two sides (example on left), or measurements from one side (example on right).}
	\label{fig:Ex_Measurement_Likelihood}
\end{figure}

In \cite{GranstromRMS:2014} the \ppp spatial model was used to track cars using data from a 2D \lidar. The cars were modelled as rectangularly shaped, see \eqref{eq:RectangularCarStateVector} and Figure~\ref{fig:illustrationOfCarState}. The measurement modelling can be simplified by assuming that the \lidar measurements are located along either one side of the assumed rectangular car, or along two sides.  Example measurement likelihoods for these two cases are shown in Figure~\ref{fig:Ex_Measurement_Likelihood}. The source density $p(\sy|\sx)$ is assumed uniform along the sides that are visible to the sensor, and a Gaussian density was used for the noise $p(\sz|\sy)$.

A second alternative to the \textsc{sprb} model with $L$ reflection points is to use a spatial model where the number of detections is binomial distributed with parameters $L$ and $p_{D}$ \cite{RisticS:2013,RisticVVF:2013}, i.e., there is an implicit assumption that the probabilities of detection are equal for all $L$ points, $p_{D}^{\ell} \equiv p_{D},\ \forall \ell$. As in the \ppp model, the detections are spatially distributed around the target state. The measurement likelihood is \cite[Eq. 5]{RisticS:2013}
\begin{align}
	p(\setZ | \sx) = \frac{L!}{(L-|\setZ|)!} p_{D}^{|\setZ|}(1-p_{D})^{L-|\setZ|}  \prod_{\sz\in\setZ} p(\sz| \sx). \label{eq:BinomMeasLik}
\end{align}
if $|\setZ|\leq L$ and $p(\setZ | \sx) = 0$ otherwise. Note the considerable similarity to \eqref{eq:PPPmeasLik}: the difference is in the assumed model for the number of detections, and the single measurement likelihood $p(\sz| \sx)$ in \eqref{eq:BinomMeasLik} is analogous to $p(\sz| \sx)$ in \eqref{eq:PPPmeasLik}. For known $L$, the conjugate prior for an unknown $p_{D}$ is the beta distribution. Bayesian approaches to estimating unknown $L$ given a known $p_{D}$, or estimating both $L$ and $p_{D}$, have to the best of our knowledge not been presented. However, a simple heuristic for determining $L$, under the assumption that $p_{D}$ is known, is given in \cite{RisticS:2013}. 

In \cite{GilholmS:2005,GilholmGMS:2005} the Poisson assumption for the number of detections is not given much motivation using direct physical modelling of sensor properties. Similarly, in \cite{RisticS:2013,RisticVVF:2013} there is no physical modelling of sensor properties to motivate the binomial distribution model for the number of detections. Indeed, both models may be crude approximations for some sensor types, e.g., \lidar. Nevertheless, experiments with real-world data show that both models are applicable to many different sensor types, regardless of whether or not the number of detections are actually Poisson/binomial distributed.  The \ppp model has been used successfully with data from \lidar \cite{GranstromO:2012a,GranstromL:2013,GranstromRMS:2014}, radar \cite{GranstromNBLS:2014,GranstromNBLS:2015_TGARS}, and camera (see Section~\ref{sec:PedGroupTracking}). The binomial model has been used successfully with camera data \cite{RisticS:2013,RisticVVF:2013}.

\subsubsection{Physics based modelling}
In \cite{BuhrenY:2006,GunnarssonSDB:2007,HammarstrandLS:2012} \textsc{sprb} models for car tracking using automotive radars are derived using a physics based approach. Naturally, it is possible to use physical modelling of the sensor properties---both the modelling of the number of detections, and the modelling of the single measurement likelihood---to derive models that do not fit into the \textsc{sprb} model or the spatial model. For example, for a high resolution radar the number of measurements and their locations in the range-Doppler image can be reasonably predicted by deterministic electromagnetic theory, see, e.g., \cite{BhallaL:1995}. In \cite{KnillSD:2016} automotive radars are modelled using direct scattering, and this model is integrated into a multi-object framework in \cite{ScheelKRD:2016}. \lidar sensors can be modelled precisely using ray-tracing \cite{PetrovskayaT:2008} which facilitates the integration into multi-object tracking algorithms using the separable likelihood approach \cite{ScheelRD:2016}.

\subsection{Shape modelling}
\label{subsec:shapemodels}

\begin{figure}
\center

\subfloat[No shape model]{
\begin{minipage}{2.5cm}\center
 \includegraphics[page=1,scale=0.91] {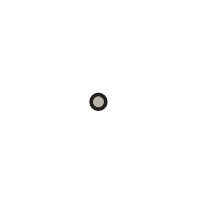}
 \label{fig:type1}
 \end{minipage}
} 
\subfloat[Basic geom. shape]{\begin{minipage}{2.5cm}\center
 \includegraphics[page=2,scale=0.91] {ExtObjTypes2015.pdf}
 \label{fig:type2} \end{minipage}
}  
\subfloat[Arbitrary shape]{\begin{minipage}{2.5cm}\center
 \includegraphics[page=3,scale=0.91] {ExtObjTypes2015.pdf}
 \label{fig:type3} \end{minipage}
} 
\caption{Illustration of the three levels of shape complexity. a) No shape model is used, the point corresponds to, e.g., the centre-of-mass. b) A basic geometric shape, such as an ellipse, is used to represent the extent of the target. c) An arbitrary shape model is used for the extent of the target. 
\label{fig:targetExtentShapeComplexities}}
\end{figure}

When it comes to modelling the shape of the object, it is useful to distinguish different complexity levels for describing the shape, because different shape complexities might require different approximations and algorithms. The different ways to model this type of extended object tracking scenario are here divided into three complexity levels:
\begin{enumerate}
 \item  The simplest level of modelling is to not model the shape at all, \iep to only estimate the object's kinematic properties. This approach has lowest computational complexity and the flexibility to track different type of objects is high because this model, even though it is simplistic in terms of object shape, is often applicable (with varying degree of accuracy).
\item  A more advanced level of modelling is to assume a specific basic geometric shape for the object, such as an ellipse, a line or a rectangle. 
\item The most advanced approach is to construct a measurement model that is capable of handling a broad variety of both different shapes and different measurement appearances. While such a model would be most general, it could also prove to be overly computationally complex.
\end{enumerate}
The three complexity levels are illustrated in Figure~\ref{fig:targetExtentShapeComplexities}, and some references whose shape modelling fall into the latter two categories are listed in Table~\ref{tab:TargetShapes}. 

\begin{table}[b]
	\centering
	\caption{Object shape (2D in 2D-space)}
	\label{tab:TargetShapes}
		\begin{tabular}{| l | l |}
			\hline
			Stick & \tblcite{GilholmS:2005,BaumFH:2012,Boersetal:06a,GranstromL:2013,VermaakIG:2005} \\
			\hline
			Circle & \tblcite{BaumKH:2010,PetrovMGA:2011,PetrovGMA:2012} \\
			\hline
			Ellipse & \tblcite{Koch:2008,BaumNH:2010,GranstromLO:2011,ZhuHL:2011,ReuterD:2011,DegermanWS:2011,LanRL:2012,ReuterWD:2012,SchusterRW:2015,AngelovaMPG:2013} \\
			\hline
			Rectangle & \tblcite{GranstromLO:2011,GranstromRMS:2014,KnillSD:2016} \\
			\hline
			Arbitrary shape & \tblcite{LundquistGO:11,BaumH:2011,LanRL:2012irregular,LanRL:2014,WahlstromO:2015,GranstromWBS:2015_MRMUK,CaoLL:2016,HirscherSRD:2016} \\
			\hline
	
	\end{tabular}
\end{table}

 The correct choice of complexity level is challenging and does not have a simple answer.
 In general, the more complex the shape, the more measurements (with less noise) are required to get a reasonable shape estimate. Furthermore, it depends on the type of sensor that is used, the types of objects, their motions, and what the tracking output will be used for. In some scenarios it may be sufficient to know the position of each object, in other scenarios it is necessary to have a detailed estimate of the size and shape of each object.

For example, in \cite{GranstromL:2013} it is shown that using \lidar data bicycles can be tracked fairly accurately without modelling the extent. However, estimation performance\footnote{Video with tracking results: \texttt{https://youtu.be/sGTGNkrprts}} is improved by using a spatial distribution model where the measurement source distribution, cf. $p(\sy|\sx)$ in \eqref{eq:ReflPointNoiseConvolution}, is modelled by a stick shape and uniform distribution and the noise distribution, cf. $p(\sz|\sy)$ in \eqref{eq:ReflPointNoiseConvolution}, is modelled by a Gaussian distribution. Specifically, by modelling the shape it becomes possible to capture rotations of the shape, and thus capture the onset of turning maneuvers. Without a shape estimate, the turning is captured at a later time \cite{GranstromL:2013}.

The 2D-\lidar bicycle tracking results are also an example of how a simple geometric shape, in this case a stick, combined with a simple Gaussian noise model, is a suitable measurement likelihood. A 2D stick shape is a crude approximation of the way a person riding a bicycle looks from a top-down perspective, however, here the stick shape is intended to model the measurement likelihood, and is not intended to be a nice visualization of the tracked bicyclist. Similarly, a rectangle shape is suitable when 2D-\lidar is used to track cars, see, e.g., \cite{GranstromLO:2011,GranstromRMS:2014,PetrovskayaT:2008}, even though many cars are only approximately rectangular. Another example is the ellipse-shape that is used to track boats and ships using marine radar in, e.g., \cite{GranstromNBLS:2014,GranstromNBLS:2015_TGARS,VivoneBGW:2016,VivoneBGNC:2015_ConvMeas,VivoneBGNC:2016_JAIF}. Typically neither boats, nor ships, are shaped like ellipses, however, the ellipse shape is suitable for the measurement modelling, and the estimated major and minor axes of the object ellipses are accurate estimates of the real-world lengths and widths of the boats/ships \cite{VivoneBGW:2016,VivoneBGNC:2015_ConvMeas,VivoneBGNC:2016_JAIF}.

{In some scenarios the objects have extents with shapes that cannot accurately be represented by a simple geometric shape like an ellipse or a rectangle. For estimation of arbitrary object shapes, the literature contains at least two different types of approaches: either the shape is modelled as a curve with some parametrization \cite{LundquistGO:11,BaumH:2011,WahlstromO:2015,CaoLL:2016,HirscherSRD:2016}, or the shape is modelled as combination of ellipses \cite{LanRL:2012irregular,LanRL:2014,GranstromWBS:2015_MRMUK}. When the shape is given a curve parametrization the noisy detections can be modelled using Gaussian processes \cite{WahlstromO:2015,HirscherSRD:2016}. Applied to car tracking using 2D-\lidar \cite{WahlstromO:2015,HirscherSRD:2016}, this allows for shape modelling with rounded corners, which is a more accurate model of actual cars than a rectangle with sharp corners is. The price of a more accurate model is an increased complexity: a general shape requires more parameters than a simple geometric shape. 

The increased complexity can be alleviated by utilizing the prior knowledge that cars are symmetric, see \cite{JAIF15_Faion} for a general concept to incorporate symmetries and \cite{HirscherSRD:2016} for a Gaussian process model example. Another approach to handling the complexity is to use different models at different distances from the sensor; in \cite{WyffelsC:JAIF} the priority of objects is ranked in three groups, specifying how accurately the different objects should be modelled. For example, for collision avoidance in autonomous driving, the objects closest to the ego-vehicle are more important than the distant objects, and this justifies ``taking'' computational resources from the distant objects and ``spending'' it on the closer objects.
}

\begin{figure}
 \center 
\subfloat[Closed curve (measurements from boundary).]{
\begin{minipage}{3.5cm}
\center
 \includegraphics[page=2,scale=0.91] {Inner2015.pdf}
 \label{fig:inner1}

\end{minipage}
 }  
 \hspace{0.5cm}
\subfloat[Closed curve (measurements from surface).]{
\begin{minipage}{3.5cm}\center
 \includegraphics[page=1,scale=0.91] {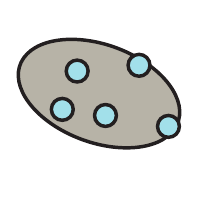}
 \label{fig:inner2}
\end{minipage}
}
\caption[Dimension of a  shape.]{One-dimensional (a) and two-dimensional (b) shape.\label{fig:targetExtentModelTypesInner}}
\end{figure}

In addition to modelling the shape itself, there are different ways to model how the measurements are spatially distributed over the shape. The types of extended object spatial distributions can be divided into two classes:
\begin{itemize}
	\item Measurements along the boundary of the object's extent. For measurements in 2D, this means that the measurements are noisy points on a curve. For measurements in 3D, the measurements are noisy points on either a curve or a surface. Measurements along the boundary are obtained, e.g., when \lidar is used in automotive applications.
	\item Measurements inside the volume/area of the object's extent, i.e., the measurements form a cluster.
	For example, two-dimensional radar detections of marine vessels can be interpreted as measurements from the inner of a two-dimensional shape, e.g., an ellipse, see \cite{GranstromNBLS:2015_TGARS} and Section~\ref{sec:XbandRadarTracking} for an experimental example.
\end{itemize}
In Table~\ref{tab:TargetShapesDim} some references are listed according to the shape dimension and measurement type, and Figure~\ref{fig:targetExtentModelTypesInner} provides an illustration. To our knowledge there is no explicit work about the estimation of 3D shapes in 3D space, probably because there are rarely sensors for this case. However, most algorithms for 2D shapes in 2D space can be generalized rather easily to the 3D case.

\begin{table}[b]
	\centering
	\caption{Shape Dimensions}
	\label{tab:TargetShapesDim}
		\begin{tabular}{| l | l |}
			\hline
			Curve in 2D/3D space: & \tblcite{Porrill1990,Zhang1997,GilholmS:2005,BaumFH:2012,Boersetal:06a,GranstromL:2013} \\
			\hline
			Surface in 2D space: & \tblcite{BaumNH:2010,DegermanWS:2011,GranstromLO:2011,Koch:2008,LanRL:2012,PetrovMGA:2011,ReuterD:2011,ReuterWD:2012,SchusterRW:2015,ZhuHL:2011} \\
			\hline
			Surface in 3D space: &\tblcite{Fusion12_Faion-CylinderTracking,Faion2015a} \\
			\hline
		\end{tabular}
\end{table}

When the measurements lie on the boundary of the extended object, the resulting theoretical problem shares similarities with  traditional curve fitting, where a curve is to be matched with noise points \cite{Fitzgibbon1999,Chernov2010}.
However, the curve fitting problem only considers static scenarios, i.e., non-moving curves. Additionally, the noise is usually isotropic and non-recursive non-Bayesian methods have been developed. Hence, curve fitting algorithms usually cannot directly be applied in the extended object tracking context. For a discussion of the rare Kalman filter-based approaches for curve fitting, we refer to \cite{Porrill1990,Zhang1997}.

To summarize the discussion about shape modelling, we note that it is important that the shape model is not only a reasonable representation of the true object shape but is also suitable for the measurement modelling, and that the shape model has a complexity that is appropriate for the sensor, the tracked object, and the computational resources.

\subsection{Dynamics modelling}
\label{sec:DynamicsModelling}
The object dynamic model describes how the object state evolves over time; for a moving object this describes how the object moves. This involves the position and the kinematic states that describe the motion---e.g., velocity, acceleration, turn-rate---however, it also involves descriptions of how the extent changes over time (typically it rotates when the object turns) and how the number of measurements changes over time (often there are more measurements the closer to the sensor the object is). 

There are two probabilistic parts to dynamics modelling that are important: the transition density and the Chapman-Kolmogorov equation. The transition density is denoted
\begin{align}
	p(\sx_{k+1} | \sx_{k}),
\end{align}
and describes the transition of the state from time step $k$ to time step $k+1$, i.e., from $\sx_{k}$ to $\sx_{k+1}$. The Chapman-Kolmogorov equation 
\begin{align}
	p(\sx_{k+1})  = \int p(\sx_{k+1} | \sx_{k}) p(\sx_{k}) \diff \sx_{k}. \label{eq:ChapmanKolmogorovEquation}
\end{align}
describes how, given a prior state density $p(\sx_{k})$ and a transition density, the predicted density $p(\sx_{k+1})$ is computed.

In many cases the dynamics for the position and the kinematic states can be modelled using any of the models that are standard in point object tracking, see \cite{RongLiJ:2003} for a comprehensive overview. Examples include the constant velocity (\cvmod), or white-noise acceleration, model, the (nearly)-constant acceleration (\camod) model, and the coordinated, or constant, turn (\ctmod) model. Detailed descriptions of \cvmod, \camod and \ctmod models are given in \cite{RongLiJ:2003}. When the tracked objects are cars, so called bicycle-models, introduced in \cite{RiekertS:1940}, are suitable for describing the target motion, see, e.g., \cite[Ch. 10--11]{SchrammHB:2014} for an overview and introduction to bicycle-models.

When the extended object is a rigid body its size and shape does not change over time, however the orientation of the shape (typically) rotates when the object turns. If the object is described by a set of points on a rigid body, see Section~\ref{sec:SetOfRigidPoints}, the point of rotation must be specified, and the centre-of-mass is a suitable choice. For the more common spatial models, see Section~\ref{sec:SpatialMeasurementModelling}, a typical assumption for the extent is to assume that its orientation is aligned with the heading of the object, e.g., this is the case in the bicycle models that are used in \cite{GranstromL:2013,GranstromRMS:2014}. When the heading and orientation are aligned the rotation of the extent does not have to be explicitly modelled as it is implicitly modelled by the object's heading. However, if this is not the case, the point of the rotation must be specified---again a suitable choice is the object's centre-of-mass. 

When there are multiple objects present a common assumption is that the objects evolve independently of each other, resulting in the object estimates being predicted independently. Obviously, an independent prediction may result in physically impossible (e.g. overlapping/intersecting) object state estimates. To better model target interactions one can use, e.g., social force modelling \cite{HelbingM:1995}; this is done in \cite{ReuterD:2011} where \lidar is used to track pedestrians. In group object tracking, where several objects form groups while remaining distinguishable, it is possible to apply, e.g., leader-follower models, allowing for the individual objects to be predicted dependently, see e.g. \cite{ClarkG:2007,Pang:LG2011}. A Markov Chain Monte Carlo (\textsc{mcmc}) approach to inferring interaction strengths between targets in groups is presented in \cite{MurphyOBG:2016}.

%%%%%%%%%%%%%%%%%%%%%%%%%%%%%%%%%%%%%%%%%%%%%%%%%%%%%%%%%%%%%%%%%%%%%%%%%%%%%%%%%%%%%%%%%%
%%%%%%%%%%%%%%%%%%%%%%%%%%%%%%%%%%%%%%%%%%%%%%%%%%%%%%%%%%%%%%%%%%%%%%%%%%%%%%%%%%%%%%%%%%
%%%%%%%%%%%%%%%%%%%%%%%%%%%%%%%%%%%%%%%%%%%%%%%%%%%%%%%%%%%%%%%%%%%%%%%%%%%%%%%%%%%%%%%%%%

\section{Tracking a single extended object}
\label{sec:ShapeEstimation}
In this section we overview some widely-used approaches for single extended object tracking, namely random matrix models and star-convex models. 

\subsection{Random Matrix Approach}
\label{sec:RandomMatrixModel}

The random matrix model was originally proposed by Koch \cite{Koch:2008}, and is an example of a spatial model (Section~\ref{sec:SpatialMeasurementModelling}). It models the extended object state as the combination of a kinematic state vector $\sx_{k}$ and an extent matrix\footnote{The book by Gupta and Nagar \cite{GuptaN:2000} is a good reference for various matrix variate distributions.} $\ext_{k}$. The vector $\sx_{k}$ represents the object's position and its motion properties, such as velocity, acceleration and turn-rate. The $d\times d$ matrix $\ext_{k}$ represents the object's extent, where $d$ is the dimension of the object; $d=2$ for tracking with 2D position and $d=3$ for tracking with 3D position. The matrix $\ext_{k}$ is modelled as being symmetric and positive definite, which implies that the object shape is approximated by an ellipse. The ellipse shape may seem limiting, however the model is applicable to many real scenarios, e.g. pedestrian tracking using \lidar \cite{GranstromO:2012a} and tracking of boats and ships using marine radar \cite{GranstromNBLS:2014,GranstromNBLS:2015_TGARS,VivoneBGW:2016,VivoneBGNC:2015_ConvMeas,VivoneBGNC:2016_JAIF,SchusterRW:2015}.

\begin{figure*}
	\centering
	{
	\subfloat[]{\includegraphics[width=0.2\textwidth]{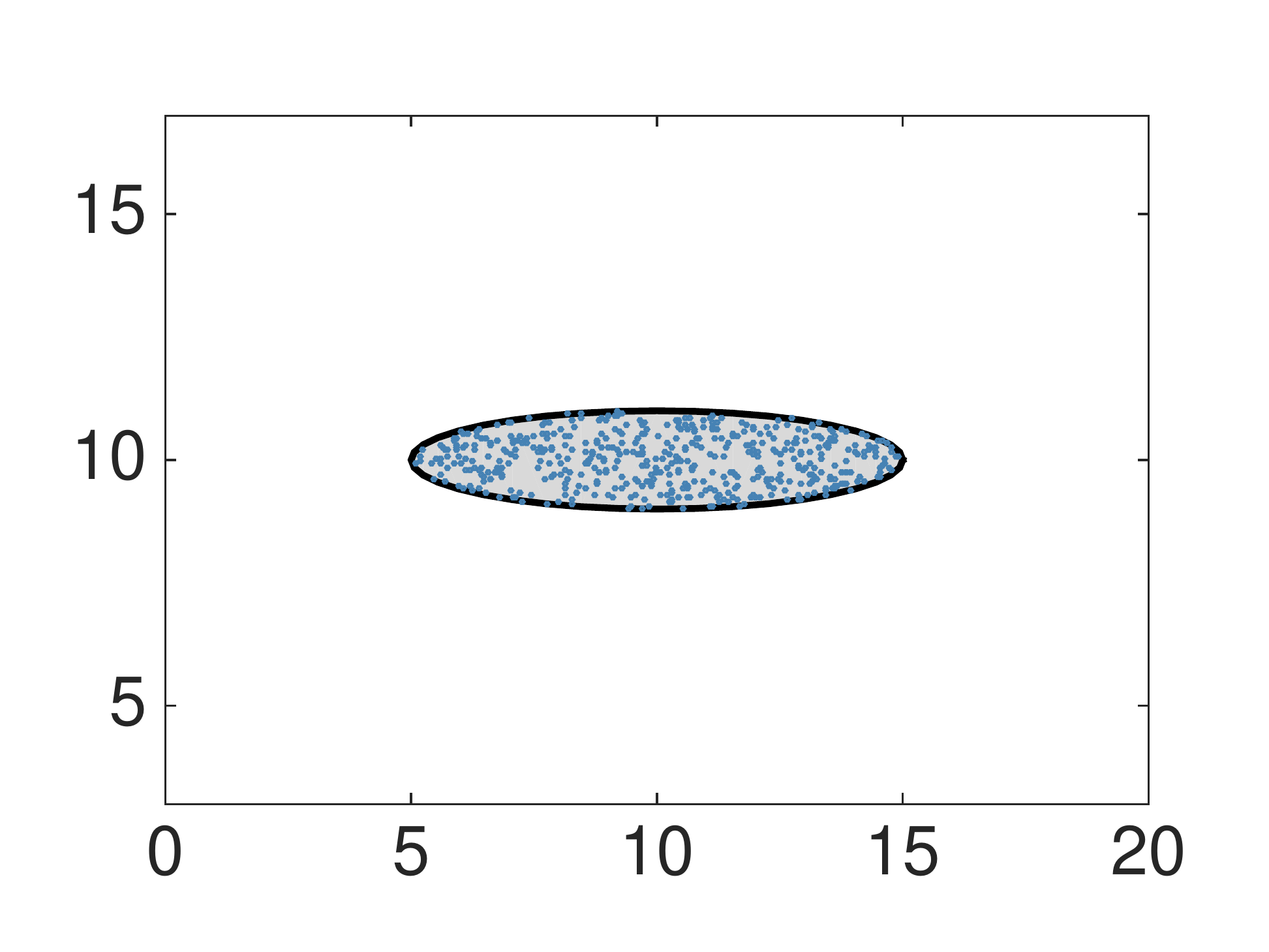}}
	\hfil
	\subfloat[]{\includegraphics[width=0.2\textwidth]{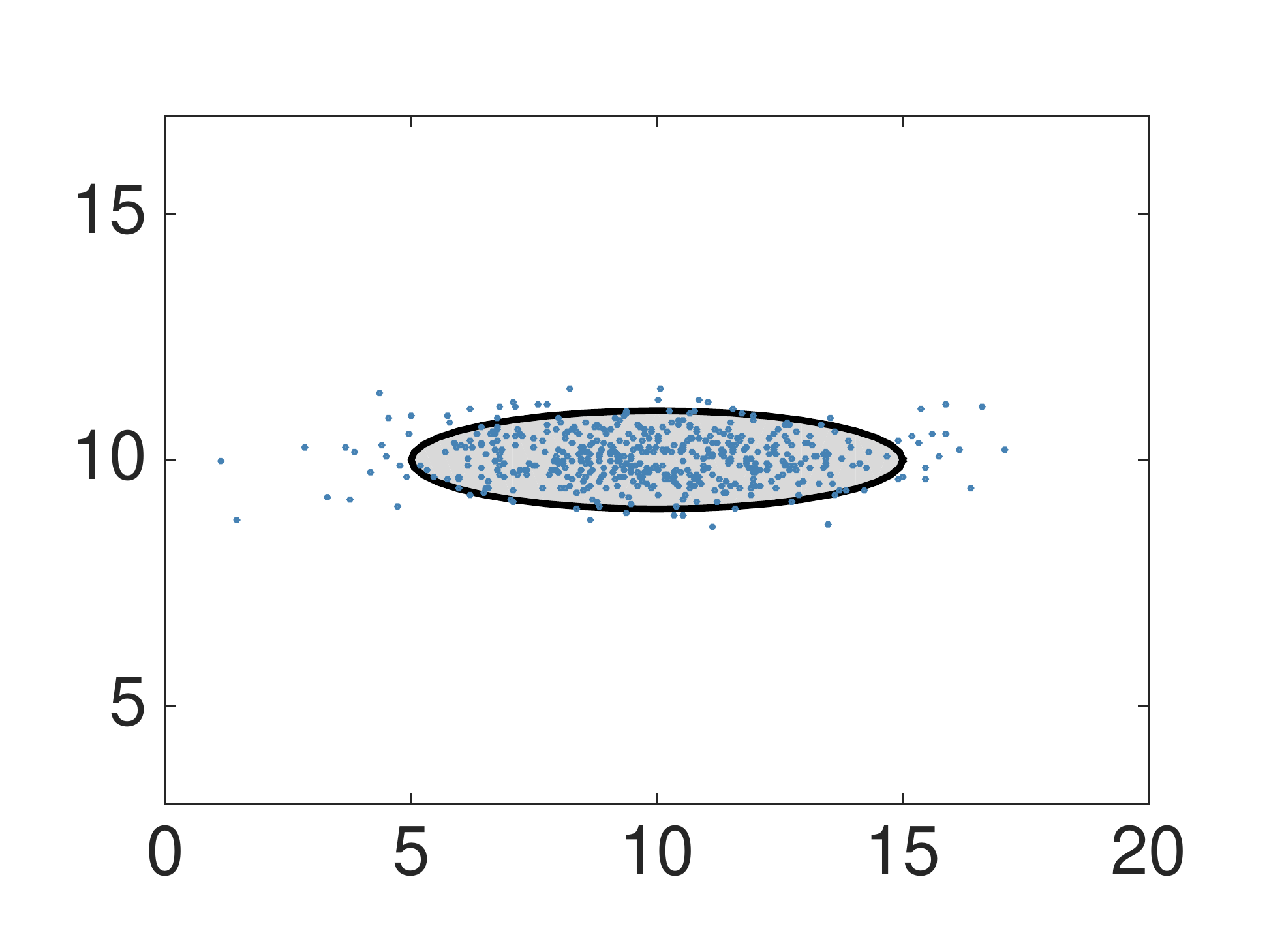}}
	\hfil
	\subfloat[]{\includegraphics[width=0.2\textwidth]{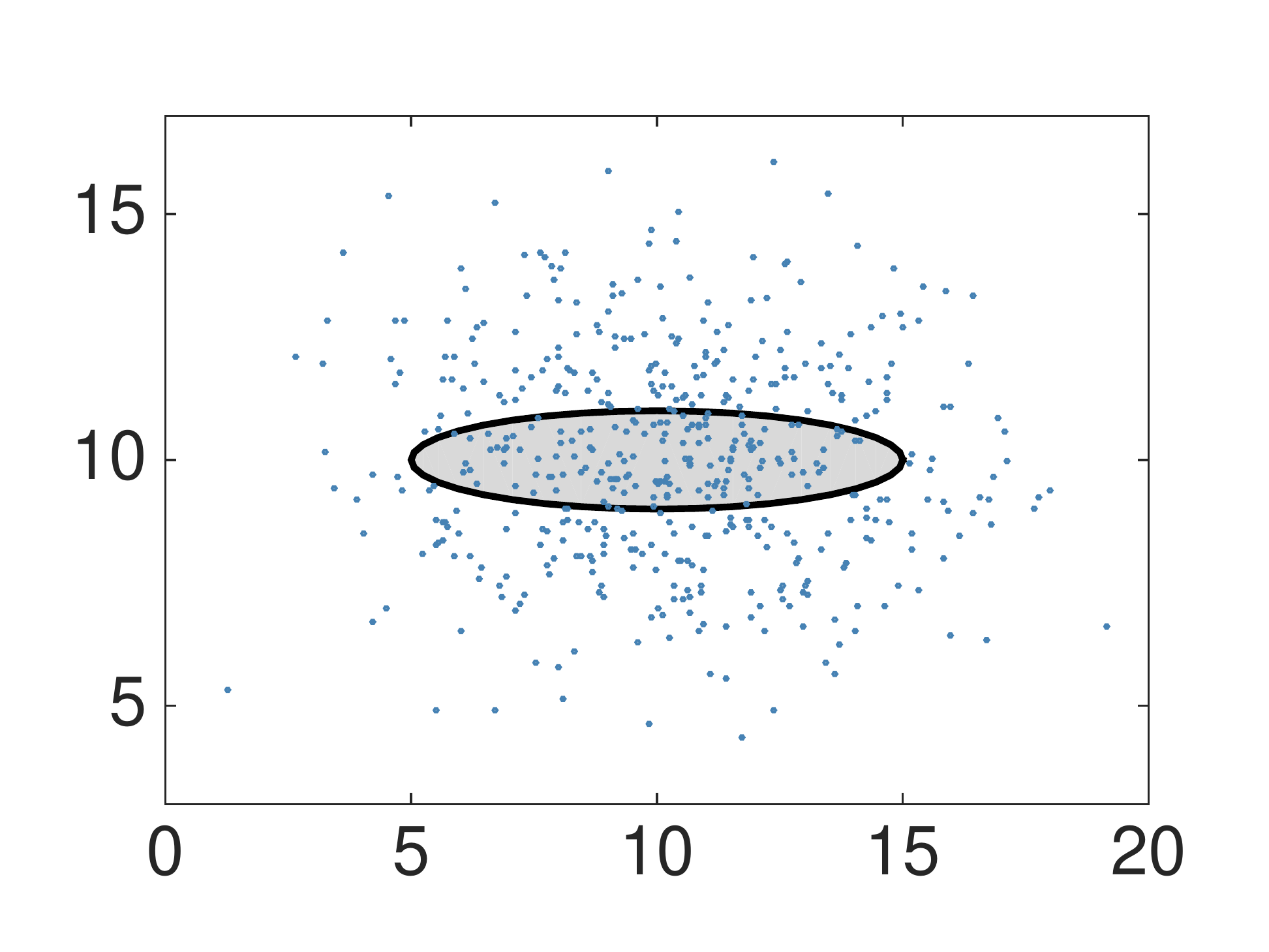}}
	\hfil
	\subfloat[]{\includegraphics[width=0.2\textwidth]{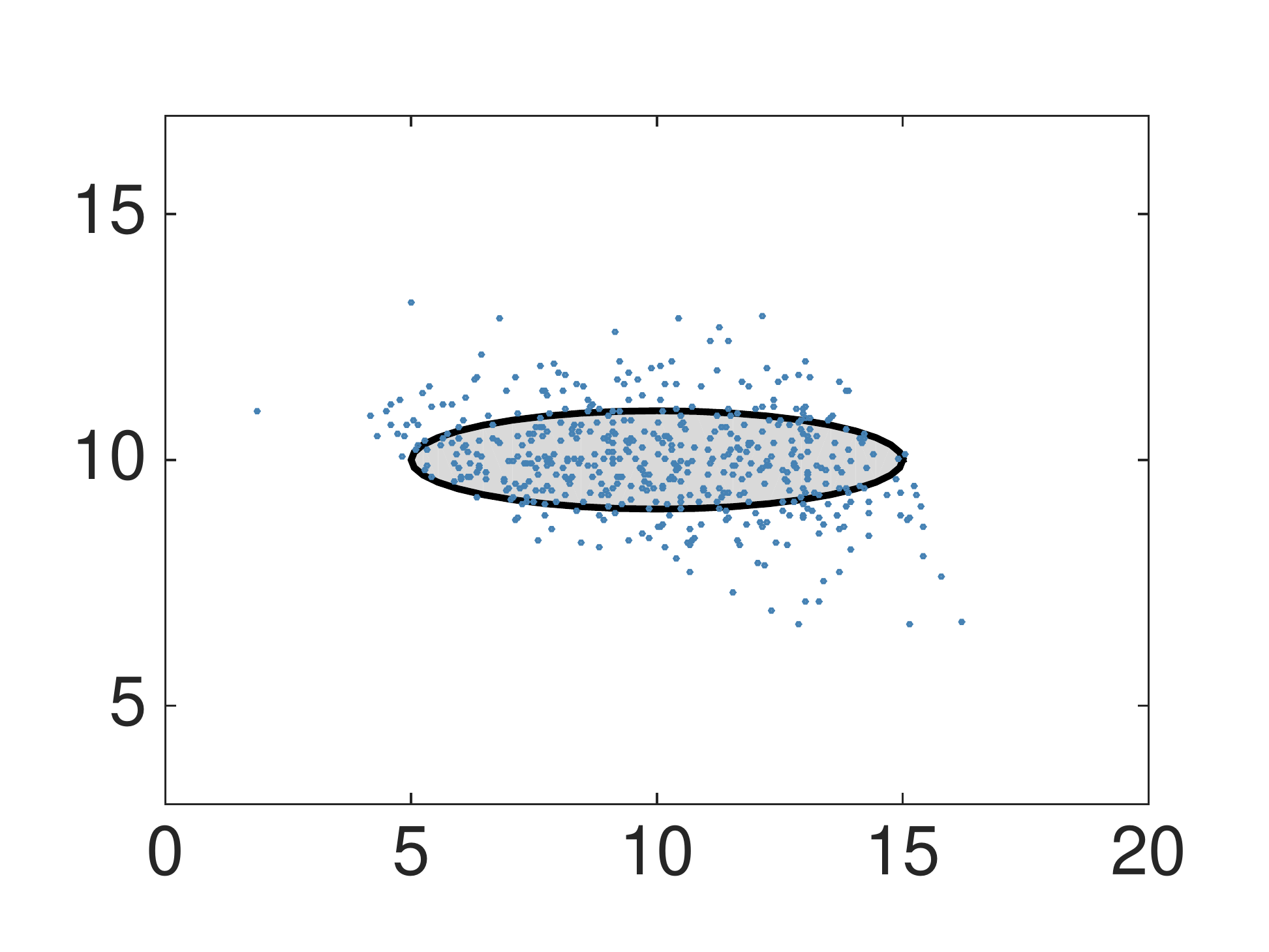}}
	\hfil
	\subfloat[]{\includegraphics[width=0.2\textwidth]{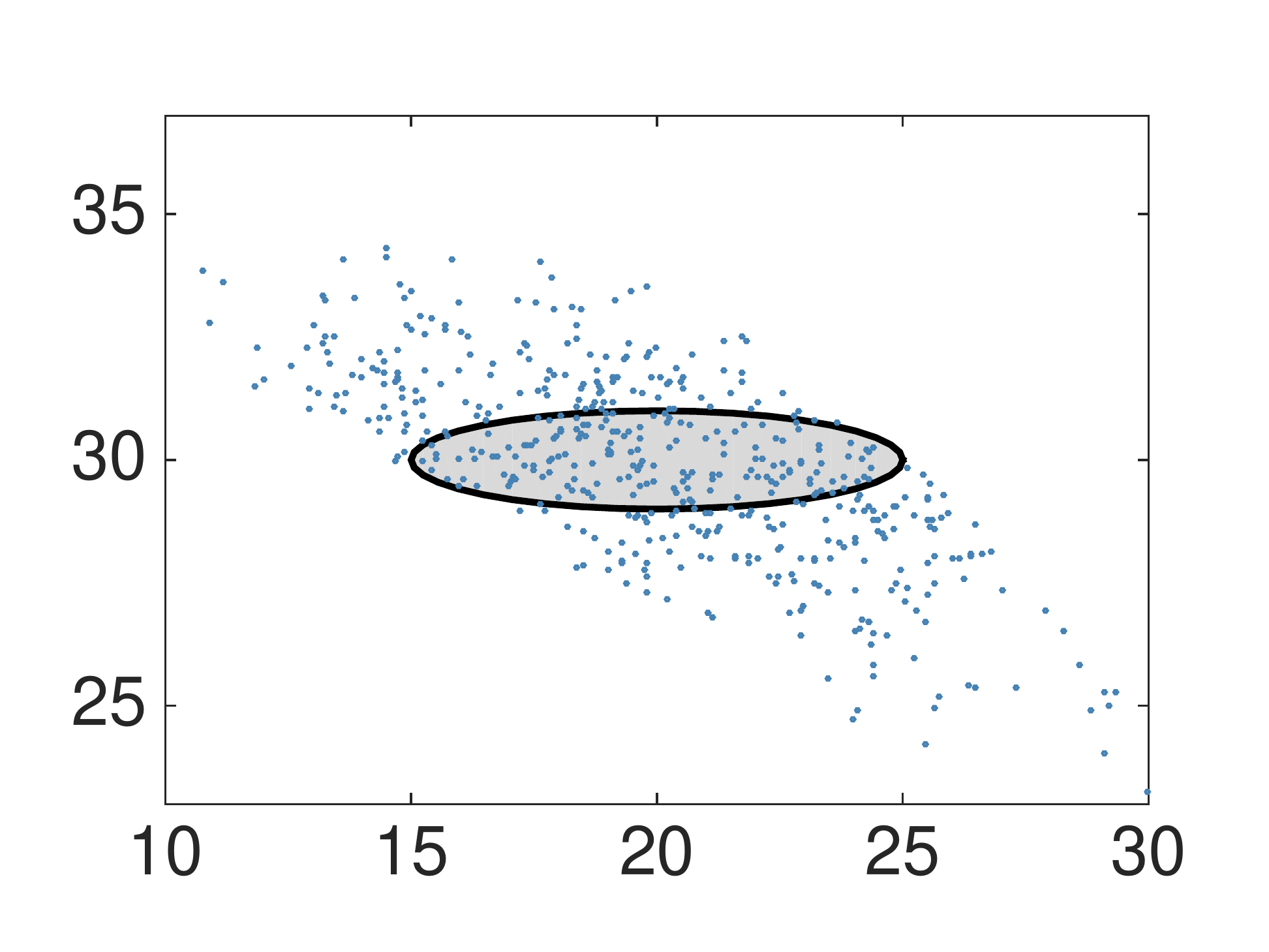}}
	}
	\caption{Illustration of the random matrix measurement model. The sensor is located in the origin. a) Uniform reflection points, no noise. b) Gaussian approximation of uniform distribution. c) Uniform reflection points, Cartesian Gaussian noise. d)--e) Uniform reflection points, polar Gaussian noise. Note how the spread due to noise is larger when the object is further away (e).}
	\label{fig:RMillustration}
\end{figure*}

\subsubsection{Original measurement model}
In the original model \cite{Koch:2008} the measurements are assumed independent, and conditioned on the object state $\sx_k,\ext_k$ the single measurement likelihood---cf. \eqref{eq:PPPmeasLik}, \eqref{eq:BinomMeasLik}---is modelled as Gaussian,
\begin{subequations}
\begin{align} 
p\left(\sz_{k}|\sx_{k},\ext_{k}\right) = & \Npdfbig{\sz_{k}}{ (H_k \otimes \mathbf{I}_{d}) \sx_{k}}{\ext_{k}}. \label{eq:Linear_Gaussian_Measurements}
\end{align}%
where $\otimes$ is Kronecker product, $\mathbf{I}_{d}$ is an identity matrix of the same dimensions as the extent, the noise covariance matrix is the extent matrix, and $(H_k \otimes \mathbf{I}_{d})$ is a measurement model that picks out the Cartesian position from the kinematic vector $\sx_k$.

For Gaussian measurements, the conjugate priors for unknown mean and covariance are the Gaussian and the inverse Wishart distributions, respectively. This motivates the object state distribution \cite{Koch:2008}
\begin{align}
p\left(\sx_{k},\ext_{k}|\setZ^{k}\right) = & p\left(\sx_{k}|\ext_{k},\setZ^{k}\right)p\left(\ext_{k}|\setZ^{k}\right) \\
= & \Npdfbig{\sx_{k}}{m_{k|k}}{P_{k|k}\otimes\ext_{k}} \nonumber \\
& \times \IWishpdf{\ext_{k}}{v_{k|k}}{V_{k|k}}, \label{eq:Koch_GIW_model}
\end{align}%
\label{eq:Koch_random_matrix_model}%
\end{subequations}
where the kinematic vector is Gaussian distributed with mean $m_{k|k}$ and covariance $P_{k|k}\otimes\ext_{k}$, and the extent matrix is inverse Wishart distributed with $v_{k|k}$ degrees of freedom and scale matrix $V_{k|k}$. Owing to the specific form of the conditional Gaussian distribution, where the covariance is the Kronecker product of a matrix $P_{k|k}$ and the extent matrix, non-linear dynamics, such as turn-rate, can not be included in the kinematic vector. In this model the kinematic state $\sx_{k}$ is limited to consist of a spatial state component $\mathbf{r}_{k}$ that represents the center of mass (i.e., the object's position), and derivatives of $\mathbf{r}_{k}$ (typically velocity and acceleration, although higher derivatives are possible) \cite{Koch:2008}. It follows from this that the motion modelling for the kinematic state is linear \cite{Koch:2008}, see further in Section~\ref{sec:RandomMatrixDynamicModelling}.

The measurement update is linear without approximation \cite{Koch:2008}, the details are given in Table~\ref{tab:KochRMupdate}. For the kinematic state a Kalman-filter-like update is performed, and the extent state is updated with two matrices $N$ and $Z$, where the matrix $N$ is proportional to the spread of the centroid measurement $\bar{\sz}$ (mean measurement) around the predicted centroid $(H_k \otimes \mathbf{I}_{d})m$, and the matrix $Z$ is proportional to the sum of the spreads of the measurements around the centroid measurement.

\subsubsection{Improved noise modelling}

An implicit assumption of the original random matrix model \eqref{eq:Koch_random_matrix_model} is that the measurement noise is negligible compared to the extent. In some scenarios this assumption does not hold, for example when marine X-band radar is used \cite{VivoneBGNC:2016_JAIF}. If the measurement noise is not modelled properly the filtering will lead to a biased estimate, see, e.g., \cite{GranstromNBLS:2015_TGARS}.

To alleviate this problem Feldmann \etal \cite{FeldmannF:2008,FeldmannF:2009,FeldmannFK:2011} suggested to use a measurement likelihood that is a convolution of a source distribution and a noise distribution, see \eqref{eq:ReflPointNoiseConvolution}. The noise is modelled as zero mean Gaussian with constant covariance,
\begin{align}
p(\sz_k|\sy_k) = \mathcal{N}(\sz_k;\sy_k,R),
\label{eq:MeasNoise}
\end{align}
and the measurement sources are modelled as uniformly distributed on the object,
\begin{align}
p(\sy_k|\sx_k,\ext_k) = \mathcal{U}(\sy_k; \sx_k, \ext_k).
\label{eq:UniformReflectionPoints}
\end{align}
A uniform distribution is appropriate, e.g., when marine radar is used to track boats and ships, see  \cite{GranstromNBLS:2014,GranstromNBLS:2015_TGARS,VivoneBGW:2016,VivoneBGNC:2015_ConvMeas,VivoneBGNC:2016_JAIF}. The drawback of the uniform distribution is that the convolution \eqref{eq:ReflPointNoiseConvolution} is not analytically tractable.

It is shown in \cite{FeldmannFK:2011} that for an elliptically shaped object the uniform distribution \eqref{eq:UniformReflectionPoints} can be approximated by a Gaussian distribution
\begin{align}
p(\sy_k|\sx_k,\ext_k) = \mathcal{N}(\sy_k;\mathbf{H}_{k}\sx_k,z \ext_k)
\label{eq:GaussianReflectionPoints}
\end{align}
where $z$ is a scaling factor and $\mathbf{H}_{k}$ is a measurement model that picks out the position. A simulation study in \cite{FeldmannFK:2011} showed that $z=1/4$ is a good parameter setting; this result is experimentally verified in \cite{VivoneBGNC:2016_JAIF}. The difference between the uniform distribution \eqref{eq:UniformReflectionPoints} and its Gaussian approximation \eqref{eq:GaussianReflectionPoints} is illustrated in Figure~\ref{fig:RMillustration}, see subfigures a and b.

With the Gaussian noise model \eqref{eq:MeasNoise} and the Gaussian approximation \eqref{eq:GaussianReflectionPoints} the solution to the convolution \eqref{eq:ReflPointNoiseConvolution} is
\begin{align}
p(\sz_k|\sx_k,\ext_k) & =  \Npdfbig{\sz_{k}}{\mathbf{H}_k\sx_{k}}{z\ext_{k}+R}. \label{eq:Feldmann_measurement_model}
\end{align}%
An example with elliptic extent $\ext$ and circular measurement noise covariance $R$ is given in Figure~\ref{fig:RMillustration}, see subfigure c. The inclusion of the constant noise matrix $R$ means that, with a Gaussian inverse Wishart prior of the form \eqref{eq:Koch_random_matrix_model}, the update is no longer analytically tractable. Feldmann \etal \cite{FeldmannF:2008,FeldmannF:2009,FeldmannFK:2011} proposed to approach this by modelling the extended object state with a factorised state density
\begin{subequations}
\begin{align}
p\left(\sx_{k},\ext_{k}|\setZ^{k}\right)  = &  p\left(\sx_{k}|\setZ^{k}\right)p\left(\ext_{k}|\setZ^{k}\right)  \\
= & \Npdfbig{\sx_{k}}{m_{k|k}}{P_{k|k}} \nonumber \\
& \times\IWishpdf{\ext_{k}}{v_{k|k}}{V_{k|k}}. \label{eq:Feldmann_random_matrix_state}
\end{align}%
 \label{eq:Feldmann_random_matrix_model}%
\end{subequations}
Note the assumed independence between the kinematic state $\sx_{k}$ and $\ext_{k}$ in \eqref{eq:Feldmann_random_matrix_state}, an assumption that cannot be fully theoretically justified\footnote{After updating with a set of measurements $\setZ$ the kinematic state $\sx$ and extent state $\ext$ are necessarily dependent.}. 

Despite this theoretical drawback of a factorised density \eqref{eq:Feldmann_random_matrix_model}, there are some practical advantages to using the state distribution \eqref{eq:Feldmann_random_matrix_state}, instead of \eqref{eq:Koch_GIW_model}. The factorised model allows for a more general class of kinematic state vectors $\sx_{k}$, e.g. including non-linear dynamics such as heading and turn-rate, and the Gaussian covariance is no longer intertwined with the extent matrix. Further, the measurement model is better when the size of the extent and the size of the sensor noise are within the same order of magnitude \cite{FeldmannFK:2011}. The assumed independence between $\sx_{k}$ and $\ext_{k}$ is alleviated in practice by the measurement update which provides for the necessary interdependence between kinematics and extent estimation, see \cite{FeldmannFK:2011}. 

\begin{table}
\caption{Random matrix update from \cite{Koch:2008}}
\label{tab:KochRMupdate}
\vspace{-5mm}
\rule[0pt]{\columnwidth}{1pt}
\textbf{Input:} Parameters $m,P,v,V$ of conditional state density \eqref{eq:Koch_random_matrix_model}, measurement model $H$, set of detections $\setW$, $n=|\setW|$

\textbf{Output:} Updated parameters $m_{+},P_{+},v_{+},V_{+}$
\begin{align*}
\begin{array}{rcl}
m_{+} &= & m + (K \otimes \mathbf{I}_{d}) \varepsilon \\
P_{+} &= & P - K S K^{\tp} \\
v_{+} &= & v + n \\
V_{+} &= & V+ N + Z \\
\varepsilon &= &  \bar{\sz} - (H \otimes \mathbf{I}_{d}) m^{}\\
\bar{\sz}  &= &  \frac{1}{n}\sum_{\sz^{i} \in \setW}{\sz^{i}} \\
Z &= & \sum_{\sz_{k}^{i}\in \setW} \left(\sz^{i} - \bar{\sz}\right)\left(\sz^{i} - \bar{\sz}\right)^{\tp} \\
S  &= & H P H^{\tp} + \frac{1}{n}\\
K  &= & P H^{\tp} S^{-1}\\
N^{} &= & S^{-1} \varepsilon \varepsilon^{\tp}
\end{array}
\end{align*}
\rule[0pt]{\columnwidth}{1pt}
\end{table}

With the measurement likelihood \eqref{eq:Feldmann_measurement_model} and the state density in \eqref{eq:Feldmann_random_matrix_model} the updated extent estimate is unbiased, however the measurement update requires approximation. The update presented in \cite{FeldmannFK:2011}, for details see Table~\ref{tab:FeldmannRMupdate}, is based on the assumption that the extent is approximately equal to the predicted estimate,
\begin{align}
	\ext_{k}\approx\hat{\ext}_{k|k-1}=E[\ext_{k}|\setZ^{k-1}], \label{eq:RMpredAssumption}
\end{align}
and on the approximation of non-linear functions of the extent using matrix square roots computed with Cholesky factorisation, $\hat{X} = \hat{X}^{\frac{\tp}{2}} \hat{X}^{\frac{1}{2}}$. After some clever approximations the update of the kinematic state is again a Kalman-filter-like update, and the extent state shape matrix is again updated with two matrices $\hat{N}$ and $\hat{Z}$ proportional to the spreads around the predicted measurement and the centroid. Note that the difference to the original approach, see $N$ and $Z$ in Table~\ref{tab:KochRMupdate} is in the scaling of the two matrices. 

A simulation study in \cite{FeldmannFK:2011} shows that the noisy measurement model \eqref{eq:Feldmann_measurement_model} and the factorised state model \eqref{eq:Feldmann_random_matrix_model} does indeed outperform the original model \eqref{eq:Koch_random_matrix_model} when the measurement noise is non-negligible. A performance analysis of the update in Table~\ref{tab:FeldmannRMupdate} based on the posterior Cram{\'e}r-Rao lower bounds can be found in \cite{SaritasO:2016}.

\begin{table}
\caption{Random matrix update from \cite{FeldmannFK:2011}}
\label{tab:FeldmannRMupdate}
\vspace{-5mm}
\rule[0pt]{\columnwidth}{1pt}
\textbf{Input:} Parameters $m,P,v,V$ of factorised state density \eqref{eq:Feldmann_random_matrix_model}, measurement model $\mathbf{H}$, measurement noise covariance $R$, scaling factor $z$, set of detections $\setW$, $n=|\setW|$

\textbf{Output:} Updated parameters $m_{+},P_{+},v_{+},V_{+}$
\begin{align*}
\begin{array}{rcl}
m_{+} &= & m + K \varepsilon \\
P_{+} &= & P -K S K^{\tp}\\
v_{+} &= & v + n\\
V_{+} &= & V+\hat{N} + \hat{Z} \\
\varepsilon^{} &= &  \bar{\sz} - \mathbf{H} m\\
\bar{\sz}  &= &  \frac{1}{n}\sum_{\sz^{i} \in \setW}{\sz_{}^{i}} \\
Z_{}^{} &= & \sum_{\sz_{k}^{i}\in \setW} \left(\sz_{}^{i} - \bar{\sz}_{}^{}\right)\left(\sz_{}^{i} - \bar{\sz}_{}^{}\right)^{\tp} \\
S^{}  &= & \mathbf{H} P^{} \mathbf{H}^{\tp} + \frac{Y}{n}\\
K^{}  &= & P^{}\mathbf{H}^{\tp} S^{-1}\\
\hat{X} & = & V \left(v-2d-2\right)^{-1} \\
Y &= & z \hat{X} + R\\
\hat{N} &= & \hat{X}^{\frac{1}{2}} S^{-\frac{1}{2}}  \varepsilon \varepsilon^{\tp} S^{-\frac{\tp}{2}} \hat{X}^{\frac{\tp}{2}} \\
\hat{Z} & = & \hat{X}^{\frac{1}{2}} Y^{-\frac{1}{2}}  Z Y^{-\frac{\tp}{2}} \hat{X}^{\frac{\tp}{2}}
\end{array}
\end{align*}
\rule[0pt]{\columnwidth}{1pt}
\end{table}

For the models \eqref{eq:Feldmann_measurement_model} and \eqref{eq:Feldmann_random_matrix_model} two additional updates are presented in \cite{Orguner:2012,ArdeshiriOG:2015}. The update presented in \cite{Orguner:2012} is based on variational Bayesian approximation\footnote{Variational Bayes, or simply variational inference, is a type of approximate inference that builds upon approximating the true distribution with a factorised distribution, i.e, approximation under assumed independence. Thus, variational Bayes is a suitable estimation method for the state model \eqref{eq:Feldmann_random_matrix_state}, since this model already makes the necessary factorisation assumption and approximates the distribution $p\left(\sx_{k},\ext_{k}|\setZ^{k}\right)$ with a factorised distribution $p\left(\sx_{k}|\setZ^{k}\right)p\left(\ext_{k}|\setZ^{k}\right)$. Variational Bayes, and other approximate inference methods, are described further in, e.g., \cite[Ch. 10]{Bishop:2006}.}, where the unknown measurement sources $\sy$, cf. \eqref{eq:ReflPointNoiseConvolution}, are estimated as so called hidden variables. The update is iterative, and can be run either for a fixed number of iterations, or until some convergence criterion is met. The details are given in Table~\ref{tab:OrgunerRMupdate}. 

A simulation study in \cite{Orguner:2012} shows that the variational update has smaller estimation error than the update based on Cholesky factorisation (Table~\ref{tab:FeldmannRMupdate}), at the price of higher computational cost. It is reported that the update on average converges in $5$ iterations, however, to be on the safe side $20$ iterations were performed in each update in the simulation study \cite{Orguner:2012}.

An update based on linearisation of the natural logarithm of the measurement likelihood \eqref{eq:Feldmann_measurement_model} is presented in \cite{ArdeshiriOG:2015}, details are given in Table~\ref{tab:TohidRMupdate}. A simulation study in \cite{ArdeshiriOG:2015} shows that the log-linearised update gives results that almost match the variational update, at a lower computational cost.

\begin{table}
\caption{Random matrix update from \cite{Orguner:2012}}
\label{tab:OrgunerRMupdate}
\vspace{-5mm}
\rule[0pt]{\columnwidth}{1pt}
\textbf{Input:} Parameters $m,P,v,V$ of factorised state density \eqref{eq:Feldmann_random_matrix_model}, measurement model $\mathbf{H}$, measurement noise covariance $R$, scaling factor $z$, set of detections $\setW$, $n=|\setW|$

\textbf{Output:} Updated parameters $m_{+},P_{+},v_{+},V_{+}$

Initialize
\begin{align*}
\begin{array}{rcl}
\sy^{i,(0)} & = & \sz^{i} \\
\Sigma^{(0)} & = & z V(v-2d-2)^{-1} \\
m_{+}^{(0)} &= & m \\
P_{+}^{(0)} &= & P \\
v_{+} &= & v + n \\
V_{+}^{(0)} &= & V \\
\end{array}
\end{align*}

Iterate until convergence
\begin{align*}
\begin{array}{rcl}
\sy^{i,(t+1)} & = & \Sigma^{(t+1)} \left( v_{+} (z V_{+}^{(t)})^{-1} \mathbf{H} m_{+}^{(t)} + R^{-1}  \sz^{i}\right) \\
\Sigma^{(t+1)} & = & \left( v_{+} (z V_{+}^{(t)})^{-1} + R^{-1} \right)^{-1} \\
m_{+}^{(t+1)} &= & P_{+}^{(t+1)} \left( P^{-1}m + n \mathbf{H}^{\tp} v_{+} (z V_{+}^{(t)})^{-1} \frac{1}{n}\sum_{i}\sy^{i,(t)}\right) \\
P_{+}^{(t+1)} &= & \left( P^{-1} + n \mathbf{H}^{\tp} v_{+} (z V_{+}^{(t)})^{-1} \mathbf{H} \right)^{-1} \\
V_{+}^{(t+1)} &= & V + \frac{1}{z} \sum_{i} (\sy^{i,(t)}-\mathbf{H}m_{+}^{(t)})(\sy^{i,(t)}-\mathbf{H}m_{+}^{(t)})^{\tp} \\
& & + \frac{n}{z}\mathbf{H}P^{(t)}\mathbf{H}^{\tp} + \frac{n}{z}\Sigma^{(t)} \\
\end{array}
\end{align*}

Output ($T$ is the final iteration)
\begin{align*}
\begin{array}{rcl}
m_{+} &= & m_{+}^{(T)}  \\
P_{+} &= & P_{+}^{(T)} \\
v_{+} &= & v + n \\
V_{+} &= & V_{+}^{(T)} \\
\end{array}
\end{align*}
\rule[0pt]{\columnwidth}{1pt}
\end{table}

To improve the measurement modelling for the original conditional state model \eqref{eq:Koch_GIW_model} the following measurement likelihood was proposed in \cite{LanRL:2012,LanRL:NewModelApproachAccepted},
\begin{align}
p\left(\sz_{k}|\sx_{k},\ext_{k}\right) = & \Npdfbig{\sz_{k}}{(H_k \otimes I) \sx_{k}}{B_{k}\ext_{k}B_{k}^{\tp}} \label{eq:LanRL_measurement_model}
\end{align}
where $B_{k}$ is a known parameter matrix. The update, see details in Table~\ref{tab:LanRMupdate}, builds upon the approximation \cite[Eq. 28]{LanRL:2012}
\begin{align}
	B_k \ext_k B_k^{\tp} \approx \gamma_{k} \ext_{k}
\end{align}
where $\gamma_{k}$ is a scalar that is given by setting the determinants of both sides equal \cite[Eq. 29]{LanRL:2012}
\begin{align}
	\det( B_k \ext_k B_k^{\tp} ) = \det( \gamma_{k} \ext_{k} ) \Rightarrow \gamma_{k} = \det( B_{k} )^{2/d}
\end{align}
Under the assumption that the extent is approximately equal to the predicted estimate \eqref{eq:RMpredAssumption} the measurement model \eqref{eq:LanRL_measurement_model} can model additive Gaussian noise approximately by setting
\begin{align}
	B_{k} = (z\hat{\ext}_{k|k-1}+R)^{1/2}\hat{\ext}_{k|k-1}^{-1/2}.
\end{align}
Note that similarly to the update presented in \cite{FeldmannFK:2011}, this requires matrix square roots. In addition to modelling noise, the matrix $B_{k}$ can be used to model distortion of the observed extent \cite{LanRL:2012}.

\subsubsection{Non-linear measurements}

Both the original measurement likelihood \eqref{eq:Linear_Gaussian_Measurements} and the noise adapted measurement likelihoods \eqref{eq:Feldmann_measurement_model} and \eqref{eq:LanRL_measurement_model} are linear with respect to the kinematic state $\sx_k$, and the noise covariance in \eqref{eq:Feldmann_measurement_model} and \eqref{eq:LanRL_measurement_model} is constant. However, when real-world data is used the measurement model is often non-linear, e.g., a radar measures range and azimuth to the object's position instead of measuring the position directly as in \eqref{eq:Linear_Gaussian_Measurements} and \eqref{eq:Feldmann_measurement_model}. Further, due to the polar noise the noise covariance in Cartesian coordinates is not constant, but increases with increasing sensor-to-object distance. 

\begin{table}
\caption{Random matrix update from \cite{ArdeshiriOG:2015}}
\label{tab:TohidRMupdate}
\vspace{-5mm}
\rule[0pt]{\columnwidth}{1pt}
\textbf{Input:} Parameters $m,P,v,V$ of factorised state density \eqref{eq:Feldmann_random_matrix_model}, measurement model $\mathbf{H}$, measurement noise covariance $R$, scaling factor $z$, set of detections $\setW$, $n=|\setW|$

\textbf{Output:} Updated parameters $m_{+},P_{+},v_{+},V_{+}$
\begin{align*}
\begin{array}{rcl}
m_{+} &= & m + K \varepsilon \\
P_{+} &= & P -K S K^{\tp}\\
v_{+} &= & v + n\\
V_{+} &= & V + M \\
\varepsilon^{} &= &  \bar{\sz} - \mathbf{H} m\\
\bar{\sz}  &= &  \frac{1}{n}\sum_{\sz^{i} \in \setW}{\sz_{}^{i}} \\
Z_{}^{} &= & \sum_{\sz_{k}^{i}\in \setW} \left(\sz_{}^{i} - \bar{\sz}_{}^{}\right)\left(\sz_{}^{i} - \bar{\sz}_{}^{}\right)^{\tp} \\
S^{}  &= & \mathbf{H} P^{} \mathbf{H}^{\tp} + \frac{z \hat{X} + R}{n}\\
K^{}  &= & P^{}\mathbf{H}^{\tp} S^{-1}\\
\hat{X} & = & V \left(v-2d-2\right)^{-1} \\
C &= & \mathbf{H} P \mathbf{H}^{\tp}  + z \hat{X} + R\\
M & = & n \hat{X} + n z \hat{X} C^{-1}(\frac{Z}{n} + \varepsilon \varepsilon^{\tp} - C) C^{-1} \hat{X}
\end{array}
\end{align*}
\rule[0pt]{\columnwidth}{1pt}
\end{table}

\begin{table}
\caption{Random matrix update from \cite{LanRL:NewModelApproachAccepted}}
\label{tab:LanRMupdate}
\vspace{-5mm}
\rule[0pt]{\columnwidth}{1pt}
\textbf{Input:} Parameters $m,P,v,V$ of conditional state density \eqref{eq:Koch_random_matrix_model}, measurement model $H$, parameter matrix $B$, set of detections $\setW$, $n=|\setW|$

\textbf{Output:} Updated parameters $m_{+},P_{+},v_{+},V_{+}$
\begin{align*}
\begin{array}{rcl}
m_{+} &= & m + (K \otimes \mathbf{I}_{d}) \varepsilon \\
P_{+} &= & P - K S K^{\tp} \\
v_{+} &= & v + n\\
V_{+} &= & V+ N + \hat{Z} \\
\varepsilon &= &  \bar{\sz} - (H \otimes \mathbf{I}_{d}) m^{}\\
\bar{\sz}  &= &  \frac{1}{n}\sum_{\sz^{i} \in \setW}{\sz^{i}} \\
Z &= & \sum_{\sz_{k}^{i}\in \setW} \left(\sz^{i} - \bar{\sz}\right)\left(\sz^{i} - \bar{\sz}\right)^{\tp} \\
S  &= & H P H^{\tp} + \frac{\det(B)^{{2}/{d}}}{n}\\
K  &= & P H^{\tp} S^{-1}\\
N^{} &= & S^{-1} \varepsilon \varepsilon^{\tp} \\
\hat{Z} & = & B^{-1} Z B^{-\tp}
\end{array}
\end{align*}
\rule[0pt]{\columnwidth}{1pt}
\end{table}

In \cite{VivoneBGW:2016,VivoneBGNC:2015_ConvMeas,VivoneBGNC:2016_JAIF} non-linear radar measurements are handled by performing a polar to Cartesian conversion in a pre-processing step, and by modelling the the noise covariance $R(\sy)$ as a function of the reflection point. The measurement noise model \eqref{eq:MeasNoise} is modified to
\begin{align}
p(\sz_k|\sy_k,\sx_k,\ext_k) = \mathcal{N}(\sz_k;\sy_k,R(\sy_k)). \label{eq:NonConstantGaussianNoise}
\end{align}
After conversion to Cartesian coordinates the spread of the measurements due to noise is larger the further the object is from the sensor, see Figure~\ref{fig:RMillustration}, subfigures d and e.  With the Gaussian noise model \eqref{eq:NonConstantGaussianNoise} and the Gaussian approximation \eqref{eq:GaussianReflectionPoints}, the convolution of the two (cf. \eqref{eq:ReflPointNoiseConvolution})
\begin{align}
p&(\sz_k|\sx_k,\ext_k) \nonumber \\
&= \int\mathcal{N}(\sz_k;\sy_k,R(\sy_k))\mathcal{N}(\sy_k;\mathbf{H}\sx_k,z \ext_k)\diff\sy_k,
\label{eq:NoisyReflectionMeasurements}
\end{align}
does not have an analytical solution. In \cite{VivoneBGW:2016,VivoneBGNC:2015_ConvMeas,VivoneBGNC:2016_JAIF} this is handled by approximating the noise covariance as
\begin{align}
	R(\sy) \approx & R(\hat{\sy}_k), \\
	\hat{\sy}_{k} = & H_k \hat{\sx}_{k|k-1}=H_k E[\sx_{k}|\setZ^{k-1}].
\end{align}
This allows any of the updates presented in \cite{FeldmannFK:2011,Orguner:2012,ArdeshiriOG:2015} to be used (see Tables~\ref{tab:FeldmannRMupdate}, \ref{tab:OrgunerRMupdate} and \ref{tab:TohidRMupdate}).

Non-linear range and azimuth measurement for the conditional state model \eqref{eq:Koch_GIW_model} and the measurement likelihood \eqref{eq:LanRL_measurement_model} are modelled in \cite{LanL:2016}, where linearisation and a Variational Bayes scheme are used to handle the non-linearities in the update. Radar doppler rate is integrated into the measurement modelling in \cite{SchusterRW:2015}.

\subsubsection{Dynamic modelling}
\label{sec:RandomMatrixDynamicModelling}
In the original random matrix model \cite{Koch:2008} the transition density is modelled as
\begin{subequations}
\begin{align}
p & \left(\sx_{k+1},\ext_{k+1}|\sx_{k},\ext_{k}\right)  \nonumber \\
\approx & p\left(\sx_{k+1}|\ext_{k+1},\sx_{k}\right)p\left(\ext_{k+1}|\ext_{k}\right), \\
= & \Npdfbig{\sx_{k+1}}{(F_k\otimes \mathbf{I}_{d})\sx_{k}}{D_k \otimes \ext_{k+1}} \nonumber \\
& \times \Wishpdf{\ext_{k+1}}{n_{k}}{\ext_{k}/n_{k}} \label{eq:Koch_transition_densityGIW}
\end{align}%
\label{eq:Koch_transition_density}%
\end{subequations}
and in \cite{FeldmannFK:2011} a slightly different transition density was proposed,
\begin{subequations}
\begin{align}
p & \left(\sx_{k+1},\ext_{k+1}|\sx_{k},\ext_{k}\right) \nonumber \\
 \approx & p\left(\sx_{k+1}|\sx_{k}\right)p\left(\ext_{k+1}|\ext_{k}\right). \\
= & \Npdfbig{\sx_{k+1}}{\mathbf{F}_k \sx_{k}}{Q_k} \nonumber \\
& \times \Wishpdf{\ext_{k+1}}{n_{k}}{\ext_{k}/n_{k}}
\end{align}%
\label{eq:Feldmann_transition_density}%
\end{subequations}
In both cases we have a linear Gaussian transition density for the kinematic vector, and for the extent a Wishart transition density where the parameter $n_k>0$ governs the noise level of the prediction: the smaller $n_k$ is, the higher the process noise. 

The predicted parameters of the kinematic state are simple to compute. For the extent state, rather than solving the Chapman-Kolmogorov equation, a simple heuristic is used in which the expected value is kept constant and the variance is increased \cite{Koch:2008}. This corresponds to exponential forgetting for the extent state, see \cite{GranstromO:2014} for additional discussion. The predicted parameters are given in Table~\ref{tab:KochRMprediction} and Table~\ref{tab:FeldmannRMprediction}. 

This model for the extent's time evolution is sufficient when the object manoeuvres are sufficiently slow. In practice, this means that the object turns slowly enough for the rotation of the extent to be very small from one time step to another. The kinematics transition density $p\left(\sx_{k+1}|\sx_{k}\right)$ in \eqref{eq:Feldmann_transition_density} is assumed independent of the extent. This neglects factors such as wind resistance, which can be modelled as a function of the extent $\ext_{k}$, however the assumption is necessary to retain the functional form \eqref{eq:Feldmann_random_matrix_state} in a Bayesian recursion.

An alternative to the heuristic extent predictions from \cite{Koch:2008,FeldmannFK:2011} is to analytically solve the Chapman-Kolmogorov equation \eqref{eq:ChapmanKolmogorovEquation} for a Wishart transition density, and approximate the resulting density with an inverse Wishart density. Different approaches to this is discussed in, e.g., \cite{Koch:2008,LianHLYZ:2010,LanRL:2012,LanRL:NewModelApproachAccepted,GranstromO:2014}.

\begin{table}
\caption{Random matrix prediction from \cite{Koch:2008}}
\label{tab:KochRMprediction}
\vspace{-5mm}
\rule[0pt]{\columnwidth}{1pt}
\textbf{Input:} Parameters $m,P,v,V$ of conditional state density \eqref{eq:Koch_random_matrix_model}, motion model $F$, motion noise covariance $D$, sampling time $T_s$, temporal decay constant $\tau$

\textbf{Output:} Predicted parameters $m_{+},P_{+},v_{+},V_{+}$
\begin{align*}
\begin{array}{rcl}
m_{+} &= & (F \otimes \mathbf{I}_{d}) m \\
P_{+} &= & FPF^{\tp} + D \\
v_{+} &= & e^{-T_{s}/\tau} v \\
V_{+} &= & \frac{v_{+} -2d -2}{v-2d-2} V
\end{array}
\end{align*}
\rule[0pt]{\columnwidth}{1pt}
\end{table}

\begin{table}
\caption{Random matrix prediction from \cite{FeldmannFK:2011}}
\label{tab:FeldmannRMprediction}
\vspace{-5mm}
\rule[0pt]{\columnwidth}{1pt}
\textbf{Input:} Parameters $m,P,v,V$ of factorised state density \eqref{eq:Feldmann_random_matrix_model}, motion model $\mathbf{F}$, motion noise covariance $Q$, sampling time $T_s$, temporal decay constant $\tau$

\textbf{Output:} Predicted parameters $m_{+},P_{+},v_{+},V_{+}$
\begin{align*}
\begin{array}{rcl}
m_{+} &= & \mathbf{F} m \\
P_{+} &= & \mathbf{F}P\mathbf{F}^{\tp} + Q \\
v_{+} &= & 2d+2 + e^{-T_{s}/\tau} (v-2d-2) \\
V_{+} &= & \frac{v_{+} -2d -2}{v-2d-2} V
\end{array}
\end{align*}
\rule[0pt]{\columnwidth}{1pt}
\end{table}

In \cite{LanRL:2012,LanRL:NewModelApproachAccepted} the following transition density is used, where transformations of the extent are allowed via known parameter matrices $A_{k}$,
\begin{subequations}
\begin{align}
p & \left(\sx_{k+1},\ext_{k+1}|\sx_{k},\ext_{k}\right)  \nonumber \\
\approx & p\left(\sx_{k+1}|\ext_{k+1},\sx_{k}\right)p\left(\ext_{k+1}|\ext_{k}\right), \\
= & \Npdfbig{\sx_{k+1}}{(F_k\otimes \mathbf{I}_{d})\sx_{k}}{D_k \otimes \ext_{k+1}} \nonumber \\
& \times \Wishpdf{\ext_{k+1}}{n_{k}}{A_{k}\ext_{k}A_{k}^{\tp}} 
\end{align}%
\label{eq:LanRongLi_prediction}%
\end{subequations}
The solution to the Chapman-Kolmogorov equation \eqref{eq:ChapmanKolmogorovEquation} is not Gaussian inverse Wishart of the form \eqref{eq:Koch_random_matrix_model}, however, using moment matching it can be approximated as such. The predicted parameters are given in Table~\ref{tab:LanRMprediction}. The parameter matrices $A_{k}$ correspond to, \egp, rotation matrices. Rotation matrices are useful for a turning target, because the extent rotates as the target turns. By using the prediction in Table~\ref{tab:LanRMprediction} with three motion models, with different matrices $A_k$ corresponding to \textit{i)} no rotation, \textit{ii)} clockwise rotation and \textit{iii)} counter-clockwise rotation, the target motion can be predicted better compared to using the prediction in Table~\ref{tab:KochRMprediction}, leading to improved estimation, see \cite{LanRL:2012}.

The extent transition density $p\left(\ext_{k+1}|\ext_{k}\right)$ in \eqref{eq:Koch_transition_density}, \eqref{eq:Feldmann_transition_density}, and \eqref{eq:LanRongLi_prediction}, assumes independence of the prior kinematic state $\sx_{k}$. The extent of an object going through a turning manoeuvre will typically rotate during the turn, because the extent is aligned with the object's heading. This implies that the extent transition density should be dependent on the turn-rate, i.e., it should be dependent on the kinematic state $\sx_k$.

The inverse Wishart transition density is generalized in \cite{GranstromO:2014,GranstromO:2011} to allow for transformation matrices $M(\sx_{k})$ that are functions of the kinematic state, which means that the rotation angle can be coupled to, e.g., the turn-rate, and estimated online. The following transition density is used with the factorised state density \eqref{eq:Feldmann_random_matrix_model},
\begin{subequations}
\begin{align}
p & \left(\sx_{k+1},\ext_{k+1}|\sx_{k},\ext_{k}\right) \nonumber \\
 \approx & p\left(\sx_{k+1}|\sx_{k}\right)p\left(\ext_{k+1}|\sx_{k},\ext_{k}\right). \\
= & \Npdfbig{\sx_{k+1}}{\mathbf{f}_k( \sx_{k})}{Q_k} \nonumber \\
& \times \Wishpdf{\ext_{k+1}}{n_{k}}{\frac{M({\sx_{k}})\ext_{k}M({\sx_{k}})^{\tp}}{n_{k}}}
\end{align}%
\label{eq:GranstromOrguner_prediction}%
\end{subequations}
Note that a non-linear motion model $\mathbf{f}(\cdot)$ is used. 

\begin{table}
\caption{Random matrix prediction from \cite{LanRL:NewModelApproachAccepted}}
\label{tab:LanRMprediction}
\vspace{-5mm}
\rule[0pt]{\columnwidth}{1pt}
\textbf{Input:} Parameters $m,P,v,V$ of conditional state density \eqref{eq:Koch_random_matrix_model}, motion model $F$, motion noise covariance $D$, motion noise degrees of freedom $n$, parameter matrix $A$

\textbf{Output:} Predicted parameters $m_{+},P_{+},v_{+},V_{+}$
\begin{align*}
\begin{array}{rcl}
m_{+} &= & (F \otimes \mathbf{I}_{d}) m \\
P_{+} &= & FPF^{\tp} + D \\
v_{+} &= & \frac{2n(\lambda-1)(\lambda-2)}{\lambda(\lambda+n)} + 2d+4\\
V_{+} &= & \frac{n}{\lambda-1}(v - 2d-2)AVA^{\tp}\\
\lambda &= & v -2d-2
\end{array}
\end{align*}
\rule[0pt]{\columnwidth}{1pt}
\end{table}

Similarly to \eqref{eq:LanRongLi_prediction}, the solution to the Chapman-Kolmogorov equation is not of the desired form, i.e, not a factorised Gaussian inverse Wishart \eqref{eq:Feldmann_random_matrix_model}. By minimising the Kullback-Leibler divergence, the predicted density can be approximated as Gaussian inverse Wishart of the form \eqref{eq:Feldmann_random_matrix_model}. The parameters of the prediction are given in Table~\ref{tab:GranstromRMprediction}. The proof that the solution $s$ to the non-linear equation is unique is given in \cite{GranstromO:2011}. %The second order derivatives in the third order Taylor expansion approximation of the expected values are

%\begin{align}
%	& \frac{\diff^{2} \log(\det(M_{\sx}VM_{\sx}^{\tp}))}{\diff \sx_{i} \diff \sx_{j}} \nonumber \\
%	= & \tr\left( - \left(M_{\sx}VM_{\sx}^{\tp}\right)^{-1} \frac{\diff M_{\sx}VM_{\sx}^{\tp}}{\diff \sx_{i}} \left(M_{\sx}VM_{\sx}^{\tp}\right)^{-1}\frac{\diff M_{\sx}VM_{\sx}^{\tp}}{\diff \sx_{j}} + \left(M_{\sx}VM_{\sx}^{\tp}\right)^{-1} \frac{\diff^{2} M_{\sx}VM_{\sx}^{\tp}}{\diff\sx_{i} \diff\sx_{j}}  \right)
%\end{align}

%If, e.g., the transformation matrix is a rotation matrix
%\begin{align}
%	M(\sx) = M(\omega) = \begin{bmatrix} \cos(\omega T_s) & -\sin(\omega T_s) \\ \sin(\omega T_s) & \cos(\omega T_s) \end{bmatrix}
%\end{align}
%where $\omega$ is the turn rate of the target, and $T_s$ is the sample time, then we have
%\begin{align}
%	C_1 = & \log(\det(M(\hat{\omega}) V M(\hat{\omega})^{\tp})) + \frac{1}{2} P_{\omega,\omega} W''(\hat{\omega}) \\
%	C_2 = & W(\hat{\omega}) + \frac{1}{2} P_{\omega,\omega}W''(\hat{\omega})
%\end{align}

A comparison of the predictions resulting from the transition densities \eqref{eq:Feldmann_transition_density}, \eqref{eq:LanRongLi_prediction} and \eqref{eq:GranstromOrguner_prediction}, i.e., the predictions in Tables~\ref{tab:FeldmannRMprediction}, \ref{tab:LanRMprediction}, \ref{tab:GranstromRMprediction}, is presented in \cite{GranstromO:2014}. For a target that moves according to a constant turn motion model, see, e.g., \cite[Sec. V.A]{RongLiJ:2003}, the prediction in Table~\ref{tab:GranstromRMprediction} is shown to give lowest filtering and prediction errors when the true turn-rate is unknown. If the true turn-rate is assumed to be known, the two predictions in Tables~\ref{tab:LanRMprediction} and \ref{tab:GranstromRMprediction} perform similarly. 
Average cycle times for Matlab implementations are reported for the prediction in Table~\ref{tab:GranstromRMprediction} and the prediction in Table~\ref{tab:LanRMprediction}; the prediction in Table~\ref{tab:GranstromRMprediction} is shown to be about three times faster than the prediction in Table~\ref{tab:LanRMprediction} with three modes. Note that any prediction or update can be speeded up, e.g., by parallelising computations or implementing in a fast low level language, like C++. Because of this it is important to interpret any differences in average cycle time with care.

When there are many measurements per object the measurement update will dominate the prediction and compensate for dynamic motion modelling errors. However, when multiple objects are located next to each other the prediction is important even in scenarios with many measurements per object, and accurate motion modelling can be crucial for estimation performance, see \cite{GranstromO:2014,GranstromWBS:2015_MRMUK,Granstrom:MultiEllipseArxiv}.

\subsubsection{Further extensions of the random matrix model}
Multiple extended object tracking is overviewed in Section~\ref{sec:MultipleExtendedTargetTracking}, here we briefly mention some \mtt algorithms where the random matrix model has been used. In \cite{WienekeK:2010,WienekeD:2011,WienekeK:2012} it is used in the Probabilistic Multi-Hypothesis Tracking (\textsc{pmht}) framework \cite{StreitL:1993} to track persons in video data. The random matrix model has also been used in several \rfs-type filters for multiple extended object tracking in clutter \cite{GranstromO:2012a,LundquistGO:2013,BeardRGVVS:2015journal,GranstromFS:2016fusion}. \textsc{jpda}-type \mtt algorithms are presented in \cite{SchusterRW:JAIF,SchusterRW:2015,VivoneB:2016}. Multi object tracking requires the predicted likelihood
\begin{align}
	p(\setZ) = \iint p(\setZ | \sx,\ext) p(\sx, \ext) \diff \sx \diff \ext
\end{align}
In \cite[Appendix A]{GranstromO:2012a} it is shown that for the original model \cite{Koch:2008} the predicted likelihood is proportional to a generalized matrix variate beta type 2 distribution. In \mtt algorithms it is necessary to maintain several object hypotheses due to the many involved uncertainties. When the random matrix model is used the number of hypotheses can be reduced using the merging algorithm presented in \cite{GranstromO:2012e}.

\begin{table}
\caption{Random matrix prediction from \cite{GranstromO:2014}}
\label{tab:GranstromRMprediction}
\vspace{-5mm}
\rule[0pt]{\columnwidth}{1pt}
\textbf{Input:} Parameters $m,P,v,V$ of factorised state density \eqref{eq:Feldmann_random_matrix_model}, motion model $\mathbf{f}(\cdot)$, motion noise covariance $Q$, motion noise degrees of freedom $n$, matrix transformation function $M(\cdot)$

\textbf{Output:} Predicted parameters $m_{+},P_{+},v_{+},V_{+}$
\begin{align*}
\begin{array}{rcl}
m_{+} &= & \mathbf{f}( m ) \\
P_{+} &= & \mathbf{F}P\mathbf{F}^{\tp} + Q \\
v_{+} & = & (d+1)\left(2 + \frac{(s-d-1)(n-d-1)(v-2d-2)}{sn(v-d-1) - (s-d-1)(n-d-1)(v-2d-2)} \right) \\
V_{+} & = & \frac{v_{+}-d-1}{v-d-1} \frac{s-d-1}{s} \frac{n-d-1}{n} C_{2} \\
\mathbf{F} & = & \left. \nabla_{\sx} \mathbf{f}(\sx) \right |_{\sx=m} \\
C_{1} & = & E\left[ \log(\det(M(\sx) V M(\sx)^{\tp})) \right] \\
C_{2} & = & E\left[ M(\sx) V M(\sx)^{\tp} \right]
\end{array}
\end{align*}
where $s$ is the unique solution to $h(s)=0$, 
\begin{align*}
	h(s) = d \log\left(\frac{s}{2}\right) - \sum_{i=1}^{d} \psi_{0}\left(\frac{s-i+1}{2}\right) + C_{1} - \log(\det(C_{2}))
\end{align*}
and $\psi_{k}(\cdot)$ is the poly-gamma function of order $k$. A solution to  $h(s)=0$ can be found using numerical root-finding. The second order Halley's iteration is
\begin{align*}
\begin{array}{rcl}
	s^{(t+1)} & = & s^{(t)} - \frac{2 h(s^{(t)}) h'(s^{(t)})}{2 (h'(s^{(t)}))^{2} - h(s^{(t)}) h''(s^{(t)})}
\end{array}
\end{align*}
where the first and second order differentiations of $h(s)$ w.r.t. $s$ are
\begin{align*}
\begin{array}{rcl}
	h'(s) & = & \frac{d}{s} - \frac{1}{2} \sum_{i=1}^{d} \psi_{1}\left(\frac{s-i+1}{2}\right) \\
	h''(s) & = & -\frac{d}{s^2} - \frac{1}{4} \sum_{i=1}^{d} \psi_{2}\left(\frac{s-i+1}{2}\right)
\end{array}
\end{align*}
The expected values can be approximated using Taylor expansion,
\begin{align*}
\begin{array}{rcl}
C_{1} & \approx & \log(\det( M(m)VM(m)^{\tp} )) \\
& & + \sum_{i=1}^{n_{x}} \sum_{j=1}^{n_{x}} \left.\frac{\diff^{2} \log(\det(M(\sx)VM(\sx)^{\tp}))}{\diff \sx_{i} \diff \sx_{j}}\right|_{\sx=m} P_{i,j} \\
C_{2} & \approx & M(m)VM(m)^{\tp} \\
& & + \sum_{i=1}^{n_{x}} \sum_{j=1}^{n_{x}} \left.\frac{\diff^{2} (M(\sx)VM(\sx)^{\tp})}{\diff \sx_{i} \diff \sx_{j}}\right|_{\sx=m} P_{i,j} 
\end{array}
\end{align*}
where $\sx_{i}$ is the $i$th element of $\sx$, $P_{i,j}$ is the $i,j$th element of $P$, and the differentiations are ($M_{\sx} = M(\sx)$ for brevity)
\begin{align*}
\begin{array}{rcl}
	\frac{\diff^{2} \log(\det(W))}{\diff \sx_{i} \diff \sx_{j}} & = & \tr\left( W^{-1} \frac{\diff^{2} W}{\diff\sx_{i} \diff\sx_{j}} -W^{-1} \frac{\diff W}{\diff \sx_{i}} W^{-1}\frac{\diff W}{\diff \sx_{j}} \right) \\
	\frac{\diff M_{\sx}VM_{\sx}^{\tp}}{\diff \sx_{j}} & = &  \frac{\diff M_{\sx}}{\diff \sx_{j}}V M_{\sx}^{\tp} + M_{\sx} V \frac{\diff M_{\sx}^{\tp}}{\diff \sx_{j}} \\
	\frac{\diff^{2} M_{\sx}VM_{\sx}^{\tp}}{\diff \sx_{i} \diff \sx_{j}} & = & \frac{\diff^{2} M_{\sx}}{\diff \sx_{i} \diff \sx_{j}}VM_{\sx}^{\tp} + \frac{\diff M_{\sx}}{\diff \sx_{j}}V\frac{\diff M_{\sx}^{\tp}}{\diff \sx_{i}} \nonumber \\
	& & + \frac{\diff M_{\sx}}{\diff \sx_{i}}V\frac{\diff M_{\sx}^{\tp}}{\diff \sx_{j}} + M_{\sx} V \frac{\diff^{2} M_{\sx}^{\tp}}{\diff \sx_{i} \diff \sx_{j}}
\end{array}
\end{align*}
\rule[0pt]{\columnwidth}{1pt}
\end{table}

Elliptically shaped group objects are tracked under kinematical constraints in \cite{KochF:2009}. A multiple model framework is used to handle different object types in \cite{LanL:2013JointTrackClass,CaoLL:2013}, leading to joint tracking and classification. New object spawning, and merging of two object's into a single object, is modelled within the random matrix framework in \cite{GranstromO:2013spawn}. The \mtt algorithms mentioned above all consider a single sensor. In \cite{VivoneBGW:2016fusion} %VivoneGBW:MultiSensor
the multi sensor case is considered, and four different updates are derived and compared using marine radar data. A random matrix estimator based on a Rao-Blackwellised state density, with a Gaussian for the kinematic state density and a particle approximation for the extent state density, is shown to have best performance, albeit at higher average cycle time that the other estimators \cite{VivoneBGW:2016fusion}. The random matrix model is applied to mapping in \cite{FatemiGSRH:2016_PMBradarmapping}, where a batch measurement update is presented, allowing all data to be processed at once instead of sequentially.

The random matrix model assumes an ellipse shape for the object's extent. For objects with irregular, non-ellipsoidal, extents, the shape can be approximated as a combination of several elliptically shaped subobjects. Using multiple instances of a simpler shape alleviates the limitations posed by the implied elliptic object shape\footnote{As the number of ellipses grows, their combination can form nearly any given shape.}, and also retains, on a subobject level, the relative simplicity of the random matrix model. In \cite{LanRL:2014} a single extended object model is given where the extended object is a combination of multiple subobjects with kinematic state vectors $\sx_{k}^{(i)}$ and extent matrices $\ext_{k}^{(i)}$, and each subobject is modelled using \eqref{eq:Koch_GIW_model}. Note that this model assumes independence between the subobjects. By modelling the subobject kinematic vectors as dependent random variables estimation performance can be improved significantly, see \cite{GranstromWBS:2015_MRMUK,Granstrom:MultiEllipseArxiv}. In \cite{LanL:2014} the non-ellipsoidal extended object model \cite{LanRL:2014} is used in a joint tracking and classification framework.  The work \cite{Zong2019_mbf} derives a multi-Bernoulli filter for extended targets based on sub-random matrices.

For performance evaluation of estimates computed using any of the random matrix predictions/updates, the Gaussian Wasserstein distance is the best performance measure \cite{MFI16_Yang}. For the random matrix prediction/update presented in \cite{FeldmannFK:2011}, see Tables~\ref{tab:FeldmannRMupdate} and \ref{tab:FeldmannRMprediction}, the posterior Cramer Rao Lower Bound \textsc{crlb} is given in \cite{SaritasO:2016}.

\subsection{Star-Convex Shape Approaches}
\label{sec:RandomHypersurfaceModel}

The star-convex shape approaches based on the random hypersurface model \cite{BaumH:2009,BaumH:2014} or its variant the Gaussian process model \cite{WahlstromO:2015,HirscherSRD:2016} constitute a general extended object tracking framework that employs
 \begin{itemize}
 \item a parametric representation of the shape contour,
 \item a Gaussian distribution for representing the uncertainty of the joint state vector of the kinematic and shape parameters, and
\item  nonlinear Kalman filters for performing the measurement update.
 \end{itemize}
 In contrast to the random matrix model that inherently relies on the elliptic shape, the approaches in this subsection are designed for general star-convex shapes (without using multiple subobjects). However, the increased flexibility comes at the price of much more complex closed-form formulas.
 
In the following, we first discuss the benefits of nonlinear Kalman filters for extended object tracking.
Next, the random hypersurface model and the Gaussian process model for star-convex shapes are introduced.
Finally, an overview of recent developments and trends in the context of random hypersurface models is given.
 
\subsubsection{Review -- Nonlinear Kalman Filtering}
Consider a general nonlinear measurement function (time index is omitted) in the form 
\begin{equation} \label{eqn:measeqn} 
\sz=h(\sx,\sv) \enspace , 
\end{equation}
which maps the state $\sx$ and the noise $\sv$ to the measurement $\sz$. We assume that both the prior probability density function of the state and noise density are Gaussian, i.e.,
 $p(\sx)=\Npdfbig{\sx}{m}{P}$  and  $p(\sv)= \Npdfbig{\sv}{0}{R}$.
In order to calculate the posterior density function 
\begin{equation}\label{eqn:posterior}
p(\sx | \sz) = \frac{p(\sz|\sx ) \cdot p(\sx)}{p(\sz)} \enspace ,
\end{equation}
  it is necessary to determine the likelihood function $p(\sz|\sx )$ based on \eqref{eqn:measeqn}.
Unfortunately, as the noise in \eqref{eqn:measeqn} is non-additive, no general closed-form solution for the likelihood is available.
As a consequence, nonlinear estimators that work with the likelihood function (e.g., standard particle filters) cannot be applied directly to  this kind of measurement equation.
However, there are nonlinear filters that do not explicitly calculate the likelihood function -- instead they exclusively work with the measurement equation \eqref{eqn:measeqn}.
The most prominent examples are \emph{nonlinear Kalman filters}, which directly apply the Kalman filter formulas to the nonlinear measurement equation \eqref{eqn:measeqn} in order to approximate the  mean $m^+$ and covariance $P^+$ of the posterior density \eqref{eqn:posterior} as 
 \begin{eqnarray}\label{eqn:nonlinearkf}
  m^+ &=& m+ \cov[\sz,\sx] P^{-1}  (\sz-\Exp{\sz}) \\ \label{eqn:nonlinearkf2}
  P^+ &=& P- \cov[\sx, \sz]  \cov[\sz, \sz]^{-1}   \cov[\sz, \sx] \enspace .
  \end{eqnarray}
  Of course, in case of high nonlinearity of the measurement equation, this can be a pretty rough approximation. The exact posterior is only obtained in case of a linear measurement equation. 
  
  Analytic expression for the required  moments $\Exp{\sz}$, $\cov[\sz,\sx]$, and $\cov[\sz, \sz]$ in \eqref{eqn:nonlinearkf} and \eqref{eqn:nonlinearkf2} are only available for special cases, e.g., polynomial measurement equations.
However, a huge variety of approximate methods has been developed in the past such as  the unscented transform \cite{Julier_UnscentedFiltering}. More advanced methods are discussed in  detail in \cite{Lefebvre2004}.

\begin{figure}
\center

\subfloat{
\begin{minipage}{4cm}\center
 \includegraphics[page=1,width=4cm] {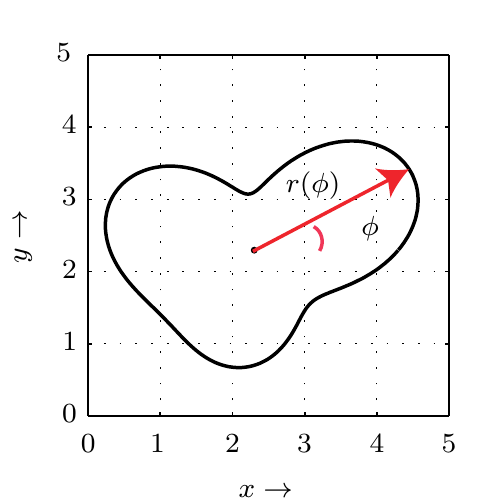}
 \label{fig:pointtarget}
 \end{minipage}
 } 
\subfloat{
\begin{minipage}{4.5cm}\center
 \includegraphics[page=1,width=4.5cm] {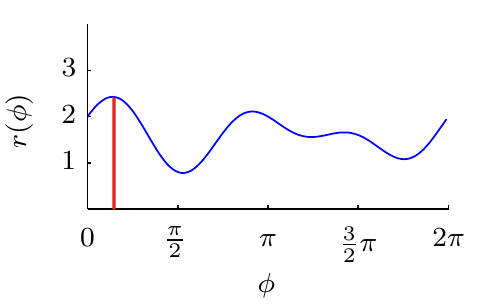}
 \label{fig:pointtarget}
 \end{minipage}

 } 
\caption{Illustration of the representation of a star-convex contour (left) with a radius function $r(\phi)$ (right).}
\label{fig:radiusfunction}
\end{figure}

\subsubsection{Random Hypersurface Model for Star-Convex Shapes}
In the following, it is shown how the extended object tracking problem can be formulated as a measurement equation with non-additive noise \eqref{eqn:measeqn}  using the  concept of a random hypersurface model.
Based on the derived measurement equation, nonlinear Kalman filters can be used to estimate the shape of extended objects as described above.

For this purpose, we first define a suitable parametrisation of
  a star-convex shape  based on the  so-called radius function
 $r(\xShape{k},\phi)$,  which maps a shape parameter vector $\xShape{k}$ and an angle  $\phi$ to a contour point (relative to a centre $\sxm_k$), see Figure~\ref{fig:radiusfunction} for an illustration.
 A  reasonable (finite dimensional) shape parameter vector $\xShape{k}$ can be defined by  a  Fourier series expansion \cite{Zhang2005}  with $N_F$ Fourier coefficients, i.e.,
 \begin{eqnarray*}\label{eqn:fourier_linear}
 r(\xShape{k},\phi) &=&  {R}(\phi) \cdot \xShape{k} \enspace ,
 \end{eqnarray*}
 where
 \begin{eqnarray*}
 {R}(\phi)&=& [\tfrac{1}{2}, \cos(\phi), \sin(\phi), \ldots , \cos(N_F\phi), \sin(N_F\phi)  ] \enspace , \\ 
 \xShape{k}&=& \tvect{\rv{a}_k^{(0)}, \rv{a}_k^{(1)},  \rv{b}_k^{(1)}, \ldots \rv{a}_k^{(N_F)},\rv{b}_k^{(N_F)} } \enspace .
 \end{eqnarray*}
 Fourier coefficients with small indices capture coarse shape features while coefficients with larger indices represent finer details.

 The overall state vector $\sxa_k$ consists of the shape parameters $\xShape{k}$, location $\sxm_k$, and kinematic parameters $\sxk_k$, i.e., 
 \begin{equation}\label{eqn:rhm_state}
 \sxa_k = \tvect{\xShape{k}^T , \sxm^T_k, \sxk^T_k} \enspace
 \end{equation}

 A suitable measurement equation  following the  random hypersurface philosophy  is formulated in polar form,
 \begin{equation}\label{eqn:starconv_measeqn}
  \sz_k =  s_k \cdot r(\xShape{k},\phi_k) + \sxm_k + \sv_k
 \end{equation}
where $s_k \in [0, 1]$ is  (multiplicative) noise  that specifies the relative distance of the measurement source from the center, and $\phi_k$ gives the angle to the measurement vector.
In \cite{BaumH:2011}, it has been shown that $s_k^2$ is uniformly distributed in case the measurement sources are uniformly distributed over the shape. It can be approximated by a Gaussian distribution with mean  $0.8$ and covariance $\frac{1}{12}$. 
By this means, the problem of estimating a (filled) shape has been reduced to a ``curve fitting'' problem, because for a fixed scaling factor $s_k$, \eqref{eqn:starconv_measeqn} specifies a closed curve. See also the discussion in Section~\ref{subsec:shapemodels}. 

The parameter $\phi_k$ can be interpreted as a nuisance parameter (or latent variable) as in errors-in-variables models for regression and curve fitting.
A huge variety of approaches for dealing with nuisance parameters has been developed in different areas.
The most simple (and most inaccurate approach) is to replace the unknown  $\phi_k$ with a point estimate, e.g.,  the angle between $\sxm_k - \sz_k$ and the $x$-axis. This approach can be seen as greedy association model \cite{Fusion15_Faion}.

Having derived the measurement equation \eqref{eqn:starconv_measeqn}, a measurement update can be performed using the formulas \eqref{eqn:nonlinearkf} and \eqref{eqn:nonlinearkf2}. As \eqref{eqn:starconv_measeqn} is polynomial for given $\phi_k$, closed-form formulas for the  moments in the update equations are available.

As the greedy association model yields to a bias in case of high noise, a so-called \emph{partial likelihood} has been developed, which outperforms the greedy association model in many cases \cite{Fusion15_Faion,Faion2015a}, e.g., high noise scenarios.
For star-convex shapes, the partial likelihood model  can be obtained from an algebraic reformulation of \eqref{eqn:starconv_measeqn} and, hence, does not come with additional complexity \cite{Fusion15_Faion,Faion2015a}. 

A further natural approach would be to assume $\phi_k$ to be uniformly distributed on the interval $[0, 2\pi]$, however,  a nonlinear Kalman filter implicitly approximates a uniform distribution by  a Gaussian distribution. Consequently, a reasonable mean for this Gaussian approximation is not obvious due to the circular nature of $\phi_k$.

Finally, we would like to note that due to the Gaussian state representation, prediction can be performed as usual in Kalman filtering, i.e., closed-form formulas are available for linear dynamic models and for nonlinear dynamic models, nonlinear Kalman filters can be employed.

\subsubsection{Gaussian Process Model for Star-Convex Shapes}
Instead of using a Fourier series expansion for modelling the shape contour, \cite{WahlstromO:2015} proposed to use Gaussian processes for star-convex shapes.
A Gaussian Process \cite{RasmussenW:2006} is a stochastic process which is completely defined by a mean function $\mu(u)$ and a kernel function $k(u,u')$:
\begin{align}
f(u) \sim \mathcal{GP}(\mu(u), k(u,u')).
\end{align}
For a finite number of inputs $u_1,\ldots,u_n$, a Gaussian process follows
\begin{align}
[f(u_1)~ \cdots ~ f(u_n)]^{\rm{T}} \sim \mathcal{N}(\bf{\mu}, K),
\end{align}
where
\begin{align}
\bf{\mu} &= [\mu(u_1) ~ \cdots ~ \mu(u_n)]^{\rm{T}}\\
K &= \begin{bmatrix}
k(u_1,u_1) & \cdots & k(u_1,u_n) \\
\vdots & \ddots & \vdots \\
k(u_n,u_1) & \cdots & k(u_n,u_n)
\end{bmatrix}.
\end{align}
Gaussian processes are often used in machine learning. In contrast to machine learning approaches, where batch processing it typically applied, tracking applications require a recursive estimate of the Gaussian process for shape representation. Thus, the function $f(u)$ is approximated by a finite number of function values or basis points which are updated over time. Consequently, the Gaussian process is described using a constant number of parameters which resembles the parameterization used in the random hypersurface model. However, the basis points are uniformly distributed over the angle interval, i.e., a separation of the basis points into points for coarse and fine shape features (cf. parameters for coarse and fine in \eqref{eqn:fourier_linear}) is not possible.

The kernel function $k$ restricts the kind of functions which can be represented by the Gaussian process, e.g. to symmetric functions \cite{WahlstromO:2015,HirscherSRD:2016}. Besides the Kalman filter based implementations, a Rao-Blackwellised particle filter implementation of the Gaussian process model for star-convex objects has been proposed in \cite{oezkanwahlstroem_raoblack}.
 
\subsubsection{Further developments, extensions and variations}
In the same manner as for star-convex shapes \cite{BaumH:2011}, the concept of a random hypersurface model can be applied for circular and elliptic shapes \cite{BaumNH:2010}. In this case, it is more suitable to describe the shape with an implicit function instead of a parametric form.

In many applications, the object to be tracked is symmetric, e.g., an aircraft or a vehicle. In this case specific improvements and adoptions can be performed in order incorporate symmetry information \cite{JAIF15_Faion,HirscherSRD:2016}.
The concept of scaling the boundary of a curve in order to model an extended object has been combined with level sets in \cite{ZeaFBH:2013} in order to model arbitrary connected shapes.
A closed-form likelihood for the use in nonlinear filters based on the RHM measurement equation \eqref{eqn:starconv_measeqn} has been derived in \cite{SteinbringBZFH:2015}.
Elongated objects are considered in \cite{Zeo_Fusion_2016}.
The RHM idea can be used in the same manner to model three-dimensional shapes in three-dimensional space.
In addition,  two-dimensional shapes in three-dimensional space can also be modelled with RHM ideas \cite{Faion2015a,JAIF15_Faion}. For example, in \cite{JAIF15_Faion}, measurements from a cylinder are modelled by means of translating a ground shape, i.e., a circle.

It is interesting to note that clutter detections that are not from the extended object, can improve shape estimation  \cite{Fusion15_Zea, MFI16_Zea} by modelling them as negative information. Furthermore, camera calibration can be performed by means of tracking an extended object  \cite{MFI15_Faion}.

\subsubsection{Multiplicative Error Model}
The basic idea of the RHM is to model one dimension of the spatial extent with a random scaling factor and the other one with, e.g.,  a greedy association model (GAM).
By this means,  Bayesian inference becomes tractable with a nonlinear Kalman filter.

A recent line of work models both dimensions with a scaling factor \cite{Fusion12_Baum,YangBaum2016_Fusion,Yang2017_ICASSP}, i.e., multiplicative noise. By this means, a uniform distribution can be matched better for simple shapes, such as circles or ellipses. The resulting model is called Multiplicative Error Model (MEM). 

The tracking of elliptically shaped object state vector is in the same vein as \eqref{eqn:rhm_state}, i.e.,
\begin{equation} \label{eqn:statemem}
 \sxa_k = \tvect{ \sxk^T_k,\xShape{k}^T } \enspace
\end{equation}
where
$\sxk^T_k$ is kinematic state, and shape variable 
$$\xShape{k}^{\tp}= [\alpha_k ~ l_{k,1} ~ l_{k,2}]^{\tp}$$
with ellipse orientation $\alpha_k$,  and semi-axes lengths $l_{k,1}$ and $l_{k,2}$. 
Then the $i$th measurement at time $k$  is modelled as  
\newcommand*\vect[1]{\begin{bmatrix}#1\end{bmatrix}}
\begin{equation}\label{eqn:MEM}
  \sz_k^i = \mathbf{H} \sxa_k+ \mat{Rot}(\alpha_k) \vect{l_{k,1}&0\\0&l_{k,2}}\vect{h_{k,1}^i\\h_{k,2}^i}+ \sv_k^i
 \end{equation}
where $\mathbf{H}_k = \vect{\mathbf{I}_2, \mathbf{0}_2  }$ is  a matrix that picks out the object position from the  kinematic state,  
\begin{align}
\mat{Rot}(\alpha_k)=\vect{\cos{\alpha_k} & -\sin{\alpha_k}\\\sin{\alpha_k} & \cos{\alpha_k}}
\end{align}
is a rotation matrix, $\sv_k^i$ is additive sensor noise, and both  $h_{k,1}^i$  and $h_{k,2}^i$ are (Gaussian) multiplicative noise terms that we assume to be  mutually independent of all other random variables. Following the reasoning for the parameter $z$ in \eqref{eq:GaussianReflectionPoints}, the variances of the multiplicative noise are set to $\sigma_{h_1}=\sigma_{h_2}=\frac{1}{4}$ in order to match an elliptic uniform spatial distribution.
In this manner, the multiplicative noise models the spatial distribution, i.e., the uncertainty of the measurement source.
The corresponding likelihood to \eqref{eqn:MEM} coincides with the likelihood used in the random matrix approach, i.e., \eqref{eq:Feldmann_measurement_model}, but the ellipse parametrisation is different.

Unfortunately, it turns out that a direct application of the Kalman filter formulas to \eqref{eqn:MEM} does not give satisfying results \cite{Fusion12_Baum} due to the strong linearities. A solution is to augment the original measurement equation \eqref{eqn:MEM} with the squared measurement $\sz^2$ using the Kronecker product and then apply a nonlinear Kalman filter. In this way, higher order moments are incorporated in the update formulas.
For this purpose, an  Extended Kalman filter  is derived  in \cite{Yang2017_ICASSP}  that results in compact update formulas for the extent, which are depicted in Table~\ref{tab:MEMupdate}. Exact prediction can be performed analytically for linear models, see Table~\ref{tab:MEMprediction}.

\begin{table}
\caption{Update of the EKF for the multiplicative error model \cite{Yang2017_ICASSP}. Source code: \protect\url{http://github.com/Fusion-Goettingen}.}
\label{tab:MEMupdate}
\vspace{-5mm}
\rule[0pt]{\columnwidth}{1pt}
\textbf{Input:} Kinematic state prior mean $m^{\sxk}$ and covariance $P^{\sxk}$, shape variable prior mean $m^\xShape{}$ and covariance $P^\xShape{}$ as defined in \eqref{eqn:statemem}, measurement matrix $\mathbf{H}$, measurement noise covariance $R$, multiplicative noise variance $\sigma_{h_1}$ and $\sigma_{h_2}$, measurement $\sz$

\textbf{Output:} Updated parameters $m^{\sxk}_{+}$, $P^{\sxk}_{+}$, $m_{+}^{\xShape{}}$ and $P_{+}^{\xShape{}}$,
\begin{align*}
\begin{array}{rcl}
m^{\sxk}_{+} &= &m^{\sxk} + \cov[\sxa,\sz]\left({\cov[\sz,\sz]}\right)^{-1}(\sz-\Exp{\sz})\\
 P^{\sxk}_{+} &=& P^{\sxk} - \cov[\sxa,\sz]\left(\cov[\sz,\sz]\right)^{-1}\left(\cov[\sxa,\sz]\right)^{\tp}\\
 m^{\xShape{}}_{+} &= &m^{\xShape{}} + \cov[\sxa,\boldsymbol{\tilde{z}}]\left({\cov[\boldsymbol{\tilde{z}},\boldsymbol{\tilde{z}}]}\right)^{-1}(\boldsymbol{z}-\Exp{\boldsymbol{\tilde{z}}})\\
 P^{\xShape{}}_{+} &=& P^{\xShape{}} - \cov[\sxa,\boldsymbol{\tilde{z}}]\left(\cov[\boldsymbol{\tilde{z}},\boldsymbol{\tilde{z}}]\right)^{-1}\left(\cov[\sxa,\boldsymbol{\tilde{z}}]\right)^{\tp}\\
% \cov[\mathcal{Z},\mathcal{Z}]&=&\diag{\cov[\sz,\sz] ~\cov[\boldsymbol{z},\boldsymbol{z}]}\\
% \cov[\sxa,\mathcal{Z}]&=&\diag{\cov[\sxm,\sz], \cov[\xShape,\mathbf{z}]}\\
% \mathcal{Z}&=&\vect{\sz~\boldsymbol{z}}^{\tp}\\
% \Exp{\mathcal{Z}}&=&\vect{\Exp{\sz}~\Exp{\boldsymbol{z}}}^{\tp}\\
  \Exp{\sz}&=&  \mathbf{H} m^{\sxk}\\
  \cov[\sxk,\sz] &=&P^{\sxk} \mathbf{H}^T\\
 \cov[\sz,\sz] &=& \mathbf{H}   P^{\sxk} \mathbf{H}^T+S\diag{\sigma_{h_1},\sigma_{h_2}}S^{\tp}+R\\
 S &=&\vect{\cos{\alpha} & -\sin{\alpha}\\\sin{\alpha} & \cos{\alpha}}\diag{l_1,l_2} \\
 \vect{\alpha~l_1~l_2}^{\tp}&=&m^{\xShape{}}\\
   \boldsymbol{\tilde{z}}&=&\vect{1&0&0&0\\0&0&0&1\\0&1&0&0}\left(\left(\sz- \Exp{\sz}\right)\otimes\left(\sz- \Exp{\sz}\right)\right)\\
\vect{\sigma_{11} &\sigma_{12}\\\sigma_{12} &\sigma_{22}} &=&   \cov[\sz,\sz] \\
\Exp{\boldsymbol{\tilde{z}}}&=&\vect{\sigma_{11} &\sigma_{22}&\sigma_{12} }^{\tp}\\
 \cov[\boldsymbol{\tilde{z}},\boldsymbol{\tilde{z}}]&=&
                    \vect{
                        3\sigma_{11}^2 &\sigma_{11}\sigma_{22}+2\sigma_{12}^2 & 3\sigma_{11}\sigma_{12}\\
						\sigma_{11}\sigma_{22}+2\sigma_{12}^2&3\sigma_{22}^2 & 3\sigma_{22}\sigma_{12}\\
						3\sigma_{11}\sigma_{12} & 3\sigma_{22}\sigma_{12} &\sigma_{11}\sigma_{22}+2\sigma_{12}^2 
						}\\
\cov[\xShape,\boldsymbol{\tilde{z}}] &=&   P^{\xShape{}}M^{\tp}\\
  M &=&
            	\vect{ -\sin{2\alpha} & \cos^2{\alpha}& \sin^2{\alpha}\\
  		            	\sin{2\alpha} & \sin^2{\alpha} &\cos^2{\alpha}\\
 		            	\cos{2\alpha} &  \sin{2\alpha} & -\sin{2\alpha}}\\%\diag{\vect{(l_{1})^2\sigma_{h_1}-(l_{2})^2\sigma_{h_2}\\2l_{1}\sigma_{h_1}\\2l_{2}\sigma_{h_2}}}\\
 		      && \cdot \vect{(l_{1})^2\sigma_{h_1}-(l_{2})^2\sigma_{h_2}&0&0\\ 0&2l_{1}\sigma_{h_1}&0\\0&0& 2l_{2}\sigma_{h_2}}
\end{array}
\end{align*}
\rule[0pt]{\columnwidth}{1pt}
\end{table}

\begin{table}
\caption{Prediction of the EKF for the multiplicative error model prediction \cite{Yang2017_ICASSP}. \newline{} Source code: \protect\url{http://github.com/Fusion-Goettingen}.}
\label{tab:MEMprediction}
\vspace{-5mm}
\rule[0pt]{\columnwidth}{1pt}
\textbf{Input:} Kinematic state prior mean $m^{\sxk}$ and  covariance $P^{\sxk}$, shape variable prior mean $m^\xShape{}$ and covariance $P^\xShape{}$, process matrices  $\mathbf{F}^{\sxk}$, $\mathbf{F}^{\xShape{}}$ with process noise covariances $Q^{\sxk}$ and $Q^{\xShape{}}$

\textbf{Output:} Parameters $m^{\sxk}_{*}$, $P^{\sxk}_{*}$, $m_{*}^{\xShape{}}$ and $P_{*}^{\xShape{}}$ for the prediction
\begin{align*}
\begin{array}{rcl}
m^{\sxk}_{*}&=& \mathbf{F}^{\sxk}m^{\sxk}\\
P^{\sxk}_{*}&=&\mathbf{F}^{\sxk}P^{\sxk}(\mathbf{F}^{\sxk})^{\tp}+Q^{\sxk}\\
m^{\xShape{}}_{*}&=&\mathbf{F}^{\xShape{}}m^{\xShape{}}\\
P^{\xShape{}}_{*}&=&\mathbf{F}^{\xShape{}}P^{\xShape{}}(\mathbf{F}^{\xShape{}})^{\tp}+Q^{\xShape{}}
\end{array}
\end{align*}
\rule[0pt]{\columnwidth}{1pt}
\end{table}

%%%%%%%%%%%%%%%%%%%%%%%%%%%%%%%%%%%%%%%%%%%%%%%%%%%%%%%%%%%%%%%%%%%%%%%%%%%%%%%%%%%%%%%%%%
%%%%%%%%%%%%%%%%%%%%%%%%%%%%%%%%%%%%%%%%%%%%%%%%%%%%%%%%%%%%%%%%%%%%%%%%%%%%%%%%%%%%%%%%%%
%%%%%%%%%%%%%%%%%%%%%%%%%%%%%%%%%%%%%%%%%%%%%%%%%%%%%%%%%%%%%%%%%%%%%%%%%%%%%%%%%%%%%%%%%%

\section{Tracking multiple extended objects}
\label{sec:MultipleExtendedTargetTracking}

In this section we overview multiple extended object tracking. Regardless of the type tracking problem---point, extended, group, etc---\mtt is a problem that has many challenges:
\begin{itemize}
	\item The number of objects is unknown and time varying.
	\item There are missed measurements, i.e., at each time step, some of the existing objects do not give measurements.
	\item The objects that are not missed give rise to an unknown number of detections.
	\item There are clutter measurements, i.e., measurements that were not caused by a target object.
	\item Measurement origin is unknown, i.e, the source of each measurement is unknown. This is often referred to as the ``data association problem''.
\end{itemize}
For multiple point object tracking the literature is vast; recently a comprehensive overview of \mtt algorithms, with a focus on point objects, was written by Vo et al \cite{VoMBSCOMV:2015}. Since many of the existing extended object \mtt algorithms are of the \rfs type, we focus on these algorithms in the following (see \ref{sec_non_RFS} for selected approaches with other \mtt algorithms). In the following subsections we will first give a brief overview of \rfs filters, then we give examples of extended and group object \mtt algorithms, and lastly we discuss the data association problem in extended object \mtt.

\subsection{Review -- \rfs filters}

A random finite set (\rfs) is a set whose cardinality is a random variable, and whose set members are random variables. In \rfs based tracking algorithms both the set of objects and the sets of measurements are modelled as \rfss. Tutorials on \rfs methods can be found in, e.g., \cite{Mahler:2013,VoVC:2013,GranstromLGO:2014}, and in-depth descriptions of the \rfs concept and of finite set statistics (\fisst) are given in the books \cite{mahler_book_2007,Mahler:2014}.

The state of the set of objects that are present in the surveillance space is referred to as the \textit{multi-object state}. Because of the computational complexity, specifically due to the data association problem, a full multi-object Bayes filter can be quite computationally demanding to run, and approximations of the data association problem are necessary. Computationally tractable filters include the Probability Hypothesis Density (\phd) filter \cite{mahler_AES_2003_PHD}, the Cardinalized \phd (\cphd) filter \cite{Mahler:2007}, the Cardinality Balanced \member (\cbmember) filter \cite{VoVC:2009}, and the \mtt conjugate priors \cite{VoV:2013,Williams:2015conjprior}.

\subsubsection{PHD and CPHD filters}
The first order moment of the multi-object state is called the \phd\footnote{The first order moment is also called intensity function, see, e.g., \cite{Mahler:2013,VoSD:2005}.}, and can be said to be to a random set as the expected value is to a random variable. A \phd filter recursively estimates the \phd under an assumed Poisson distribution for the cardinality. A consequence of the Poisson assumption is that the \phd filter's cardinality estimate has high variance, a problem that manifests itself, e.g., where there are missed measurements \cite{ErdincWBS:2005}. The \cphd filter recursively estimates the \phd and  a truncated cardinality distribution, and is known to have a better cardinality estimate compared to the \phd filter. The \phd and \cphd filters were first derived in \cite{mahler_AES_2003_PHD,Mahler:2007} using probability generating functionals\footnote{The probability generating functional is an integral transform that can be used when working with \rfs densities, see further in, e.g., \cite{mahler_book_2007,Mahler:2014}.}. In \cite{GarciaFernandezV:2015} it is shown that the \phd and \cphd filters can be derived by minimizing the Kullback-Leibler divergence \cite{KullbackL:1951} between the multi-object density and either a \ppp density (\phd filter) or an iid cluster process density (\cphd filter). 

In both the \phd filter and the \cphd filter the objects are independent identically distributed (iid); the normalized \phd is the estimated object pdf. 
When there are multiple objects the \phd has multiple modes (peaks), where each mode corresponds to one object. An exception to this is when two or more objects are located close to each other; in this case a mode can correspond to multiple objects, also called \emph{unresolved} objects. The estimated number of objects located in an area, e.g. under one of the modes, is given by integrating the \phd over that area.
Both the \phd filter and the \cphd filter are susceptible to a ``spooky effect'' \cite{FrankenSU:2009,Mahler:2014}, a phenomenon manifested by \phd mass shifted from undetected objects to detected objects, even in cases when the objects are far enough away that they ought to be statistically insulated.

Ultimately the desired output from an \mtt algorithm is a set of estimated trajectories (tracks), where a trajectory is defined as the sequence of states from the time the object appears to the time it disappears. In their most basic forms neither the \phd nor the \cphd formally estimate object trajectories. However, object trajectories can be obtained, e.g. using post-processing with labelling schemes \cite{PantaCV2009,GranstromNBLS:2014,GranstromNBLS:2015_TGARS}. 

\subsubsection{CB-MeMBer filter}

The \cbmember filter \cite{VoVC:2009} approximates the multi-object density with a multi-Bernoulli (\mb) density \cite[Ch. 17]{mahler_book_2007}. In an \mb density the objects are independent but not identically distributed, compared to the \phd and \cphd filters where the objects are iid. The Bernoulli \rfs density is a suitable representation of a single object, as it captures both the uncertainty regarding the object's state, as well as the uncertainty regarding the object's existence. As the name suggests, an \mb density is the union of several independent Bernoulli densities, and it is therefore a suitable representation of multiple objects. The \cbmember filter fixes the biased cardinality estimate of the \member filter presented in \cite[Ch. 17]{mahler_book_2007}.

\subsubsection{MTT conjugate priors}
The concepts \emph{conjugacy} and \emph{conjugate prior} are central in Bayesian probability theory. In an \mtt context, conjugacy means that if we begin with a multi-object density of a conjugate prior form, then all subsequent predicted and updated multi-object densities will also be of the conjugate prior form. Two \mtt conjugate priors can be found in the literature, both based on multi-Bernoulli representations for the set of objects.

The first is based on labeled \rfss and is called Generalized Labeled Multi-Bernoulli (\glmb) \cite{VoV:2013}. In the \glmb filter the labels are used to obtain target trajectories. Because of the unknown measurement origin, the \glmb has a mixture representation, where each component in the mixture corresponds to one possible data association history. The \glmb filter performs well in challenging scenarios, however, it is computationally expensive. A computationally efficient approximation is the Labeled Multi-Bernoulli (\lmb) filter \cite{ReuterVVD:2014}, which approximates the \glmb mixture with a single labeled multi-Bernoulli density. Both the \glmb and \lmb filters rely on handling the data association problem by computing the $M$ top ranked assignments, an analysis of the approximation error incurred by this is presented in \cite{VoVP:2014}.

The second \mtt conjugate prior is based on regular \rfss, i.e., unlabeled, and is called Poisson Multi-Bernoulli Mixture (\pmbm) \cite{Williams:2015conjprior}. The \pmbm conjugate prior allows an elegant separation of the set of objects into two disjoint subsets: objects that have been detected, and objects that have not yet been detected. A Poisson point process density is used for the undetected objects, and a multi-Bernoulli mixture is used for the detected objects. Explicitly modelling the objects that have not been detected is useful, e.g., when the sensor is susceptible to occlusions, or when the sensor is mounted to a moving platform. Similarly to the \glmb filter, in the \pmbm filter the components in the multi-Bernoulli mixture corresponds to different data association histories. A variational Bayesian approach to approximating the multi-Bernoulli mixture density with a single multi-Bernoulli density is presented in \cite{Williams:2015}, leading to the Variational Multi-Bernoulli (\vmb) filter. Note that the variational approximation does not affect the Poisson part that models the undetected objects. The \vmb filter can be understood to be to the \pmbm filter, as the \lmb filter is to the \glmb filter. However, it should be noted that the approximations used in the \vmb and \lmb are not the same.

\subsection{Examples of extended and group \mtt}
\subsubsection{PHD and CPHD filters}
A \phd filter for extended objects under the Poisson model \cite{GilholmGMS:2005}, see also Section~\ref{sec:SpatialMeasurementModelling}, was presented in \cite{mahler_FUSION_2009_extTarg}. Gaussian mixture implementations of this extended object \phd filter, for both linear and non-linear motion and measurement models, are presented in \cite{GranstromLO:2010,GranstromLO:2011,GranstromLO:2012}. The resulting filters can be abbreviated \etgmphd filters. A Gaussian inverse Wishart implementation, using the random matrix extended object model \cite{Koch:2008} (see also Section~\ref{sec:RandomMatrixModel}), is presented in \cite{GranstromO:2012a,GranstromO:2012d}, and the resulting filter is abbreviated \giwphd filter. A Gaussian mixture implementation using RHMs (see Section~\ref{sec:RandomHypersurfaceModel}) was presented in \cite{zhang2013phd}. Multiple model Gaussian mixture \phd filters can be found in \cite{GranstromL:2013,GranstromRMS:2014}; the filters are applied to tracking of cars and bicycles, under assumed rectangle and stick shape models, and it is shown that using multiple measurement models can improve the estimation results. Augmenting the implementations with gamma distributions makes it possible to estimate the unknown Poisson measurement rate for each object \cite{GranstromO:2012c}. The resulting algorithms are then called gamma Gaussian (\gamg), or gamma Gaussian inverse Wishart (\ggiw), respectively. 

An approach to group object tracking based on a point object \gmphd filter is presented in \cite{ClarkG:2007}. The extended object \phd filter presented in \cite{SwainC:2010,SwainC:2012} is derived for an object model different from the Poisson point process model \cite{GilholmGMS:2005}. The objects are modelled by a  Poisson cluster process, a hierarchic process with a parent process and a daughter process. The parent process models a Poisson distributed number of objects. For each object a daughter process models a number of reflection points that generate measurements. An implementation is proposed where the object is assumed ellipse shaped and the reflection points are located on the edge of the ellipse.

At least two different \cphd filters have been presented. The \cphd filter for extended objects presented in \cite{LianHLLS:2012} is derived under the assumption that \emph{``relative to sensor resolution, the extended objects and the unresolved objects are not too close and the clutter density is not too large''}~\cite[Corollary 1]{LianHLLS:2012}. However, this is an assumption that cannot be expected to hold in the general case. A \cphd filter capable of handling both spatially close objects and dense clutter is presented in \cite{OrgunerLG:11techreport,OrgunerArxivCPHD2010,OrgunerLG:2011,LundquistGO:2013}, and a \ggiw implementation is also presented. A comparison shows that the \ggiw-\cphd filter outperforms the \ggiw-\phd filter, especially when the probability of detection is low, and/or the clutter density is high. The price for the increased performance is that the computational cost increases.

\subsubsection{CB-MeMBer filters}

An extension of the \cbmember filter \cite{VoVC:2009} to extended objects, using the \ppp measurement model overviewed in Section \ref{sec:SpatialMeasurementModelling}, was presented in \cite{ZhangLH:2014}. A Gaussian mixture implementation is presented in \cite{ZhangLH:2014}, and Sequential Monte Carlo (\textsc{smc}) implementations of the \cbmember for extended objects can be found in \cite{LiuJZ:2015,MaLL:2016}. An extended object \cbmember filter with multiple models is presented in \cite{JiangLFZ:2016}.

\subsubsection{Conjugate priors}
Labeled \mb filters for extended object tracking are presented in \cite{BeardRGVVS:2015,BeardRGVVS:2015journal}, both a \glmb filter and its approximation the \lmb filter. \ggiw implementations are presented, and simulation results show that the labelled \mb filters outperform their \phd and \cphd counterparts. Additionally, the \glmb and \lmb filters estimate object trajectories, which the \phd and \cphd filters do only if labeling is used in post processing, see e.g., \cite{GranstromNBLS:2014,GranstromNBLS:2015_TGARS}. The \lmb filter was applied to \lidar data for rectangular objects using the separable likelihood approach \cite{ScheelRD:2016} and for star-convex objects using a modelling with Gaussian processes \cite{HirscherSRD:2016}.

A \pmbm filter for extended and group objects is derived and presented with a \ggiw implementation in \cite{GranstromFS:2016fusion,GranstromFS:2016submitted}. A simulation study showed that the extended object \pmbm filter outperforms the \phd, \cphd and \lmb filters, and an experiment with \lidar data illustrates that the \ppp model can accurately represent the occluded areas of the surveillance space. The \ggiw-\pmbm model is applied to mapping in \cite{FatemiGSRH:2016_PMBradarmapping}, where a batch measurement update is derived.

\subsubsection{Non-RFS approaches}
\label{sec_non_RFS}
A Gaussian Mixture Markov Chain Monte Carlo filter for multiple extended object tracking is presented in \cite{CarmiSG:2012}. The filter is compared to the linear \etgmphd-filter \cite{GranstromLO:2010,GranstromLO:2012}, and is shown to be less sensitive to clutter but also considerably more computationally costly (as measured by the average cycle time). The Probabilistic Multi-Hypothesis Tracker (\textsc{pmht}) \cite{StreitL:1993} allows more than one measurement per object, and the random matrix model (Section~\ref{sec:RandomMatrixModel}) has been integrated in the \textsc{pmht} framework, see \cite{WienekeK:2010,WienekeD:2011,WienekeK:2012}.  A variational Bayesian Expectation Maximisation approach to mapping with extended objects is presented in \cite{LundgrenSH:2015}.

\subsection{Multiple extended object data association}

In \mtt a data association specifies for each measurement the source from whence it came: either it is an object measurement or a clutter measurement. The possibility of multiple measurements per object means that in extended object \mtt a data association can be split into two parts:
\begin{enumerate}
	\item \textbf{Partition:} A partition of a set, denoted $\altpartition$, is defined as a division of the elements of the set into non-empty subsets, called cells \cite{mahler_FUSION_2009_extTarg} and denoted $\altcell$, such that each element belongs to one and only one cell. The cells are to be understood to contain measurements that are from the same source, i.e., all measurements in the cell are from the same extended object, or they are all clutter.
	\item \textbf{Cell association:} An association of the cells to a measurement source, either one of the objects or a clutter source.
\end{enumerate}
Note that an association from measurement to cell, and from cell to source, defines an association from measurement to source.

For Bayes optimality it is necessary to consider all possible data associations in the \mtt update. This means that in extended and group \mtt it is necessary to consider all possible partitions of the set of measurements, and for each partition one has to consider all possible cell associations. Unless the measurement set contains a trivial number of measurements (i.e., extremely few) and there is a trivial number of objects, both of these problems are intractable because there are too many possible partitions, and too many possible cell associations. Fortunately, in the literature we can find methods that allow us to handle both of these problems. Below we first discuss the complexity of the partitions and the cell associations, and then we overview the solutions to these problems that can be found in the literature. 

\subsubsection{Complexity analysis}
\begin{figure}
	\includegraphics[width = \columnwidth]{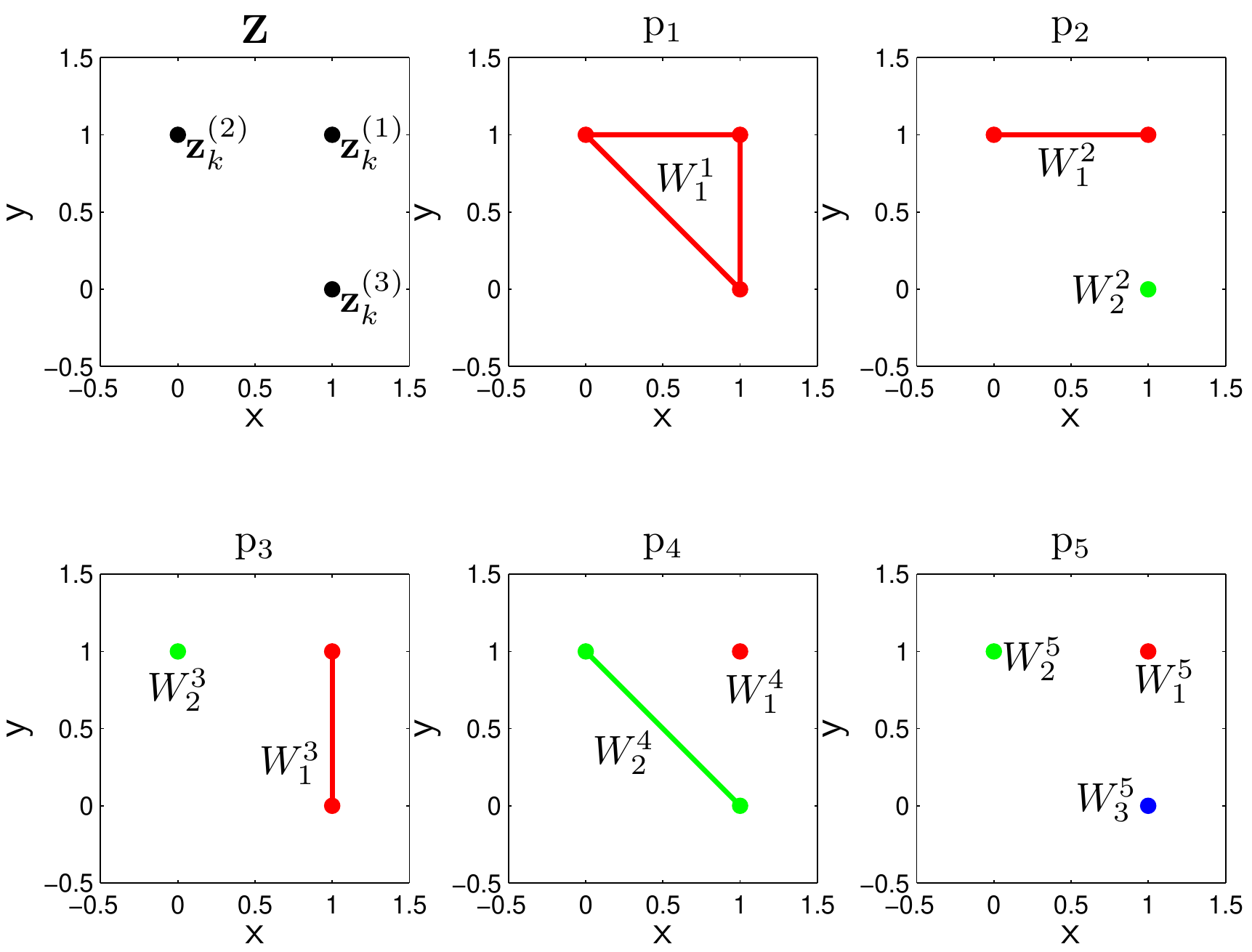}
	\caption{Partition illustration. There are three measurements $\sz_{k}^{(1)}$, $\sz_{k}^{(2)}$ and $\sz_{k}^{(3)}$, which can be partitioned in $5$ different ways. In the $j$th partition, denoted $\partition_{j}$, the $i$th cell is denoted $\cell_{i}^{j}$. With three measurements there is one partition with one cell, three partitions with two cells, and one partition with three cells. Note that the ordering of the partitions and cells is arbitrary; the particular ordering in this example is only used for notational simplicity.}
	\label{fig:measurement_partitioning}
\end{figure}

Let the set of measurements contain $n$ measurements in total. The number of possible ways to partition a set of $n$ measurements is given by the $n$th Bell number, denoted $B(n)$ \cite{Rota:1964}. The sequence of Bell numbers is log-convex\footnote{The sequence of Bell numbers is logarithmically convex, i.e., $B(n)^{2} \leq B(n-1)B(n+1)$ for $n\geq 1$ \cite{Engel:1994}. If the Bell numbers are divided by the factorials, $\frac{B(n)}{n!}$, the sequence is logarithmically concave, $\left(\frac{B(n)}{n!}\right)^{2} \geq \frac{B(n-1)}{(n-1)!} \frac{B(n+1)}{(n+1)!}$, for $n\geq 1$ \cite{Canfield:1995}. }, and $B(n)$ grows very rapidly as $n$ grows. For $n=3$ measurements there are $B(3)=5$ possible partitions; an example is shown in Figure~\ref{fig:measurement_partitioning}.
For twice the number of measurements ($n=6$) there are $B(6)=203$ possible partitions, and for $n=90$ measurements there are $B(90)>10^{100}$ possible partitions. In other words, it is computationally intractable to consider all partitions, and approximations are necessary for implementation. 

Let $|\altpartition|$ be the number of cells in the partition $\altpartition$, and let $m$ be the prior number of object estimates. Each cell can either be from one of the existing prior object estimates, or it could be from a new object. Thus, there are $|\altpartition|+m$ possible sources. The number of possible ways to associate $|\altpartition|$ cells to $|\altpartition|+m$ sources is
\begin{align}
	{m+|\altpartition| \choose |\altpartition|} = \frac{\left(m+|\altpartition|\right)!}{m! |\altpartition|!}
\end{align}
Similarly to the partitions, unless the number of cells and number of objects are very small, it is infeasible to consider all possible associations. 

\subsubsection{Complexity reduction}

The \mtt literature contain several different methods that can be used to alleviate the complexity, and that allows extended object \mtt filters to be implemented using limited computational resources.

Gating, see, e.g., \cite[Sec. 2.2.2.2]{BarShalomB:2000}, is a method that removes possible measurement-to-object associations by comparing the measurements to predictions of the objects' measurements. If the difference between the measurement and the predicted measurement is too large, the association is ruled out as infeasible. Gating has been used in a plethora of \mtt algorithms, both for point targets and extended targets. Naturally, for extended targets the gates must take into account the position of the target, the size and shape of the target, as well as state uncertainties. Using gating it is possible to group the measurements and the objects into smaller groups that, given the gating decision, are independent. This way one can solve several smaller data association problems instead of one larger data association problem.

Even after gating, there are typically too many possible partitions and cell associations. An important contribution of \cite{GranstromLO:2010,GranstromLO:2012,GranstromO:2012a} is to show how clustering can be used to find a subset of partitions. The basic insight behind the use of clustering lies in the definition of extended objects: the measurements are spatially distributed around the object. Therefore spatially close measurements are more likely to be from the same object, than spatially distant measurements. By only considering the partitions in which the cells contain spatially close measurements many partitions can be pruned, and the update becomes tractable. 

Distance Partitioning \cite{GranstromLO:2010,GranstromLO:2012} is a simple method that puts measurements in the same cell if the distance between a measurement and its closest neighbour is less than a threshold. A detailed description of Distance Partitioning is given in \cite{GranstromLO:2010,GranstromLO:2012,GranstromOML:Errata}.  By considering multiple thresholds, a subset of partitions is obtained. Finding a good subset of partitions is especially important when multiple extended objects are located in close vicinity of each other, see \cite{GranstromO:2012a,LundquistGO:2013,BeardRGVVS:2015journal}.

\begin{figure}
	\includegraphics[width = \columnwidth]{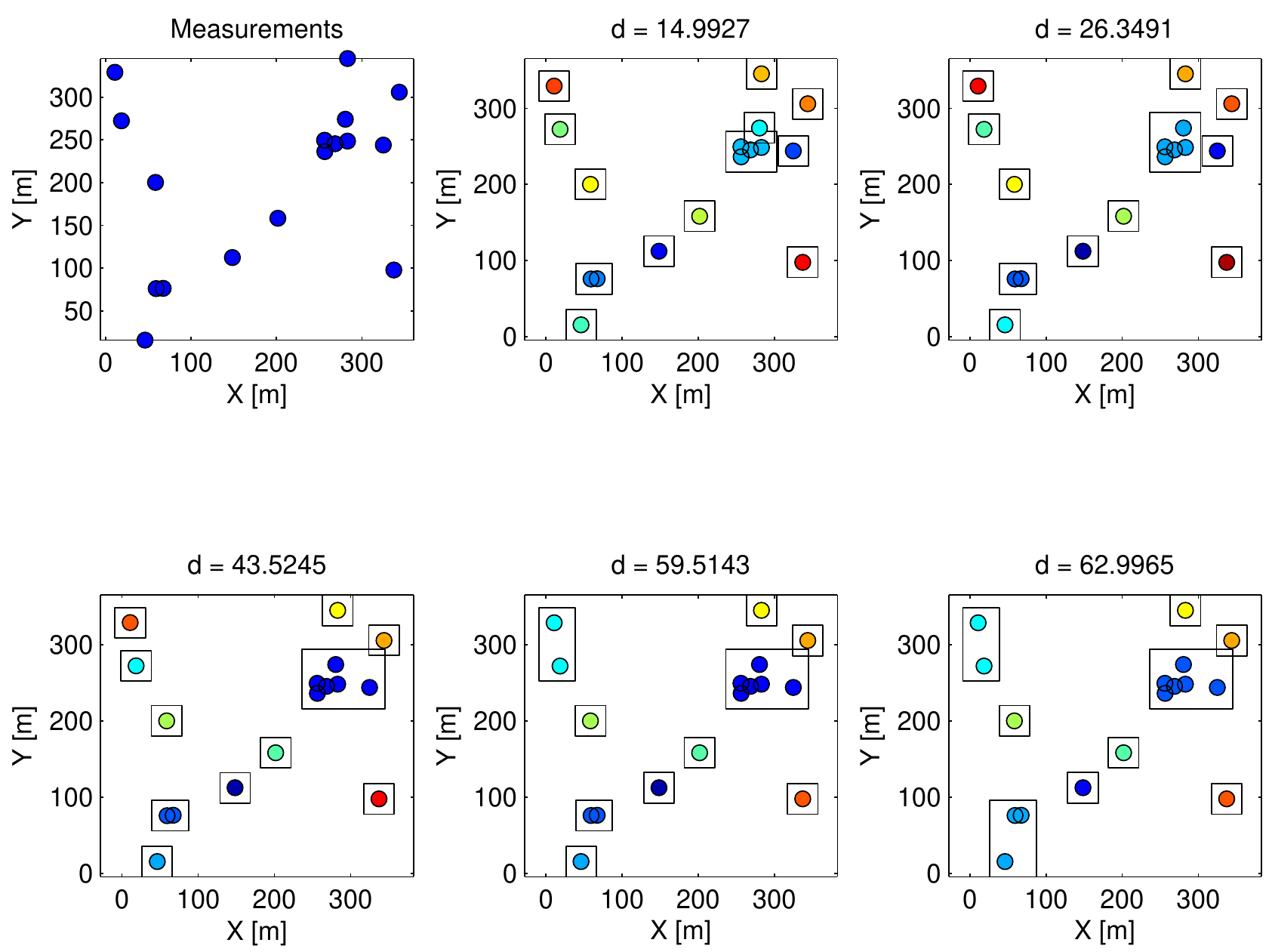}
	\caption{Illustration of the output form Distance Partitioning, with $17$ measurements. By clustering the measurements with progressively larger thresholds $d$ different partitions are obtained. The smallest and largest threshold that are used are parameters of the clustering algorithm.}
	\label{fig:meas_part_heuristic}
\end{figure}

An example where Distance Partitioning is used is given in Figure~\ref{fig:meas_part_heuristic}. In this example there are $17$ measurements, for which there are more than $10^{10}$ possible partitions. Using Distance Partitioning this is limited to $5$ partitions.
Results from both simulations and experiments have shown that, despite the very drastic reduction in the number of partitions that are considered, performance is not sacrificed when clustering is used, see, e.g., \cite{GranstromL:2013,GranstromRMS:2014,ScheelGMRD:2014,GranstromNBLS:2015_TGARS}. However, there may be scenarios where two objects are so close to each other, that their measurements may not be separated any more based on the distance. In these scenarios, prior information about the number of objects (e.g. based on the current cardinality estimate) may be used to improve partitioning (cf. \cite{GranstromO:2012a}). 

Distance Partitioning is an example of a hierarchical single linkage clustering algorithm, see, e.g., \cite{Bishop:2006} for a discussion about clustering. Other clustering methods have also been used in an extended object \mtt context, e.g., Gaussian Mixture Expectation Maximisation \cite{GranstromO:2012a}, spectral clustering \cite{YangLGY:2014}, and fuzzy adaptive resonance theory \cite{ZhangJ:2013,ZhangJ:2014}.

The extended object \phd and \cphd filters avoid the cell association through approximation, and instead the \phd is updated using all measurements. When the \phd has a distribution mixture representation, e.g., a Gaussian mixture, then the updated \phd is obtained by updating each Gaussian component in the \phd mixture with each measurement. In other extended object \mtt filters, the number of cell associations can be reduced, either by computing association probabilities or by finding the best associations. Using association probabilities means that for each measurement-object-pair we compute the probability that the object is the origin of the measurement, and the probabilities are then used in the \mtt update. \jpda association probabilities are used in \cite{SchusterRW:2015,VivoneB:2016}. Alternatively, one can find the best association assignment(s) by optimising a cost function that is related to the \mtt predicted likelihood. The single best assignment can be found using the auction algorithm \cite{Bertsekas:1988}, and the $M$ top ranked assignments can be found using Murty's algorithm \cite{Murty:1968}. Finding optimal assignments is used in the implementations of the extended object conjugate priors \cite{BeardRGVVS:2015,BeardRGVVS:2015journal,GranstromFS:2016fusion,GranstromFS:2016submitted}.
In \cite{streit2016jpda}, a JPDAF intensity filter that estimates
an intensity function for each extended object is developed.

\section{ Metrics and Performance Evaluation}

Regardless of the target type---point, extended, group or multi-path---it is important to be able to evaluate the performance of a target tracking algorithm, such that the estimates can be compared to the ground truth and different tracking algorithms can be compared to each other. For point targets the root means squared error (\rmse) is a standard metric. For Gaussian assumed state estimates, the normalised estimation error squared (\nees) is another standard performance measure, that incorporates also the estimated covariance matrix and evaluates whether or not the estimate is consistent. 

In extended object tracking the tracker output incorporates extent information, and because of this it is not trivial to answer the question: what is the distance between the estimate and the ground truth. It may seem tempting to use the \rmse, however, doing so is not always straightforward as the following two examples illustrate.
\begin{enumerate}
	\item Consider an extended object with an assumed rectangular shape and state vector
	\begin{subequations}
	\begin{align}
		\sx=\left[\cx, \cy,\ell_1,\ell_2,\varphi\right]^{\tp}
	\end{align}
	where $\cx,\cy$ is the position, $\ell_1$ and $\ell_2$ is the dimensions of the two sides, and $\varphi$ is the orientation of the side with length $\ell_1$ (and does not specify the moving direction of the object). For this state vector the two estimates
	\begin{align}
		\hat{\sx}^{(1)} & = \left[\cx, \cy,\ell_1,\ell_2,\varphi\right]^{\tp},\\
		\hat{\sx}^{(2)} & = \left[\cx, \cy,\ell_2,\ell_1,\varphi+0.5\pi\right]^{\tp},
	\end{align}
	\end{subequations}
	where width and length are switched in $\hat{\sx}^{(2)}$, define exactly the same shape in the Cartesian surveillance space, however, the \rmse errors would not be the same for the two estimates, which clearly violates intuition.
	
	\item In the random matrix model the extended object state is a combination of a vector and a matrix. The estimated vector can be compared to the ground truth using the Euclidean norm. The matrix generalisation of the Euclidean norm for vectors is the Frobenius norm, and this norm can be used to compare the estimated matrix to the ground truth. In \cite{LundquistGO:2013} it is suggested to use a weighted summation to combine the vector norm and the matrix norm, however, this leads to a problem whereby one has to determine the weights in the summation.
\end{enumerate}

In some works, see e.g., \cite{FeldmannFK:2011}, the extended object state is broken down into specific properties, such as position, velocity, orientation, extent area, and extent dimensions\footnote{For example, the semi-axes of an ellipse or the two sides of a rectangle}. This facilitates easy interpretation of the results, however, by this means it is no longer possible to rank estimates from different trackers using a single score. Furthermore, standard multi-object metrics, such as the optimal sub-pattern assignment (\ospa) metric \cite{SchuhmacherVV:2008} and the generalized \ospa (\textsc{g-ospa}) \cite{RahmathullahGFS:GOSPA} build upon single object metrics that give a single output. In other words, breaking down the extended object state into different properties does not facilitate multi-object performance evaluation.

A widely-used measure in computer vision is the so-called Intersection-over-Union (IoU), which is defined as the area of the intersection between the estimated shape and the ground truth shape, divided by the area of the union of the two shapes. In the extended object tracking context, IoU has been used, e.g., for rectangular and elliptical extended objects \cite{GranstromLO:2011}. For axis-aligned rectangles the IoU is simple to compute, however, for other shapes, or rectangles that are not axis-aligned, computing the IoU can be cumbersome. Furthermore, the IoU is always zero for non-overlapping objects\footnote{For two non-overlapping shapes, the intersection is empty, and thus the area of the intersection is zero.}, meaning that the error measure is the same regardless of how big the translational error is. This goes against intuition, which tells us that the larger the translational difference is between two shapes, the larger the error should be.

One work in this direction is \cite{MFI16_Yang}, which addresses performance metrics for elliptically shaped extended objects. By comparing several metrics and measures, the so-called Gaussian Wasserstein distance is identified as the most appropriate one. The Gaussian Wasserstein distance is available in closed-form, gives intuitive results, and is a true metric. Unfortunately, for general shapes, no analytic formulas for the Wasserstein distance exist, meaning that the Wasserstein metric is currently only suitable for objects with elliptic extents.

For star-convex shapes, the work \cite{Sun2016}  discusses a modified Hausdorff
distance that fully incorporates different shape
parametrisations.

While the existing extended object performance measures for non-elliptic shapes, such as decomposition into specific properties and IoU, have their applications, there is still a lot of work needed to specify a general single extended object performance evaluation criterion. However, for multiple extended object performance evaluation, given a chosen single object metric, the standard performance measures such as \ospa \cite{SchuhmacherVV:2008} and \textsc{g-ospa} \cite{RahmathullahGFS:GOSPA} are directly applicable.

%%%%%%%%%%%%%%%%%%%%%%%%%%%%%%%%%%%%%%%%%%%%%%%%%%%%%%%%%%%%%%%%%%%%%%%%%%%%%%%%%%%%%%%%%%
%%%%%%%%%%%%%%%%%%%%%%%%%%%%%%%%%%%%%%%%%%%%%%%%%%%%%%%%%%%%%%%%%%%%%%%%%%%%%%%%%%%%%%%%%%
%%%%%%%%%%%%%%%%%%%%%%%%%%%%%%%%%%%%%%%%%%%%%%%%%%%%%%%%%%%%%%%%%%%%%%%%%%%%%%%%%%%%%%%%%%

\begin{figure*}[h]
	\centering
	\subfloat[]{\includegraphics[width=0.5\textwidth]{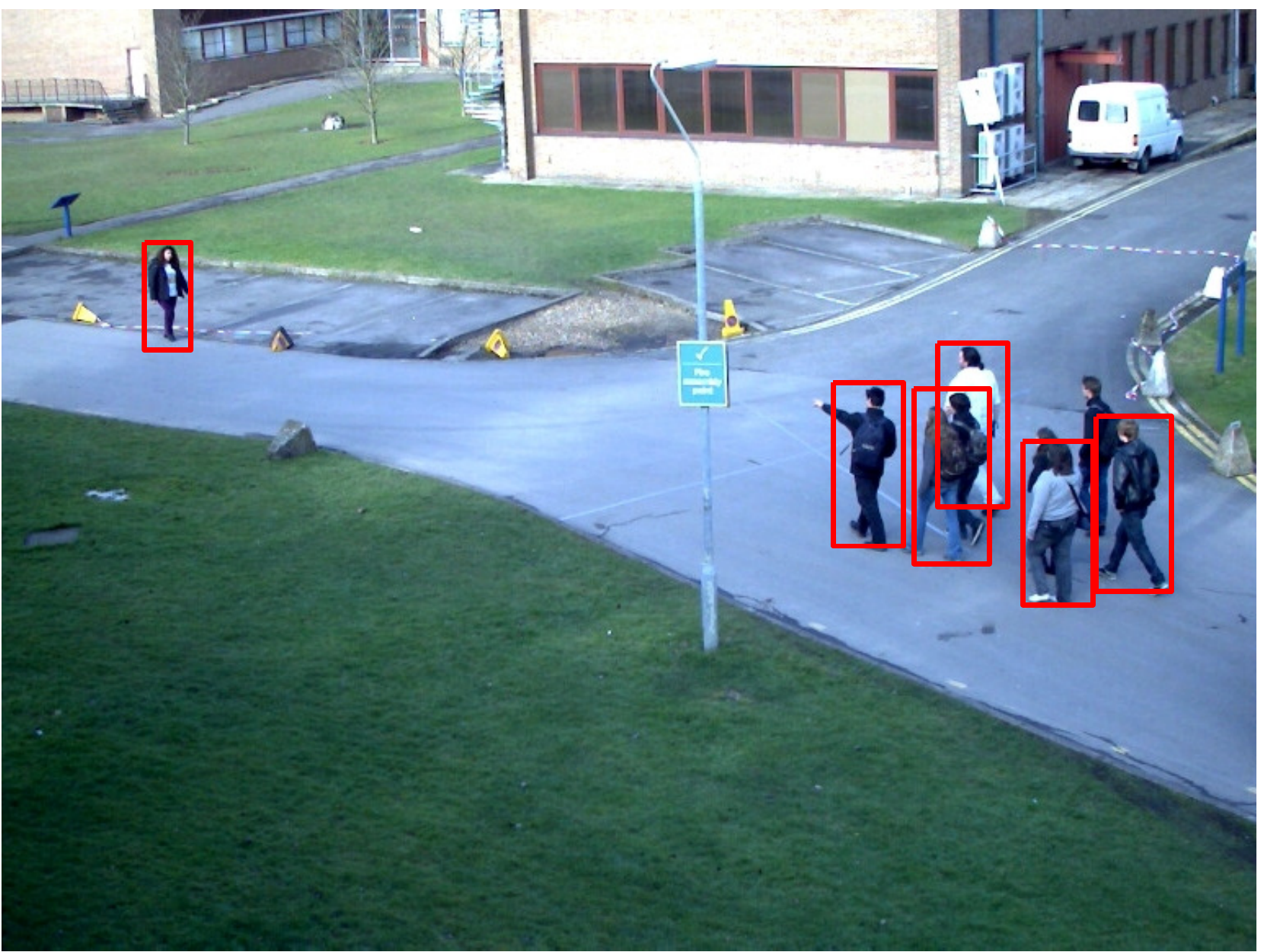}}
	\hfil
	\subfloat[]{\includegraphics[width=0.5\textwidth]{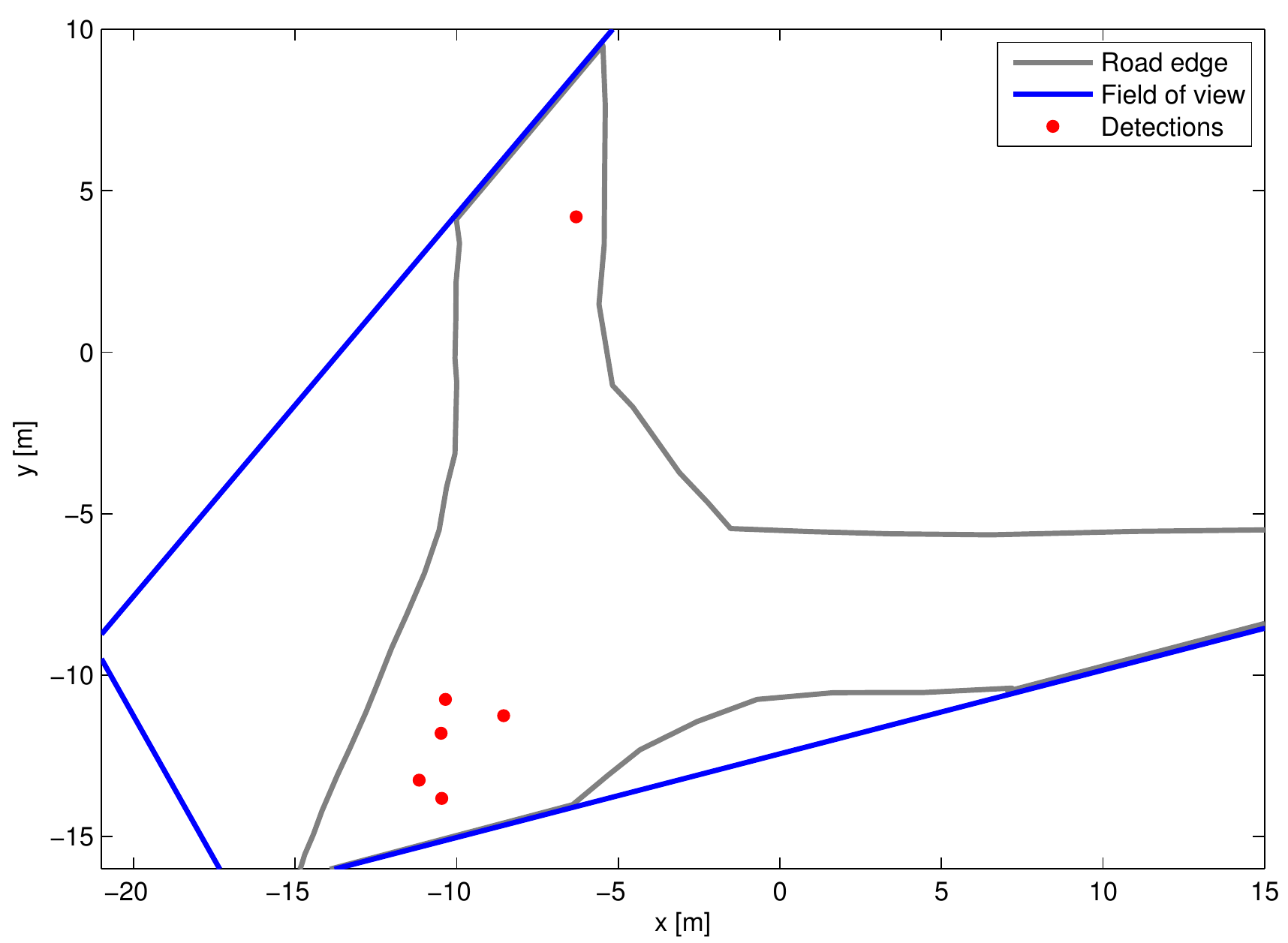}}
	
	\subfloat[]{\includegraphics[width=0.5\textwidth]{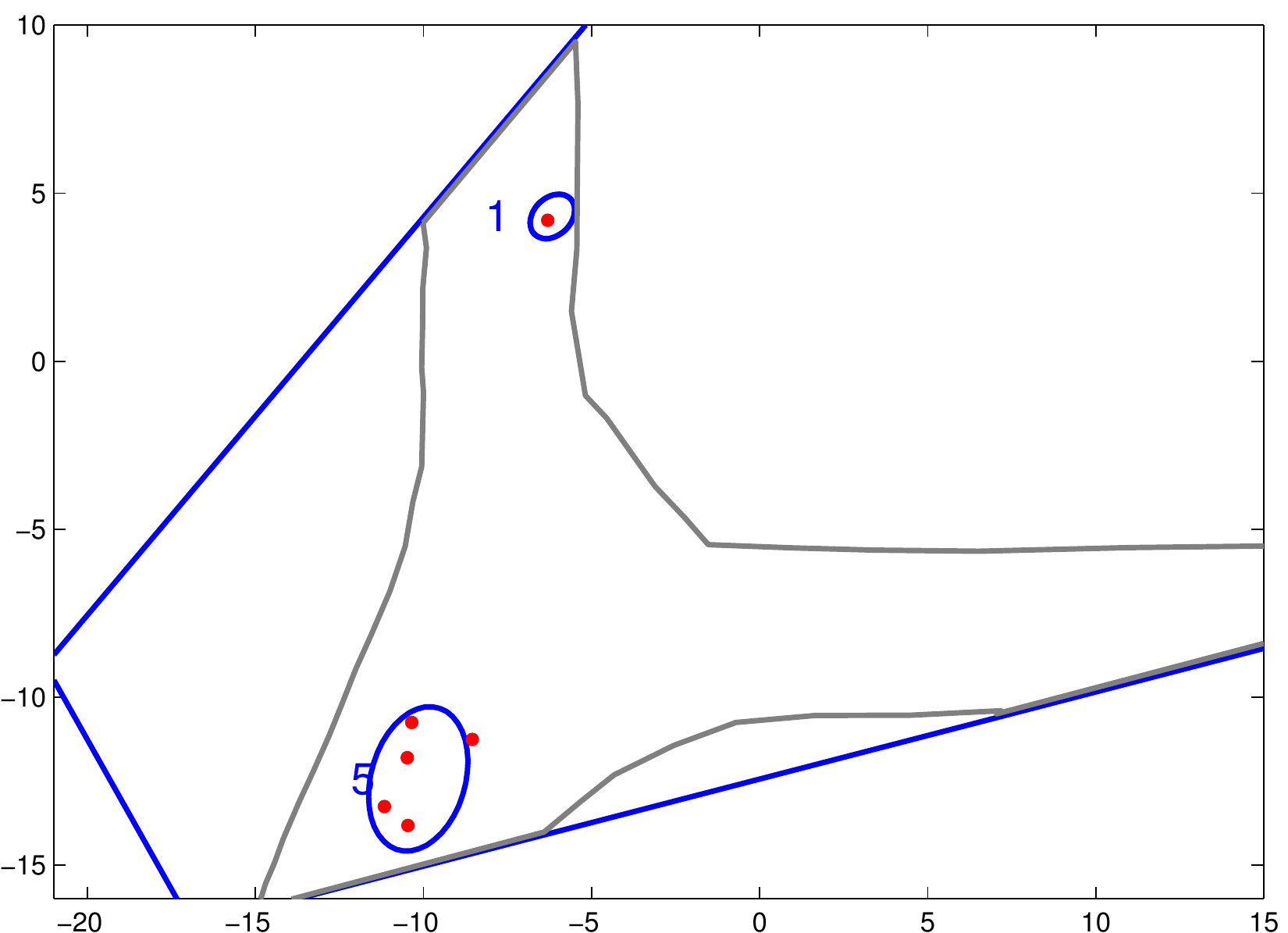}}
	\hfil
	\subfloat[]{\includegraphics[width=0.5\textwidth]{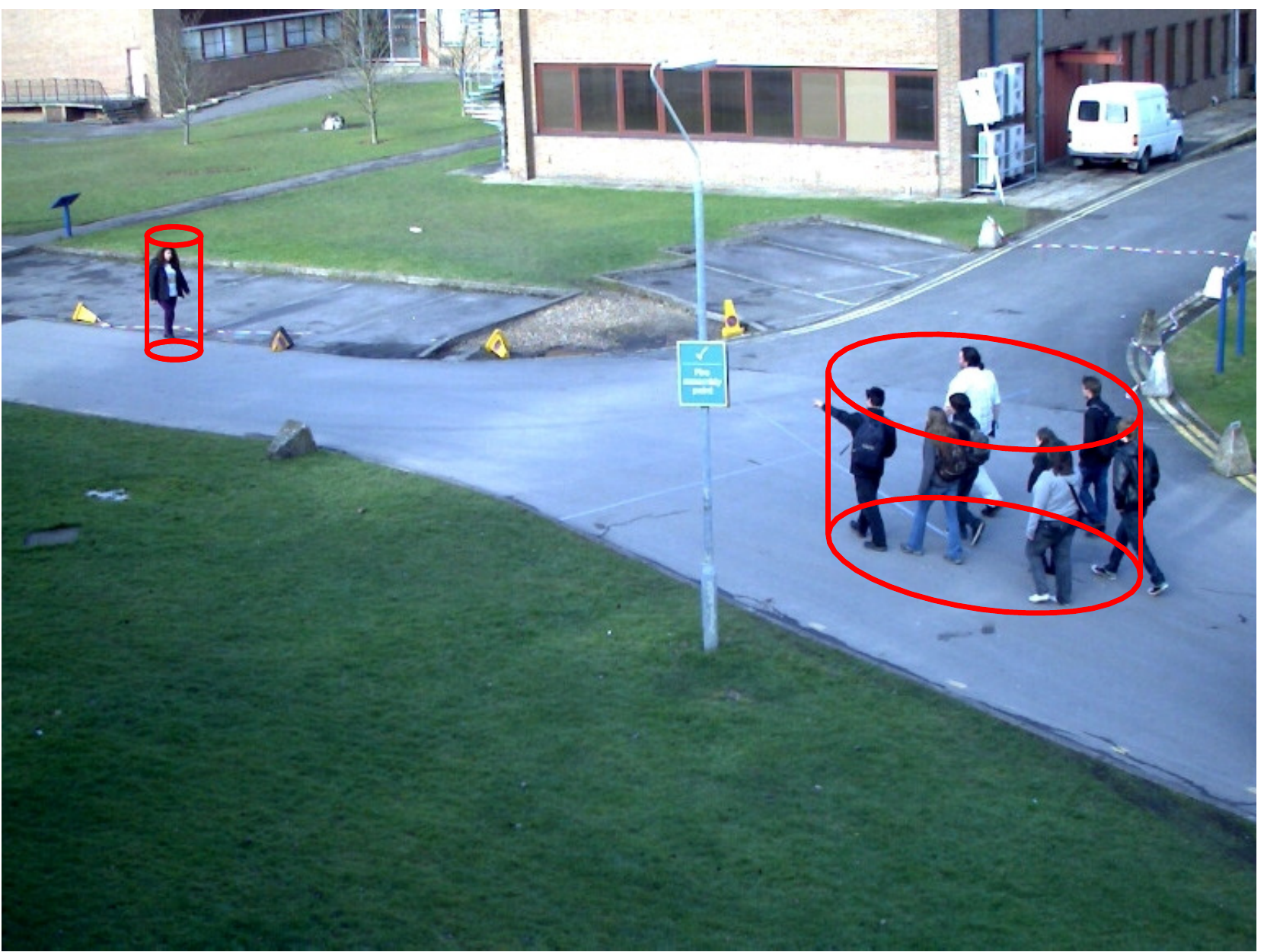}}
	\caption{Example application: tracking groups of pedestrians in video data. a) Input image with pedestrian measurements in red. b) Foot-print of measurements projected onto ground plane. c) Group tracking output, numbers are estimated Poisson rates. d) Output projected into input image, and visualised as elliptic cylinders.}
	\label{fig:group_target_example}
\end{figure*}

\begin{table}[b]
	\centering
	\caption{Experiments with different sensor types}
	\label{tab:WorkWithExperimentalData}
		\begin{tabular}{| l | l |}
			\hline
			\textbf{Sensor} & \textbf{References} \\
			\hline
			Automotive Radar & \tblcite{BuhrenY:2006,LundquistOG:2010,HammarstrandLS:2012,KnillSD:2016,ScheelKRD:2016,SchusterRW:JAIF,HammarstrandSSS:2012,GunnarssonSDB:2007} \\
			\hline
			Camera  & \tblcite{EdmanAGG:2013,DaveyWV:2013,RisticS:2013,RisticVVF:2013,BroidaCC:1990,BroidaC:1991} \\
			\hline
			 \textsc{gmti} radar  & \tblcite{PollardPR:2010} \\
			\hline
			Imaging Sonar & \tblcite{KroutKOH:2012} \\
			\hline
			\lidar  & \tblcite{BeardRGVVS:2015journal,GranstromLO:2011,GranstromLO:2012,GranstromO:2012a,ReuterD:2011,WienekeK:2012,GranstromL:2013,GranstromRMS:2014,ScheelGMRD:2014,FortinLN:2014,SchutzDD:2014,FortinLN:2013,OrgunerLG:2011,SpinelloTS:2010,NavarroSermentMH:2010,LuberTA:2011,PetrovUMGSWK:2012} \\
			\hline
			Marine Radar  & \tblcite{ErrastiAlcalaB:2014,GranstromNBLS:2014,GranstromNBLS:2015_TGARS,SchusterRW:2015,VivoneBGNC:2015_ConvMeas,VivoneBGW:2016,VivoneBGNC:2016_JAIF} \\
			 \hline
			\rgbd  & \tblcite{Faion2015a,MFI15_Faion,MFI12_Baum,Baum2012,Baum2013_thesis} \\
			\hline
			Through wall radar  & \tblcite{GennarelliVBSA:2015} \\
			\hline
			\textsc{ugs}, group tracking  & \tblcite{DamarlaK:2013} \\
			\hline
		\end{tabular}
\end{table}

\section{Extended object tracking applications}
\label{sec:ETTapplications}

Extended object tracking algorithms have been applied in many different scenarios and have been evaluated using data from many different sensors such as \lidar, camera, radar, \rgb-depth (\rgbd) sensors, and unattended ground sensors (\textsc{ugs}). A list of references that contain experiments with real data is given in Table~\ref{tab:WorkWithExperimentalData}. In this section we will present four example applications:
\begin{itemize}
	\item Tracking groups of pedestrians using a camera overlooking a footpath.
	\item Tracking marine vessels using X-band radar.
	\item Tracking cars using a \lidar mounted in the grille of an autonomous vehicle. 
	\item Tracking objects with complex shapes using an \rgbd sensor.
\end{itemize}
These four examples are complementary in the sense that they illustrate different aspects of extended object tracking: different sensor modalities; the applicability of extended object methods to group object tracking; object shapes of different complexities; and tracking in crowded scenarios with occlusions.

%%*************************************************************************
%%*************************************************************************
%%*************************************************************************
%%*************************************************************************
%%*************************************************************************

\subsection{Tracking groups of pedestrians using camera}
\label{sec:PedGroupTracking}
Automatic crowd surveillance is a complex task, and in scenes with a large number of persons it may be infeasible to track each person individually. In this case group object tracking using extended object \mtt methods is a viable alternative, as this does not require tracking and identification of each individual. In the example presented here camera data is used to track groups of pedestrians that walk along a footpath. The online available \textsc{pets} 2012 data set \cite{PetsData:2012} is used for evaluation. For each image in the dataset a pedestrian detector \cite{DollarTPB:2009,DollarBP:2010} is used, and the measurements are projected onto the ground plane using the camera parameters. 

In this data the groups of pedestrians are loosely constructed and typically do not have a detailed shape that remains constant over time. Therefore the groups can be assumed to be elliptically shaped, and the random matrix measurement model can be used \cite{Koch:2008}. The ground plane measurements are input into a \ggiw-\phd filter \cite{GranstromO:2012a,GranstromO:2012c}, and the object extractions are projected back into the camera image for visualization. The \ggiw-\phd filter is based on the Poisson model for the number of measurements from each group, i.e., for each group a Poisson rate parameter is estimated. This estimated rate can be taken as an estimate of the number of persons in the group.

Example results are shown in Figure~\ref{fig:group_target_example}.\footnote{Video with tracking results: \texttt{https://youtu.be/jN-KXQqargE}} The results show that the estimated ellipses are a good approximation of the pedestrian groups. The estimated Poisson rates tend to underestimate the number of persons in the group. The reason for this is that in groups with many persons, some individuals tend to be occluded and therefore are not detected. The estimated Poisson rate is more accurate when interpreted as a lower bound for the number of persons in the group, instead of interpreted as a count of the number of persons in the group.

%%*************************************************************************
%%*************************************************************************
%%*************************************************************************
%%*************************************************************************
%%*************************************************************************

\begin{figure}[t]
	\centering
	\subfloat[]{\includegraphics[width=0.80\columnwidth]{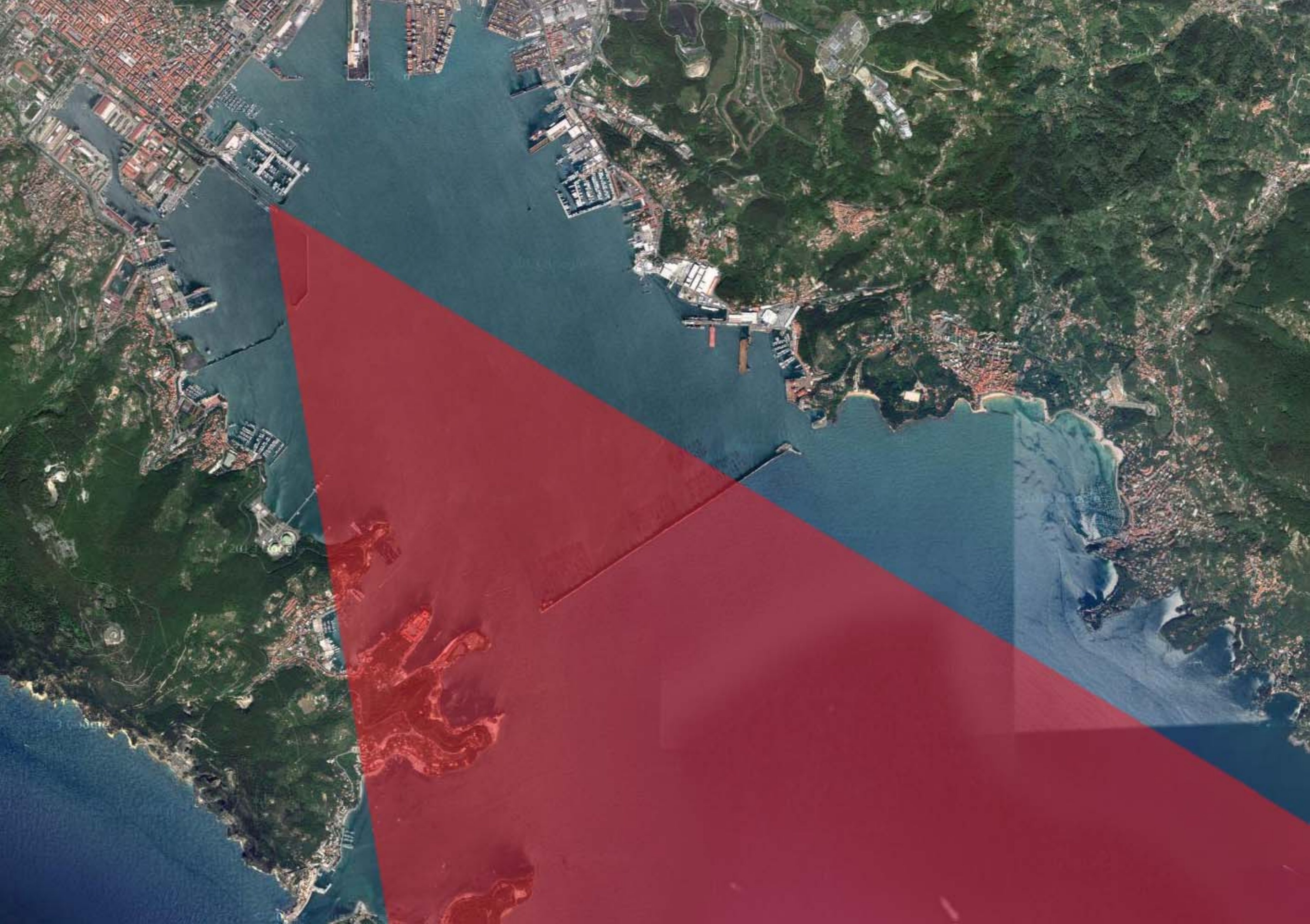}}
	\caption{Example appliation: tracking boats and ships using marine X-band radar. Aerial image of harbour, with sensor's field of view shown in red.}
	\label{fig:MarineTracking1}
\end{figure}

\begin{figure}[t]
	\centering
	\subfloat[]{\includegraphics[width=0.5\columnwidth]{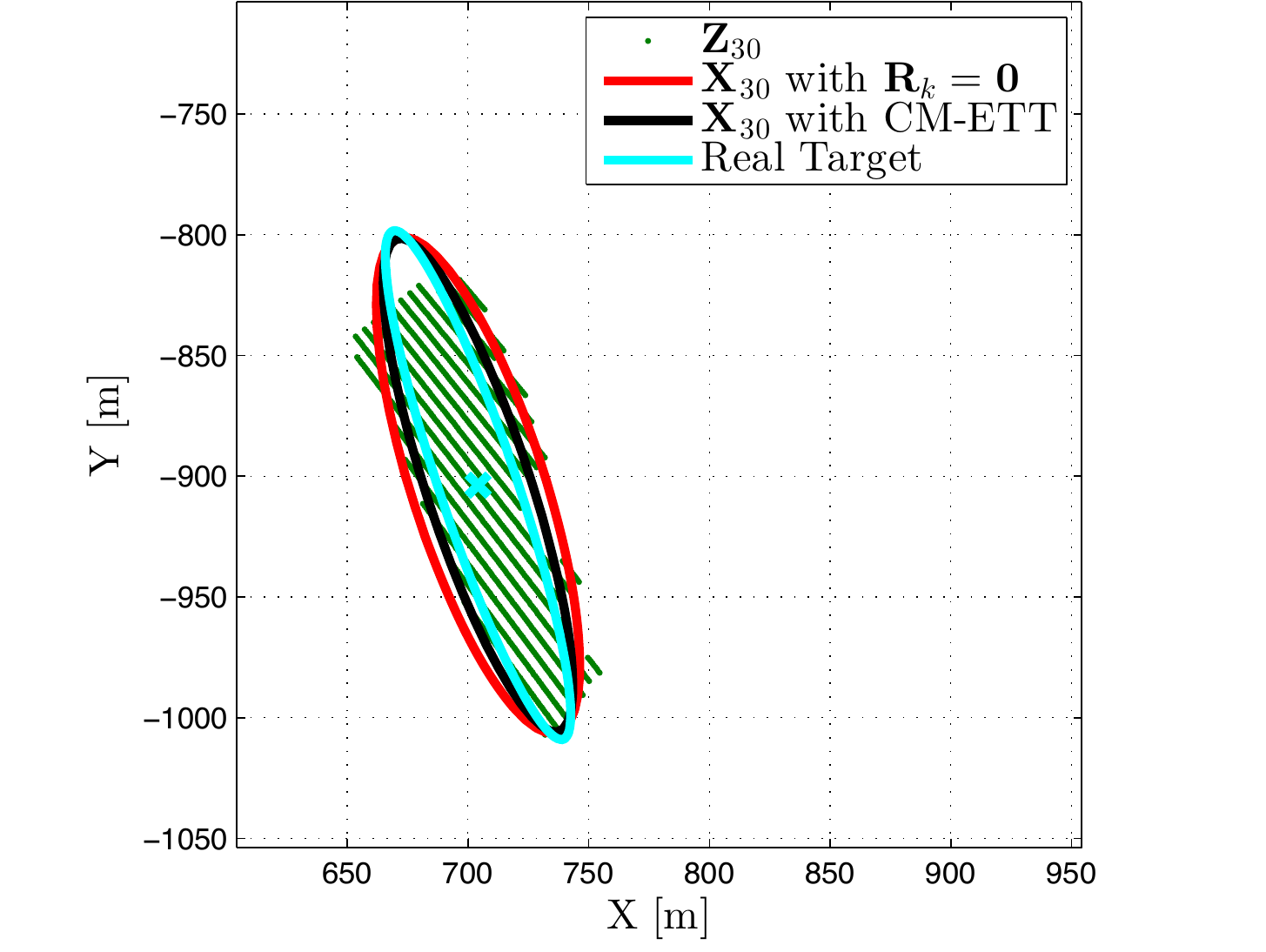}}
	\hfil
	\subfloat[]{\includegraphics[width=0.5\columnwidth]{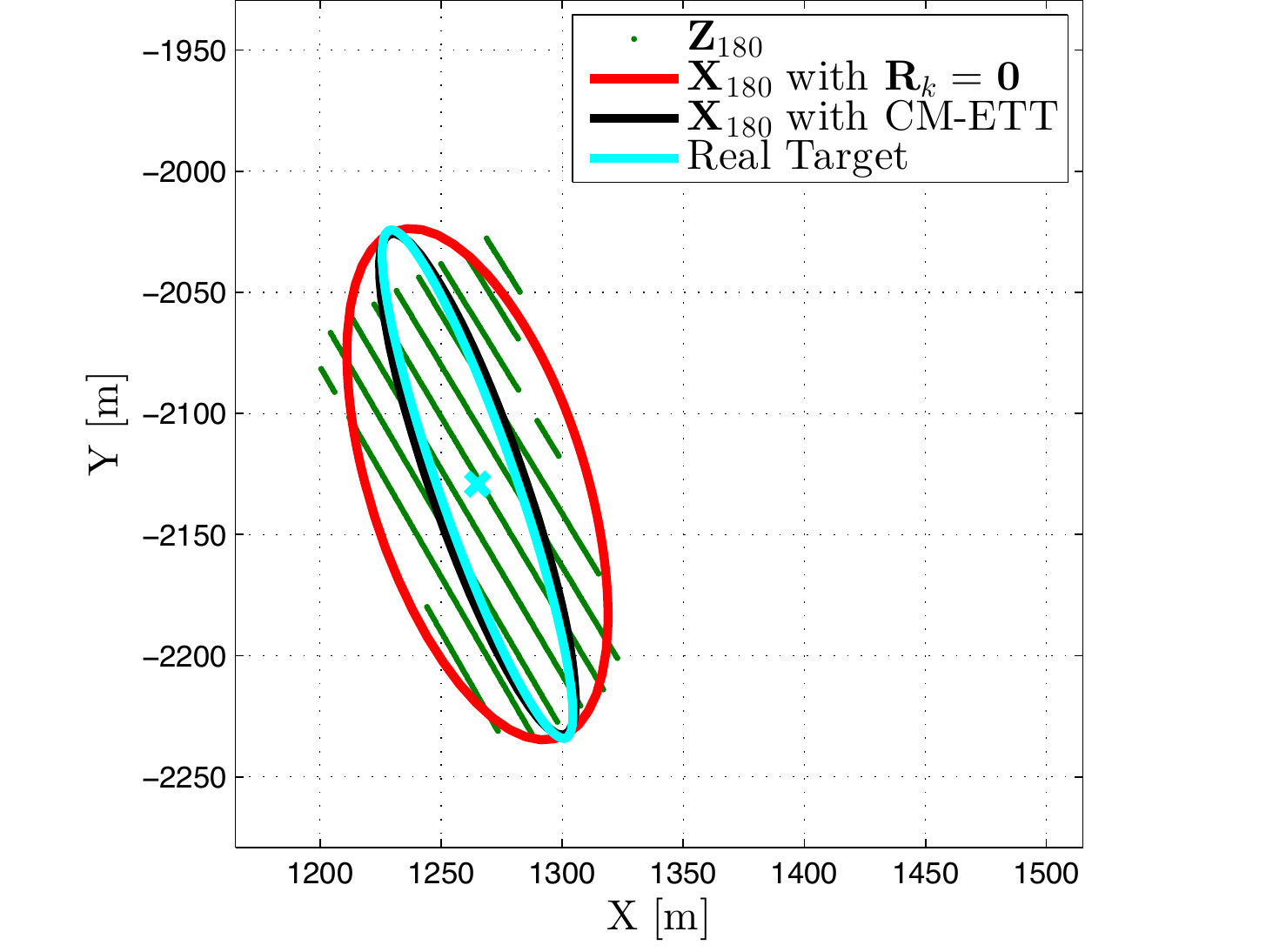}}
	\caption{Example appliation: tracking boats and ships using marine X-band radar. a) and b) Example detections (green dots), ground truth (teal ellipse), and two estimates (red and black ellipses). The the black ellipse is when the noise is correctly modelled, and the red ellipse shows the estimate when the noise is not modelled.}
	\label{fig:MarineTracking2}
\end{figure}

\subsection{Tracking marine vessels using X-band radar}
\label{sec:XbandRadarTracking}

Harbours are busy places where many vessels share the water, from small boats to large ships. To keep track of where all the vessels are, marine X-band radar can be used \cite{GranstromNBLS:2015_TGARS}. These sensor produce high-resolution data that allow the tracking algorithm to estimate the size of the vessel, further allowing the possibility to classify the tracked vessels using prior information about the size of different ships and boats. An example is given in Figure~\ref{fig:MarineTracking1}, where the field of view of the sensor is overlaid on an aerial image of a harbour.

The raw sensor data is pre-processed using a Constant False Alarm Rate (\textsc{cfar}) detector, producing polar detections (range and azimuth) \cite{GranstromNBLS:2015_TGARS}. Because boats and ships are best modelled in Cartesian coordinates, polar detections are converted to Cartesian coordinates \cite{GranstromNBLS:2015_TGARS}. The pre-processed data is suitable for use with the random matrix model, meaning that the shapes of the vessels are assumed to be ellipses. Typically neither boats, nor ships, are elliptically shaped, however, the major and minor axes of the estimated ellipses correspond to the length and width of the vessel. If measurement noise is modelled correctly low estimation errors can be achieved, however, if the noise is not modelled the size of the vessel is overestimated, especially in the cross-range dimension \cite{VivoneBGNC:2016_JAIF}. The significant difference between modelling the noise correctly, or not, is shown in Figure~\ref{fig:MarineTracking2}. If multiple radar sensors are used, the tracking results can be improved further \cite{VivoneBGW:2016fusion}. %VivoneGBW:MultiSensor

%%*************************************************************************
%%*************************************************************************
%%*************************************************************************
%%*************************************************************************
%%*************************************************************************

\subsection{Tracking cars using \lidar}
\label{sec:LidarCarTracking}
Autonomous active safety features are standard in many modern cars, and in both research and industry there is a considerable push towards fully driverless vehicles, see, e.g., \cite{KunzEtAl:2015}. For safe operation in dense scenes, such as inner city and other urban environments, an autonomous vehicle must be capable of keeping track of other objects, to avoid collisions. To this end, high resolution sensors such as \lidar and extended object tracking algorithms can be used.

The high angular resolution of \lidar sensors typically results in a large number of measurements for each object. Thus, if an extended object tracking filter is not used, preprocessing is necessary to update the object estimates. Such preprocessing commonly consists of segmentation and clustering \cite{Himmelsbach2010,Meissner2013c,Premebida.2005}, shape fitting \cite{Munz2011}, or feature extraction \cite{Nguyen.2005}. The drawback of using such algorithms is that they are heavily dependent on parametrization, and often suffer from over- or under-segmentation. Especially in scenarios in which the environment changes, or when there are different object types, it is very difficult to find appropriate parameters. Because the tracking builds upon the data that is input, any error during segmentation and clustering will manifest itself as a tracking error.

\begin{figure}
	\includegraphics[width=1.0\columnwidth]{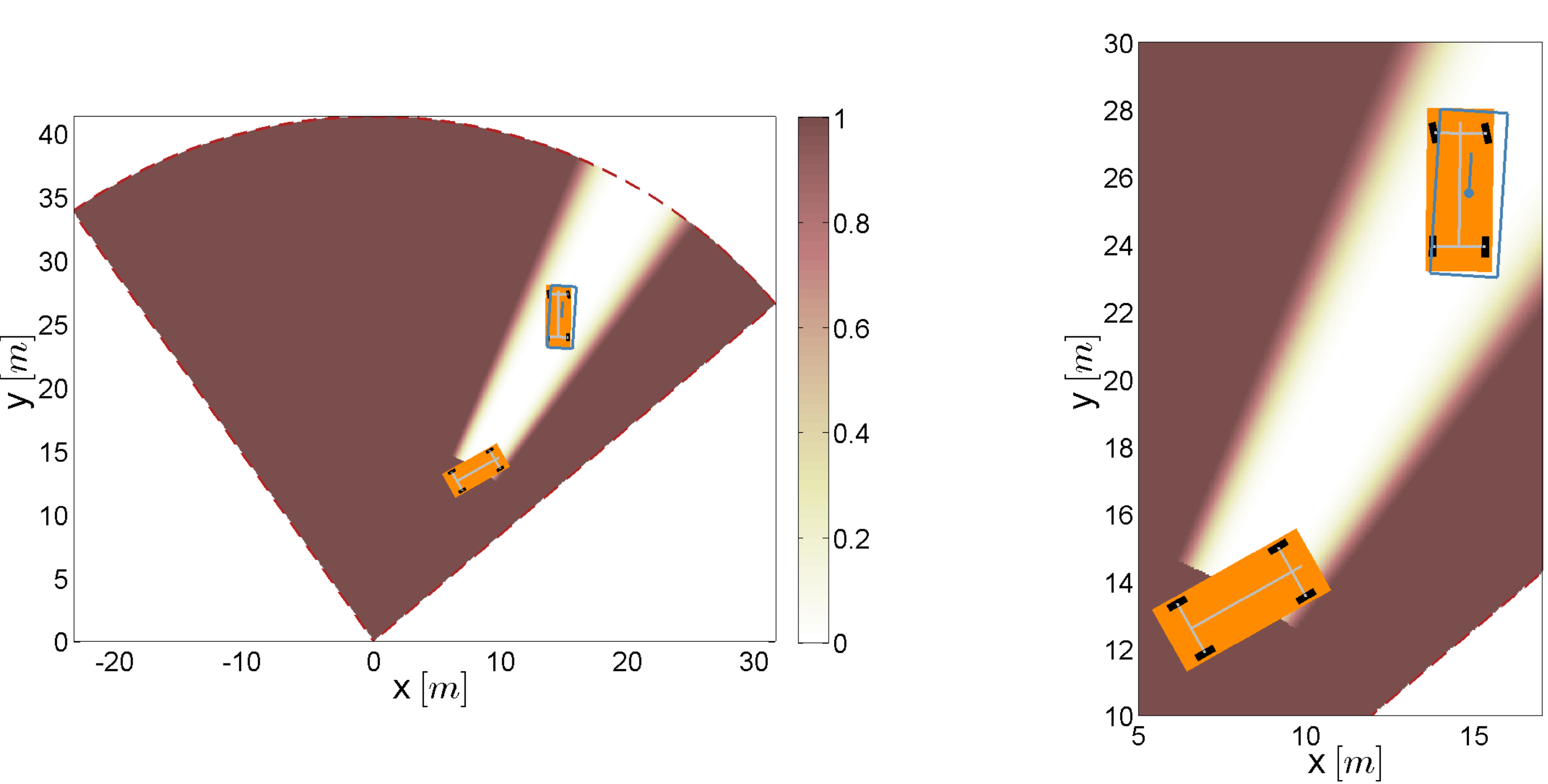}
	\caption{Occlusion example. The sensor is located in the origin; darker color means higher probability of measurement; estimates in orange, ground truth in blue. Thanks to the use of an occlusion model the occluded car can be tracked with high accuracy while it traverses an area where it cannot be detected.}
	\label{fig:Roundabout_Output_Example_Occlusion}
\end{figure}

In this section we will present experimental results where \lidar sensors and an extended object \phd filter have been used to track cars; the results presented here are a subset of the results presented in \cite{GranstromRMS:2014}. The \lidar sensor is assumed to be mounted in the grille of the ego vehicle, and the cars are assumed to be rectangular, with unknown length and width. The measurement modelling that was used is shown in Figure~\ref{fig:Ex_Measurement_Likelihood}. The tracking problem is cast as a multiple model problem, and a multiple model \phd filter is used to track multiple cars. A full description of the tracking algorithm can be found in \cite{GranstromRMS:2014}. When there are multiple cars in the sensor's field of view the cars may occlude each other, either partially of fully. To avoid loosing track of cars that are occluded a non-homogeneous probability of measurement can be used. This is illustrated in Figure~\ref{fig:Roundabout_Output_Example_Occlusion}. Similar approaches to occlusion modelling are taken in \cite{GranstromLO:2012,GranstromO:2012a,WyffelsC:2015,ReuterD:2011}.

\begin{figure}
	\includegraphics[width=1.0\columnwidth]{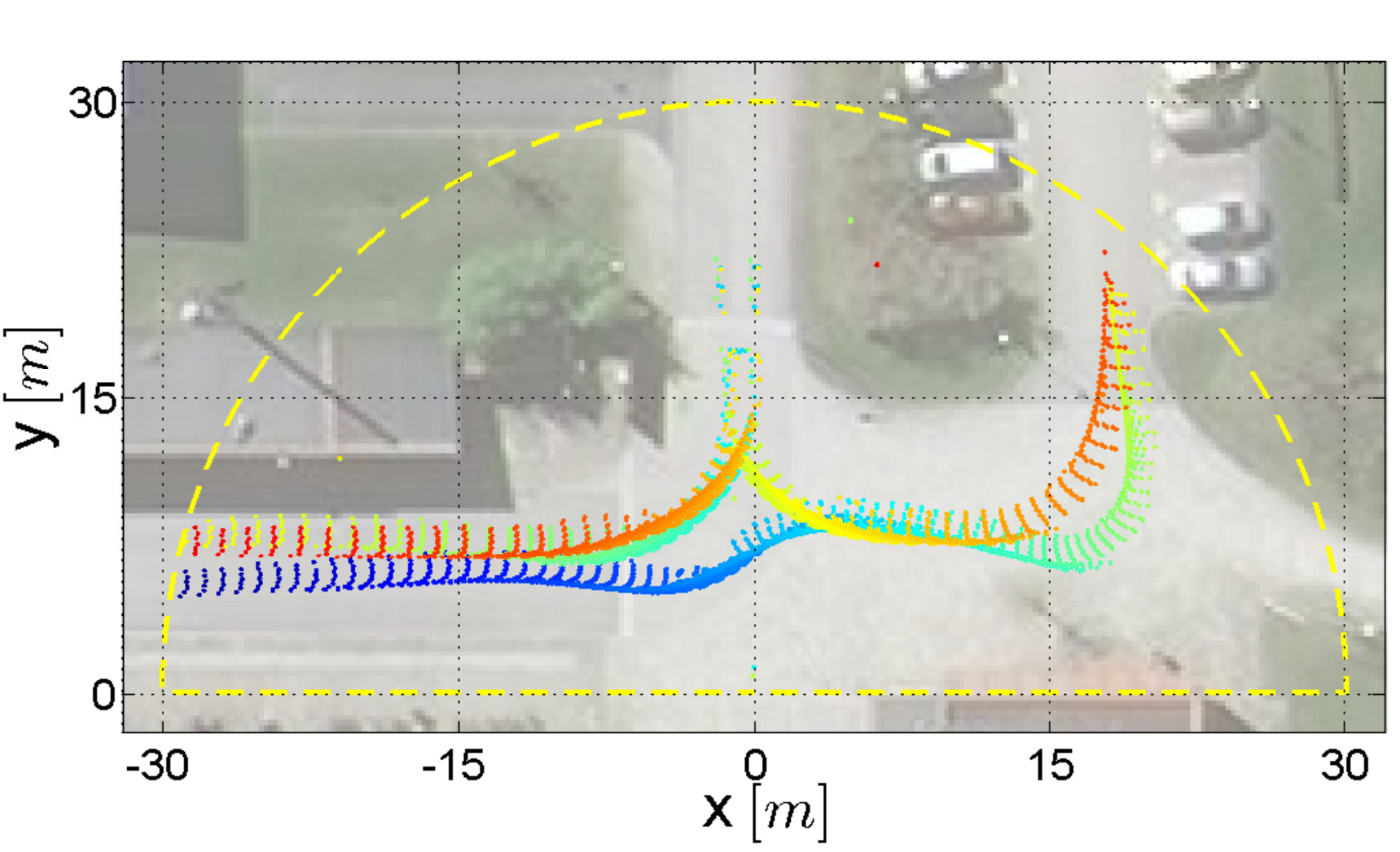}
	\includegraphics[width=01.0\columnwidth]{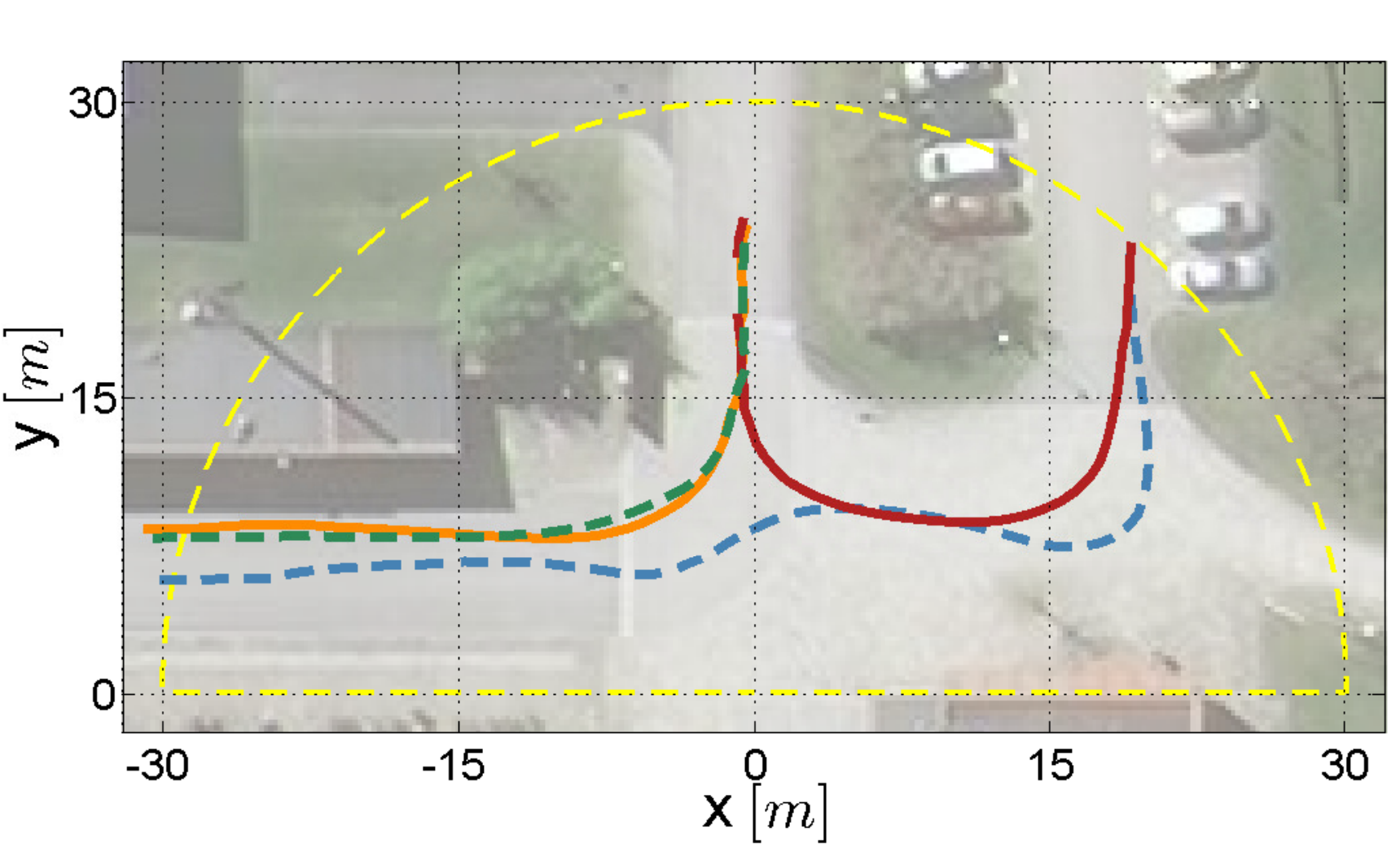}
	\caption{Results from scenario with four cars. Top: sensor data, colorcoded according to time. Bottom: Estimated positions.}
	\label{fig:MultiTarget}
\end{figure}

Experimental results in \cite{GranstromRMS:2014} show that the lateral position of the tracked cars can be estimated with an average error of less than 5cm, while the average longitudinal position error is slightly larger, around 10 to 30 cm for different datasets. The shape parameters are estimated with an average error around 2cm for the width, and around 20 cm for the length. The increased error in object length is due to the limited observability of the object length due to the aspect angle. Example detections and tracking results for a scenario with four cars is given in Figure~\ref{fig:MultiTarget}, snapshots of this data are also shown in Figure~\ref{fig:CarLidar}.

%%*************************************************************************
%%*************************************************************************
%%*************************************************************************
%%*************************************************************************
%%*************************************************************************

\subsection{Tracking complex shapes using \rgbd sensor}
\begin{figure}
\centering
\subfloat[Depth image.]{
\frame{
\includegraphics[trim=150 20 120 150,clip,width=4cm]{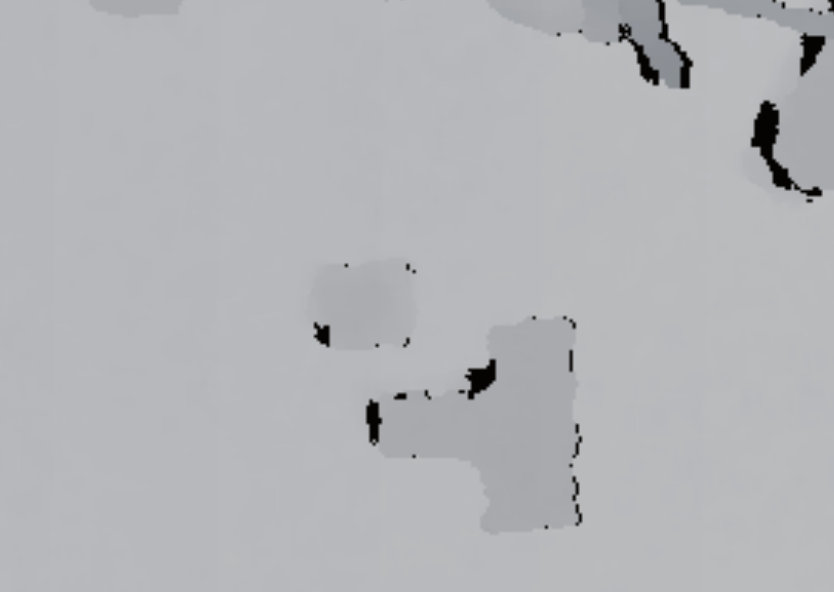}
}
}
\subfloat[\textsc{rgb} image.]{
\frame{
\includegraphics[trim=150 20 120 150,clip,width=4cm]{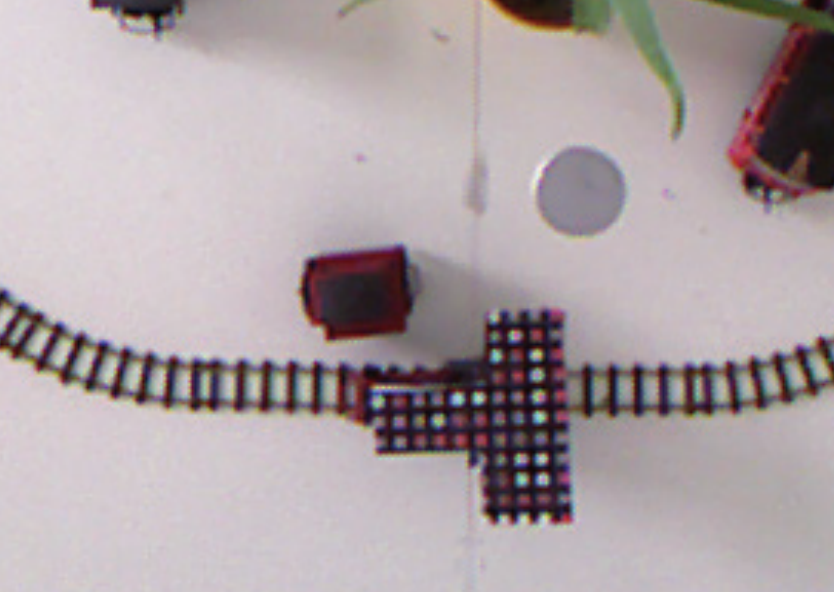}
}
}
 
 \subfloat[Measurements.\label{fig:setting2_3}]{
 \centering
\frame{
\includegraphics[trim=150 20 120 150,clip,width=4cm]{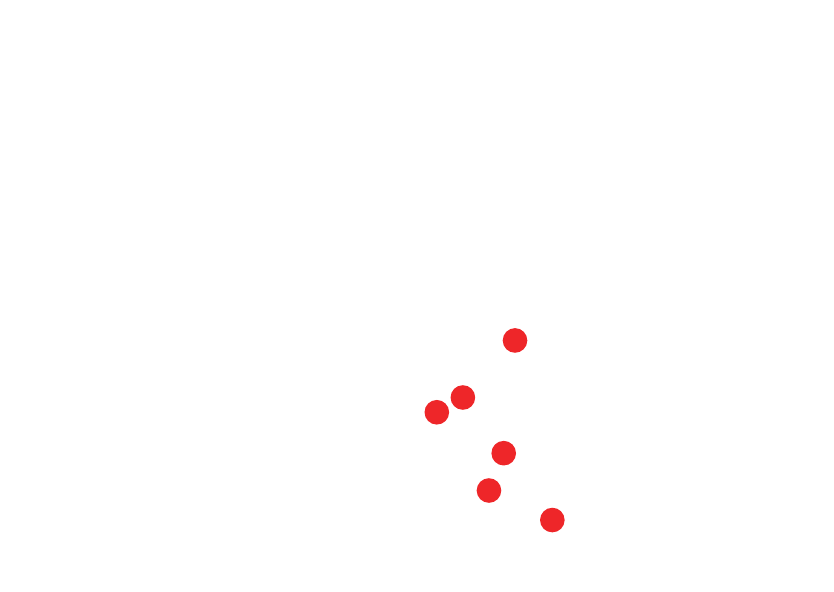}
}
}
\caption{Tracking a railway vehicle using a RGBD camera from a bird's eye view  \cite{Baum2013_thesis}.\label{fig:gmti_setting}}
\end{figure}

\begin{figure}
\centering
\subfloat[]{\hspace{-0.1cm}
\includegraphics[width=2.7cm]{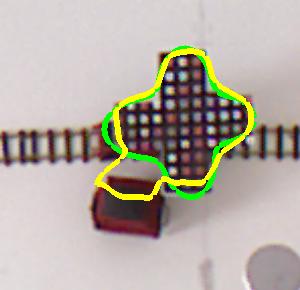}
}
\subfloat[]{\hspace{-0.1cm}
\includegraphics[width=2.7cm]{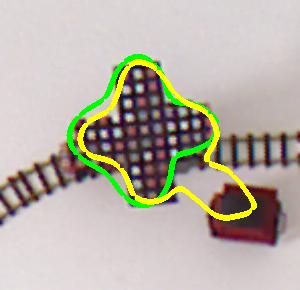}
}

\caption{Result for ``+''--shaped vehicle:  RHM (green) vs.  active contour model using depth images (yellow)  \cite{Baum2013_thesis}.\label{fig:experiment_C-shape_RGB}}
\end{figure}

In  this subsection we present an  experimental setup where complex object shapes are estimated using \rgbd sensor data. This experiment has been published first in \cite{MFI12_Baum,Baum2012,Baum2013_thesis}.
Specifically, a moving miniature railway vehicle is to be tracked from a bird's  eye view with the help of a \rgbd camera. 
An optical flow algorithm  determines the velocity of each image point in both the RGB and depth image sequences.
Based on a threshold on the velocity, we obtain measurements, i.e., points classified as ``moving'', that originate from the moving object.
In this manner,  a varying number of noisy measurements from the object's surface is received at each frame, see Figure~\ref{fig:gmti_setting} for an example frame.
Due to the noisy images and inaccuracy of the optical flow algorithm, the measurements are noisy and do not completely fill the object surface.
In fact, this is a typical  extended object tracking problem where measurements come from  a two-dimensional shape in two-dimensional space.
Figure~\ref{fig:experiment_C-shape_RGB} shows example results with an implementation of the star-convex random hypersurface approach as discussed in Section~ \ref{sec:ShapeEstimation}.
Also, Figure~\ref{fig:experiment_C-shape_RGB} shows the result obtained from an active contour (snake) algorithm  \cite{Kass1988}, which is a standard algorithm in computer vision. In general, an active  contour  model works with intensity/RGB images and not with point measurements. It  calculates a contour  by  minimizing an energy function \cite{Kass1988} that is  composed  of an external force  for pushing the contour to image features  and an internal force for regularization.
In this scenario, active contours are applied to the depth image and hence, can be unreliable in case the vehicle passes objects with similar depth, see Figure~\ref{fig:experiment_C-shape_RGB}.

\begin{figure}
\centering
\subfloat[]{\hspace{-0.1cm}
\includegraphics[width=2.7cm]{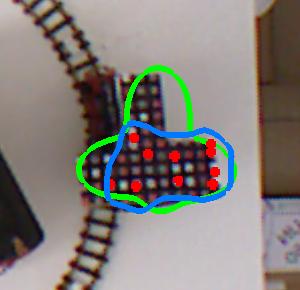}
}
\subfloat[]{\hspace{-0.1cm}
\includegraphics[width=2.7cm]{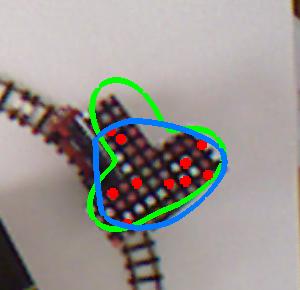}
}
\caption{Results for a  ``T''--shaped vehicle:  RHM (green) vs.  active contour model using (smoothed) point measurements (blue) \cite{Baum2013_thesis}. }
\label{fig:experiment_T_shape_10}
\end{figure}

Alternatively,  active contours can be applied to a ``smoothed'' version of the point measurements: the measurements are interpreted as an intensity image by placing  a Gaussian kernel function  at each measurement location.
As indicated by Figure~\ref{fig:experiment_T_shape_10}, active contours then aim at determining an enclosing curve of the point measurements in each frame. 
As the vehicle's surface is not covered completely by the measurements in a single frame, active contours do not give a reasonable shape estimate. Active contours are not capable of systematically accumulating individual point measurements over time -- without this capability no reasonable shape estimate can be expected.

%%%%%%%%%%%%%%%%%%%%%%%%%%%%%%%%%%%%%%%%%%%%%%%%%%%%%%%%%%%%%%%%%%%%%%%%%%%%%%%%%%%%%%%%%%
%%%%%%%%%%%%%%%%%%%%%%%%%%%%%%%%%%%%%%%%%%%%%%%%%%%%%%%%%%%%%%%%%%%%%%%%%%%%%%%%%%%%%%%%%%
%%%%%%%%%%%%%%%%%%%%%%%%%%%%%%%%%%%%%%%%%%%%%%%%%%%%%%%%%%%%%%%%%%%%%%%%%%%%%%%%%%%%%%%%%%

\section{Summary and concluding remarks}
\label{sec:Conclusions}

In this article we gave an introduction to extended object tracking, a comprehensive up-to-date overview of state-of-the-art research, and illustrated the methods using several different sensors and object types. Increasing sensor resolutions mean that there will be an increasing number of scenarios in which extended object methods can be applied. It is possible to cluster/segment the data in pre-processing and then apply standard point object methods, however this requires careful parameter tuning, thereby increasing the risk for errors. Extended object tracking, on the other hand, uses Bayesian models for the multiple measurements per object, meaning that the tracking performance is much less dependent on clustering/segmentation.

During the last 10 years an impressive number of new methods and applications have appeared in the literature, covering different approaches to extent modelling and multiple object tracking. This trend that can be expected to continue, as there are many open questions to solve, and improvements can be made. Due to the high non-linearity and high dimensionality of the problem, estimation of arbitrary shapes is still very much challenging. There is a need for performance bounds for extended object tracking methods: for a given shape model, how many measurements are required in order for the estimation algorithm to converge to an estimate with small error? Performance bounds may help in answering the question of which shape complexity is suitable when modelling the object. Naturally, in most applications one is interested in a shape description that is as precise as possible.

For arbitrary object shapes, the determination of suitable performance metrics for the evaluation of the shape estimate is still an open research question. Further, existing works on extended object tracking focus on single sensor systems (or perhaps systems with several very similar sensors). However, the fusion of complementary sensors like camera and \lidar in an extended object tracking algorithm raises new challenges due to the different measurement principles and perception capabilities.

\section{Acknowledgements}

The authors would like to thank the anonymous reviewers for their insightful comments.



\bibliographystyle{IEEEtranS}
\bibliography{ListOfReferences}

		\vspace{0cm}
		\begin{IEEEbiography}[{\includegraphics[width=1in,height=1.25in,clip,keepaspectratio]{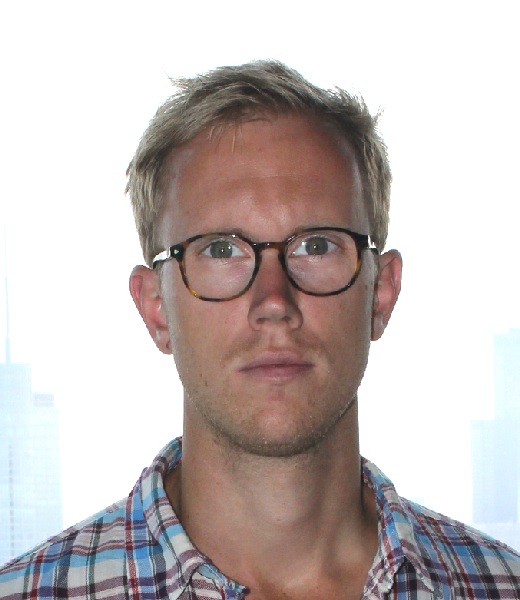}}]{Karl Granstr\"{o}m} (M'08) is a postdoctoral research fellow at the Department of Signals and Systems, Chalmers University of Technology, Gothenburg, Sweden. He received the MSc degree in Applied Physics and Electrical Engineering in May 2008, and the PhD degree in Automatic Control in November 2012, both from Link\"{o}ping University, Sweden. He previously held postdoctoral positions at the Department of Electrical and Computer Engineering at University of Connecticut, USA, from September 2014 to August 2015, and at the Department of Electrical Engineering of Link\"{o}ping University from December 2012 to August 2014. His research interests include estimation theory, multiple model estimation, sensor fusion and target tracking, especially for extended targets. He has received paper awards at the Fusion 2011 and Fusion 2012 conferences.
			\end{IEEEbiography}
			
			\begin{IEEEbiography}[{\includegraphics[width=1in,height=1.25in,clip,keepaspectratio]{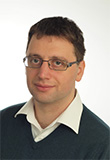}}]{Marcus Baum} is Juniorprofessor (Assistant Professor) at the University of Goettingen, Germany. He received the Diploma degree in computer science from the University of Karlsruhe (TH), Germany, in 2007, and graduated as Dr.-Ing. (Doctor of Engineering) at the Karlsruhe Institute of Technology (KIT), Germany, in 2013. From 2013 to 2014, he was postdoc and assistant research professor at the University of Connecticut, CT, USA. His research interests are in the field of data fusion, estimation, and tracking. Marcus Baum is associate administrative editor of the "Journal of Advances in Information Fusion (JAIF)" and serves as local arrangement chair of the "19th International Conference on Information Fusion (FUSION 2016)". He received the best student paper award at the FUSION 2011 conference.
			\end{IEEEbiography}
			 
			\begin{IEEEbiography}[{\includegraphics[width=1in,height=1.25in,clip,keepaspectratio]{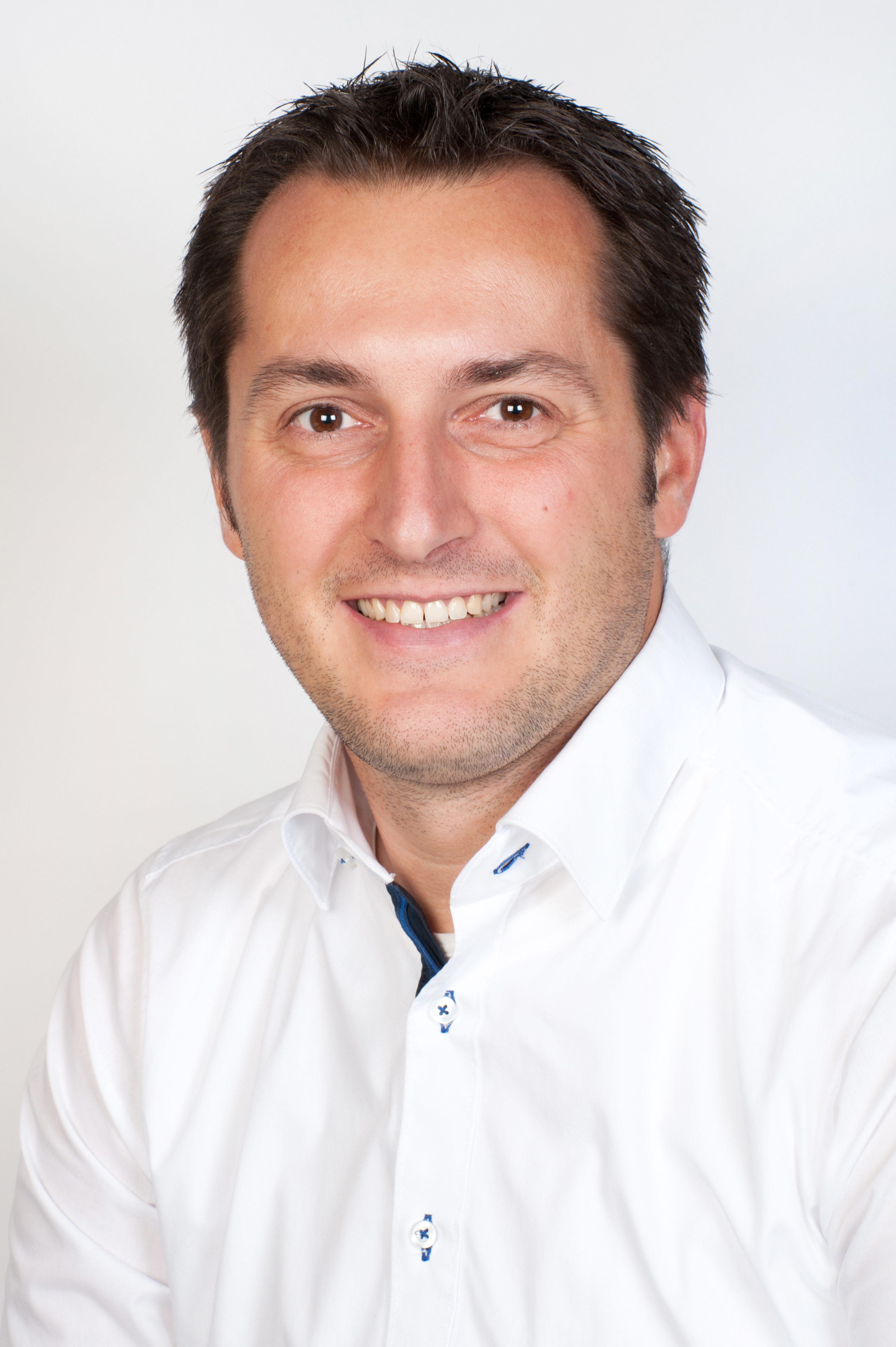}}]{Stephan Reuter} received the Diploma degree (equivalent to M.Sc. degree) and the Dr.-Ing. degree (equivalent to Ph.D.) in electrical engineering from Ulm University, Germany, in 2008 and 2014, respectively. Since 2008 he is a research assistant at the Institute of Measurement, Control and Microtechnology at Ulm University. He received the best Student Paper award at FUSION 2014 and the Uni-DAS award for an excellent PhD thesis in the field of driver assistance systems. His main research topics are sensor data fusion, multi-object tracking, extended object tracking, environment perception for intelligent vehicles, and sensor data processing.
			\end{IEEEbiography}

\end{document}